\documentclass[journal]{IEEEtran}
\usepackage{graphicx}
\usepackage{amsmath, amssymb}
\usepackage{caption, subcaption}
\usepackage{float}
\usepackage{multirow}
\usepackage{hyperref}
\usepackage{soul,color}
\usepackage{cite}
\begin{document}
%%%%%%%%%%%%%%%%%%%%%%%%%%%%%%%%%%%%
\title{Attention Mechanism Meets with Hybrid Dense Network for Hyperspectral Image Classification}
%%%%%%%%%%%%%%%%%%%%%%%%%%%%%%%%%%%%
\author{Muhammad Ahmad, Adil Mehmood Khan, Manuel Mazzara, Salvatore Distefano, Swalpa Kumar Roy, Xin Wu
\thanks{M. Ahmad is with the Department of Computer Science, National University of Computer and Emerging Sciences, Islamabad, Chiniot-Faisalabad Campus, Chiniot 35400, Pakistan. (e-mail: mahmad00@gmail.com)}
\thanks{A. M. Khan is with the Institute of Data Science and Artificial Intelligence, Innopolis University, Innopolis, 420500, Russia. (e-mail: a.khan@innopolis.ru)}
\thanks{M. Mazzara is with Institute of Software Development and Engineering, Innopolis University, Innopolis, 420500, Russia. (e-mail: m.mazzara@innopolis.ru)}
\thanks{S. Distefano is with  Dipartimento di Matematica e Informatica---MIFT, University of Messina, Messina 98121, Italy. (e-mail: sdistefano@unime.it)}
\thanks{S. K. Roy is with the Department of Computer Science and Engineering at Jalpaiguri Government Engineering College, West Bengal 735102, India (e-mail: swalpa@cse.jgec.ac.in).}
\thanks{X. Wu is with the School of Information and Electronics, Beijing Institute of Technology, 100081 Beijing, China (e-mail: xin.wu@bit.edu.cn).}
}
%%%%%%%%%%%%%%%%%%%%%%%%%%%%%%%%%%%%
\markboth{Preprint Submitted to arXiv, 2022}
{M.Ahmad \MakeLowercase{\textit{et al.}}: Hybrid Dense Attention Network}
%%%%%%%%%%%%%%%%%%%%%%%%%%%%%%%%%%%%
\maketitle
%%%%%%%%%%%%%%%%%%%%%%%%%%%%%%%%%%%%
\begin{abstract}
The nonlinear relation between the spectral information and the corresponding objects (complex physiognomies) makes pixel-wise classification challenging for conventional methods. To deal with nonlinearity issues in Hyperspectral Image Classification (HISC), Convolutional Neural Networks (CNN) are more suitable, indeed. However, fixed kernel sizes make traditional CNN too specific, neither flexible nor conducive to feature learning, thus impacting on the classification accuracy. The convolution of different kernel size networks may overcome this problem by capturing more discriminating and relevant information. In light of this, the proposed solution aims at combining the core idea of 3D and 2D Inception net with the Attention mechanism to boost the HSIC CNN performance in a hybrid scenario. The resulting  \textit{attention-fused hybrid network} (AfNet) is based on three attention-fused parallel hybrid sub-nets with different kernels in each block repeatedly using high-level features to enhance the final ground-truth maps. In short, AfNet is able to selectively filter out the discriminative features critical for classification. Several tests on HSI datasets provided competitive results for AfNet compared to state-of-the-art models. The proposed pipeline achieved, indeed, an overall accuracy of 97\% for the Indian Pines, 100\% for Botswana, 99\% for Pavia University, Pavia Center, and Salinas datasets.
% The source code will be made available at https://github.com/mahmad00.
\end{abstract}
%%%%%%%%%%%%%%%%%%%%%%%%%%%%%%%%%%%%
%%%%%%%%%%%%%%%%%%%%%%%%%%%%%%%%%%%%
\IEEEpeerreviewmaketitle
%%%%%%%%%%%%%%%%%%%%%%%%%%%%%%%%%%%%
\section{Introduction}
\label{Sec.1}

\IEEEPARstart{H}{yperspectral} Imaging (HSI) systems based on collections of electromagnetic spectrum, ranging from visible to near-infrared region, reflected by the objects of interest \cite{hong2021interpretable}. The images thus obtained are usually generated by a pre-configured HSI camera installed on either mobile (e.g. satellites, drones, aircrafts) or static (e.g., indoor, rooms, labs) setups depending upon the problem at hand \cite{app10175955, 9205804, app10196862, hong2021more, app10217783, s21093045, HKNCAA}. Thereby, HSI sensors gather a huge amount of data from hundreds of contiguous spectral bands \cite{rasti2020feature, 4786582}.

Such big and rich HSI dataset, including the different spectral bands data related by the (spatial) geo-located position, may contain hidden information and patterns. HSI Classification (HSIC) \cite{6472238} aims at discover, detect, identify and recognize such patterns. However, the spectral dataset size usually increases combinatorially with the problem size (e.g. the area, the resolution), leading to the \textit{curse of dimensionality} and thus making traditional HSIC methods inefficient \cite{ArXiv2}. To mitigate such issues, several dimensionality reduction and feature selection techniques have been proposed \cite{hong2019learning,ArXiv3}. Conventional feature extraction/selection methods rely on hand-crafted features, however, due to the spatial variability of spectral information \cite{hong2019augmented}, the extraction of discriminative and most informative features is still a big challenge \cite{ArXiv4}.

Hand-crafted features may be insubstantial in the case of HSI data, therefore, it is challenging to achieve a trade-off between discriminability and robustness on features which also considerably differ on the different HSI datasets. Furthermore, the process of  feature design and selection is strongly affected by the designer and architect knowledge and skills \cite{9307220, ArXiv4, hong2019learnable,yao2022sparsity}. To such a purpose, an automatic approach to hierarchically identify the features was developed by Hinton back in 2006 \cite{9307220, Hinton504,hong2020x}, based on a deep learning model developed in growing semantic layers until a desirable representation is achieved. Similarly, other models have been proposed on feature learning and classification as well \cite{zhang2011adaptive,9307220, 7272047,hong2021multimodal}, automatically learning and improving the underlying system representation from the available data without any prior knowledge. They can extract both linear and non-linear features, thus capable of handling HSI data in both spatial and spectral domains \cite{zhang2012real}.

In nature, HSI datasets are usually non-linear due to the undesired light-scattering phenomena given by land cover objects and particles in the atmosphere \cite{ArXiv1}. Thus rendering the use of linear transformation or feature learning methods \cite{ahmad2017graph} for HSIC. To overcome the non-linearity issues, Convolutional Neural Network (CNN) was proposed to extract both high as well as low-level features which ultimately lead to the extraction of abstract and invariant features \cite{7182258}. As a result, 2D CNN  achieved remarkable performance but unfortunately not so good for HSIC due to the missing channel-related information, i.e. 2D CNN are not able to learn spectrally discriminative features. Unlike 2D CNN, 3D ones can jointly extract the spatial-spectral information for HSIC providing higher accuracy than 2D CNN \cite{Gao2020}. However, 3D CNN models are computationally-/time-intensive due to the high number of parameters involved by 3D convolutional filters on each layer. 

For instance, in \cite{8061020}, a Spatial-Spectral Residual Network (SSRN) implemented a 3D CNN residual network based on ResNet \cite{7780459}. Despite SSRN achieves remarkable classification results, the summation method used to aggregate features at each layer requires output feature maps to have consistent scale as the residual feature maps. Hence each layer has self weights which overall lead to an explosion of network parameters \cite{ArXiv1}. Thereby, high accuracy comes at expenses of the computational power, i.e. SSRN is more  complex than traditional 3D CNN. Similarly, \cite{9307220} proposed a fast and compact 3D CNN (FC-3D-CNN) model to overcome the limitations of computational cost and reduce the number of training parameters. FC-3D-CNN achieves better results in a computationally efficient manner than SSRN due to the reduced spectral information used in the experimental process.

In general, CNN models tend to poorly perform  especially in the case of pixels of different classes but similar texture over contiguous spectral bands \cite{He17}. To deal with that, \cite{8736016, ArXiv5, ArXiv6} proposed hybrid models that combine the power of 2D and 3D convolutional layers to extract high and low-level features i.e., extraction of abstract and invariant features. The hybrid models achieve better accuracy than  state-of-the-art 2D and 3D CNN solutions. Despite the high accuracy, hybrid models still require a large number of parameters as compared to the 3D CNN network \cite{9307220} while, on the other hand, 3D CNN/SRNN has a longer processing time than hybrid models. Therefore, inception models have proved that the network topology significantly affects the complexity and the accuracy \cite{4801616}. Recently, graph convolutional network (GCN) shows the superiority in HSCI, Hong \textit{et al.} \cite{hong2021graph} proposed a novel GCN in a mini-batch fashion, called miniGCN, which solves the problem of large-scale graph computation and learning. Apart from the complexity and accuracy trade-off, all the inception models have one common property i.e., a split-transform-merge strategy which proved to be a good strategy for HSIC. Traditional 2D, 3D, hybrid, and inception models exploit the fixed convolution kernel size, however, HSI class distribution is complicated thus conventional CNN with fixed kernel size is not flexible enough. Convolutions with different spatial sizes may capture more discriminative and important information for pixel-based HSIC.

Nowadays, an attention mechanism has been extensively used to suppress redundant information while extracting features for classification. SENet \cite{Hu17} was the first network proposed to suppress the redundant features by weighting channel direction features. The work \cite{Woo18} proposed CBAM (convolutional block attention module) combines spatial attention with channel attention through pooling whereas the work \cite{Wang18} proposed a NLNet which combines the convolution operations with matrix multiplication operations to capture the long-range relationship in the global space. A combination of SENEt and NLNet was proposed in \cite{Cao19}, which consists of a simplified lightweight module GCNet for effectively extracting global context. Very recently, a novel transformer framework was rapidly developed and its spectral version, called SpectralFormer, was for the first time proposed with the application to HSIC, yielding state-of-the-art classification performance \cite{hong2021spectralformer}.

Though, attention networks have achieved remarkable results for HSIC based on the internal architecture of the attention module. These works, to some extent, put attention weights in either one or two dimensions and ignore the rest of the HSI dimensions. For instance, single and double attention networks were proposed in \cite{Fang19, Ma19}, in which the work \cite{Fang19} only consider the spectral information and ignore the spatial information, whereas, the work \cite{Ma19} was proposed to reduce the interference between the spatial and channel information. The work \cite{Pan19} was proposed to jointly explore the spectral and spatial information, where spectral-spatial dimensions were weighted by the spectral-spatial attention module. The combination of attention in more or less in one or two dimensions may improve the performance, however, it is highly recommended to integrate all channel information for better classification. 

Wang, et.al. \cite{Wang20} significantly improved the squeeze and excitation structure attention mechanism proposed in \cite{Hu17}, reducing the model complexity by a local cross-channel interaction strategy without any preprocessing, i.e., dimensional reduction. Zheng, et.al. \cite{Zheng19} worked to overcome the limitations of inconsistent class ratio and over-parameterization using a stratified sample-based training strategy. While the spectral attention module was proposed to render the soft band selection process to further refine the redundant spectrum information. However, all these spatial or spectral attention models are to some extent isolated in which the spatial attention ignores to make a connection between spectral dimension whereas, the spectral attention ignores the spatial connection and uses only the correlation between different spectral bands. 

Moreover, it has been proven fact that accuracy improves while increasing the depth of the model, thus require more parameters and higher computational burden which makes optimization an NP hard problem. Therefore, to overcome the aforesaid issues, this work explicitly investigates the possibilities of combining the core idea of 2D as well as 3D inception models into an Hybrid attention architecture to boost the pixel-based HSIC performance. We tested the model on several publicly available HSI datasets which shows competitive results compared to the state-of-the-art methods. Our proposed pipeline achieved an overall accuracy of 97\% for the Indian Pines dataset, 100\% for Botswana, 99\% for both Pavia University, Pavia Center, and Salinas datasets, respectively. In a nutshell, the contributions made in this work as summarized as follows.

\begin{enumerate}
    \item A densely connected hybrid inception net is proposed to enrich the spatial-spectral feature learning process. Different from the conventional inception model which consists of a single branch in each block, the proposed densely connected hybrid blocks are composed of parallel multiple sized filters which significantly improves the propagation and reuse of features with less number of tune-able parameters. Moreover, the proposed hybrid inception net comprehensively utilizes features of different scales from HSI dataset. 
    
    \item A triple-branch attention fusion block is introduced which boosts the robustness of the discriminative network. As compared with recently published attention blocks for HSIC, the proposed pipeline simultaneously model interactions across different spectral bands and spatial regions by re-weighting the significance of features. The triple-branch attention block adaptively emphasizes the important information and significantly suppresses the redundant and ineffective information. 
    
    \item A hybrid spectral convolutional block is used to reduce the required number of parameters for the HSIC model. Moreover, the activation inducted convolutional layer can further improve the non-linear representation capacity of the whole network.
\end{enumerate}

%%%%%%%%%%%%%%%%%%%%%%%%%%%%%%%%%%%%
The rest of the paper is structured as follows. Section \ref{sec2} presents the pipeline proposed in this paper. Section \ref{sec3} describes the experimental settings along the metrics used to compute the accuracies. Sections \ref{sec4.1}-\ref{sec4.3} exhibits the experimental datasets used to conduct the experiments and to validate the proposed methodology. Moreover, sections \ref{sec4.1}-\ref{sec4.3} demonstrates the experimental results as compared with the state-of-the-art methods proposed in the literature. Finally, section \ref{sec6} concludes the paper with possible future research directions.

%%%%%%%%%%%%%%%%%%%%%%%%%%%%%%%%%%%%
\section{Problem Formulation}
\label{sec2}

Lets assume $R = [r_1, r_2, r_3,..., r_L]^T \in R^{L \times n}$ be the HSI cube, where $n = N \times M$ samples associated with $C$ classes and $L$ band images. Each $r_i = (r_i,c_j)$ where $c_j$ be the class associated with $r_i$ sample. In nature, $r_i$ exhibit high intra-class variability and inter-class similarity, overlapping and nested regions. Therefore, HSI cube has been first divided into small spatial patches to overcome the aforesaid issues. For each patch, the ground truths are formed based on the central pixel of the patch. Principle Component Analysis (PCA) has been used before creating the patches which eliminate the redundancy among the band images i.e. $L \rightarrow B$, where $B \ll L$.

The patching process creates neighboring patches $P \in R^{S \times S \times B}$ centered at the spatial location $(a,b)$ covering $S \times S$ spatial windows \cite{9307220, 8736016}. The total of $Z$ patches given by $(U - S + 1) \times (V - S + 1)$. Thus, in total, these patches cover the width from $\frac{a + (S - 1)}{2}$ to $\frac{a - (S - 1)}{2}$ and height from $\frac{b + (S - 1)}{2}$ to $\frac{b - (S - 1)}{2}$ \cite{8736016}. The $Z$ patches are first convolved with a kernel function which computes the sum of the dot product between the input patch and kernel function to introduces the nonlinearity \cite{9307220, 8736016, Ying17,hong2020x}. The activation maps for spatial-spectral position $(x,y,z)$ at $i-th$ feature map and $j-th$ layer can be represented as $v_{i,j}^{(x,y,z)}$.

\begin{multline*}
v_{i,j}^{x,y,z} = ReLu \bigg(\sum_{\tau = 1}^{d_{i-1}} \sum_{\rho = -\gamma}^{\gamma} \sum_{\phi = -\delta}^{\delta} \sum_{\lambda = -\nu}^{\nu} \\ w_{i, j, \tau}^{\rho, \phi, \lambda} \times v_{(i-1), \tau}^{(x+\rho), (y+\phi), (z+\lambda)} + b_{i,j} \bigg)
\end{multline*}
where $d_{i - 1}$ be the total number of feature maps at $(i - 1)-th$ layer, $w_{i, j}$ and $b_{i,j}$ be the depth of the kernel and bias, respectively. Moreover, $2\gamma +1$, $2\delta +1$, and $2\nu + 1$ be the height, width, and depth of the kernel \cite{9307220}. Similarly, 2D modules do the same process with 2D input as well as the 2D kernel function. In both 3D and 2D layers, the kernel is striding over the input to cover the whole spatial dimension. More specifically, as the proposed model combines the power of 3D and 2D kernel functions, thus, 2D convolution $V^{x, y}_{i, j}$ represents the activation value of $i-th$ feature map at $(x, y)$ spatial position on $j-th$ layer and can be formulated as $v^{x, y}_{i, j}$ and finally can be formed as follows.

\begin{multline*}
v^{x, y}_{i, j} = ReLu\bigg(\sum_{\tau=1}^{d_{l-1}} \sum_{\rho = -\gamma}^{\gamma} \sum_{\phi = -\delta}^{\delta} \\ 
w_{i, j, \tau}^{\rho, \phi} \times v_{i-1, \tau}^{(x+\rho), (y+\phi)} + b_{i, j}\bigg)
\label{Eq1}
\end{multline*}
where $2\gamma + 1$ and $2\delta + 1$ be the height and width of the kernel, respectively. In short, the proposed hybrid AfNet convolutional filters are as follows with the input of $9 \times 9 \times 15$. The size of 3D filters are $3D_1 = (7 \times 7 \times 9)$, $K_1^1 = 7, K_1^2 = 7$, $K_1^3 = 9$, $3D_2 = (5 \times 5 \times 7)$, $K_2^1 = 5, K_2^2 = 5$, $K_2^3 = 7$, and $3D_3 = (3 \times 3 \times 5)$, $K_3^1 = 3, K_3^2 = 3$ and $K_3^3 = 5$ for each layer on each block with different number of filters, i.e., $(30, 20, 10)$ for first block, $(40, 20, 10)$ for second block and $(60, 30, 10)$ for third block. Similarly the size of 2D filters are $2D_1 = (3 \times 3)$, $K_1^1 = 3, K_1^2 = 3$, $2D_2 = (3 \times 3)$, $K_2^1 = 3, K_2^2 = 3$, and $2D_3 = (1 \times 1)$, $K_3^1 = 1, K_3^2 = 1$ for each layer on each block with different number of filters, i.e., $(16, 32, 64)$ for first block, $(16, 32, 64)$ for second block and $(16, 32, 64)$ for third block. A 2D fusion module has been used to incorporate the information learned hierarchically at different blocks. Finally, a 2D convolutional layer is used with $(1 \times 1)$ kernel size with total of $128$ filters to better represent the low to high level information. 

\begin{figure}[!hbt]
	\centering
	\includegraphics[width=0.48\textwidth]{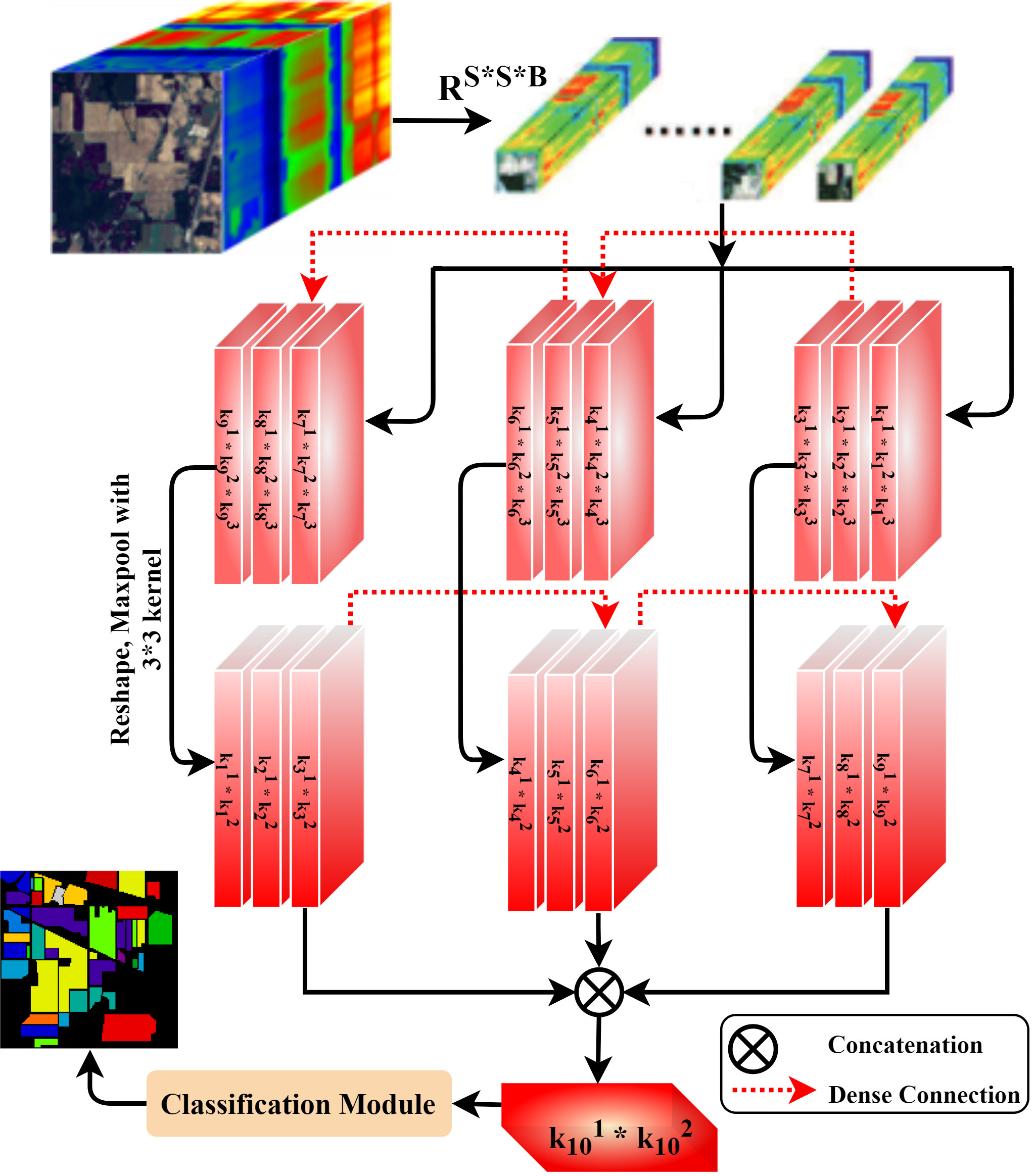}
    \caption{AfNet for HSIC, where $S \times S$ be the Height and Width of the patch, $B$ refer to the number of Bands, $(k_{1 \rightarrow 9}^{1} \times k_{1 \rightarrow 9}^{2} \times k_{1 \rightarrow 9}^{3})$ are 3D Conv layers and $(k_{1 \rightarrow 10}^{1} \times k_{1 \rightarrow 10}^{2})$ are 2D Conv. layers. }
\label{Fig.0}
\end{figure}

To decrease the number of spectral-spatial feature maps, nineteen densely connected 3D and 2D convolutional layers are used prior to the flatten layer to make sure the convolutional process discriminate the spatial information while considering different spectral bands with no loss and less number of parameters to boost the performance \cite{9307220}. The weights are randomized initially and optimized using Adam optimizer based on back-propagation with a soft-max loss function. Later the randomized weights are updated using a mini-batch size of $256$ for $50$ epochs. The overall structure of the proposed hybrid AfNet using the Indian Pines dataset as an example is presented in Figure \ref{Fig.0}.

%%%%%%%%%%%%%%%%%%%%%%%%%%%%%%%%%%%%

%%%%%%%%%%%%%%%%%%%%%%%%%%%%%%%%%%%%
\subsection{Dense Connections and Attention Blocks}

CNN extracts different features with different characteristics on each layer in which lower and middle layer features have relatively high resolution and encompass more location and detailed information. However, may have lower semantics and more noise due to fewer convolutional layers passing through. Since we know that the high-level features hold strong semantic information with low resolution and poor perception capability. Therefore, cross-layer feature fusion can be considered as an effective strategy to preserve the quality features and ultimately improve classification performance \cite{hong2021more}.

Dense connectivity (e.g., different kinds of connectivity patterns irrespective of the traditional network) has been first proposed in DenseNet and widely used framework in many real-life applications. Traditionally, all layers are connected one after another, in order to maximize the feature information flow between layers. In this hierarchy, each layer accepts the features of all previous layers in front of it as input and passes its output to the subsequent layer. However, irrespective of the traditional dense connections, this paper proposes dense connections along with the attention mechanism among different Inception Network blocks as shown in Figure \ref{Fig.0}. From Figure \ref{Fig.0}, one can see that the output of the 2nd convolutional layer of block-1 is densely connected with the 2nd layer of block-2 where the other layers of each block are densely connected traditionally as well followed by the nonlinear transformation in both cases.

Let us assume $X_i$ be the output of the ith block and $X_0$ be the output of the previous convolutional block. Thus the output of the ith convolutional block is not only related to the output of (i-1)th block but also includes the middle output of all previous blocks. Similarly, each 3D CNN block is densely connected along with the attention mechanism with the 2D CNN blocks, respectively as just explained above. However, while connecting the 3D feature maps with 2D feature maps, a reshape and max-pooling with $3 \times 3$ kernel is used. Finally, a concatenation (fusion) layer is deployed to fuse the output maps obtained from all three blocks, and subsequently, a 2D convolutional layer is used to further refine the feature maps obtained from the densely connected network. The attention blocks are flexible in the proposed model and can be positioned anywhere in the network as explained in Figure \ref{Fig.0}.

%%%%%%%%%%%%%%%%%%%%%%%%%%%%%%%%%%%%
\subsection{Overview}

For high-level intuition of the proposed model, the overall structure has been illustrated in Figure \ref{Fig.0} in which each block of the network is densely connected with the help of an attention mechanism. The proposed AfNet is an end-to-end framework for HSIC in which the input $R^{L \times n}$ is the original HSI dataset and output is regarded as the probability of each HSI pixel for $c_j$ classes.

Since HSI composed of hundreds of contiguous spectral bands, and some of these are highly correlated with each other, which provides no new information for classification. Moreover, some noisy bands exist in HSI, therefore, to eliminate the noisy and redundant bands, PCA transformation has been applied before feature learning and classification which significantly reduces the processing time and memory capacity as well. 

The HSI cube has been divided into overlapping 3D cubes to take full advantage of spatial-spectral information present in the HSI dataset. These 3D cubes are composed of the target pixel and its neighboring pixels to perform pixel-level feature learning and classification. Let's assume each 3D patch formed from HSI is $R^{S \times S \times B}$ where $S \times S$ denotes the spatial size of each patch and $B$ be the number of PCs preserved in the spectral dimension. In AfNet, a densely connected convolutional (3D and 2D) layers with rectified linear unit (ReLu) and without normalization is a basic building unit of the network. 

The 3D patches are first to go through nine attention-based 3D inter-connected convolutional layers. The obtained feature maps are further fed to nine attention-based 2D inter-connected convolutional layers which are designed to improve the propagation and reuse of features with a fewer number of tune-able parameters. Moreover, a hybrid structure can discriminate the spatial information while considering different spectral bands with no loss to increase the generalization performance. On the other hand, the proposed network is built by stacking multiscale parallel filters of various sizes. On top of that, the attention module has been added to handle the skip connections from the first block to the second and all other subsequent blocks, which improves the flow of information. As a result, the fused features provide surpass feature space as compared to the stacked single-sized convolutional layers. 

Another way around, attention blocks are incorporated to selectively filter out the information (features) which are critical for classification, i.e., weakening information that is useless for classification has been eliminated which leads to obtaining a feature representation that is more discriminative to get a higher class probability for each pixel. Following the fusion, a 2D convolutional layer is employed to aggregate the obtained features once again. Afterward, feature maps are converted into feature vectors by flattening, and finally, class labels are generated using Softmax.

%%%%%%%%%%%%%%%%%%%%%%%%%%%%%%%%%%%%
\section{Experimental Settings}
\label{sec3}

The experimental results explained in this work have been obtained through Google Colab, an online platform to execute any python environment with Graphical Process Unit (GPU), up to 358$+$ GB of cloud storage, and up to 25 GB of Random Access Memory (RAM). In all the stated experiments (not only for the proposed method but for all comparative methods as well), the training, validation, and test sets are randomly divided into three parts; 15\%/15\%/70\% (i.e., Training/Validation/Test sets).

For fair comparative analysis and to make the claims more reliable, the learning rate for all experimental methods are set to 0.001, Relu as the activation function for all hidden layers except the final (output) later on which Softmax activation function is used. Patch size is set to $9 \times 9$ for all experimental results, and $15$ most informative spectral dimensions were selected using PCA to reduce the computational burden in terms of time and space. All the models are evaluated on 100 epochs without batch normalization. 

To compute the experimental results, several metrics such as Average Accuracy (AA), Overall Accuracy (OA), and Kappa ($\kappa$) have been used. Where $\kappa$ metric is known as a statistical metric that considered the mutual information regarding a strong agreement among the classification and ground-truth maps. AA represents the average class-wise classification performance, whereas the OA is computed as the number of correctly classified examples out of the total test examples.

 \begin{equation}
     AA = \frac{TP + TN}{TP + TN + FN}
\end{equation}

\begin{equation}
     OA = \frac{1}{C} \sum_{i = 1}^C TP_i
\end{equation}

\begin{equation}
     Kappa~(\kappa) = \frac{P_o - P_e}{1 - P_e}
\end{equation}
where

\begin{multline*}
 P_e = \bigg( \frac{FN + TN}{TP + FN + FP + TN} \times \\ 
 \frac{FP + TN}{TP + FN + FP  + TN} \bigg) \\
+ \frac{TP + FN}{TP + FN + FP + TN}
\end{multline*}

\begin{equation*}
     P_o = \frac{TP + TN}{TP + FN + FP + TN}
\end{equation*}
where $C$ be the total number of classes presents in HSI dataset. Moreover, $FP$, $FN$, $TP$, and $TN$  are false positive, false negative, true positive, and false negative, respectively. 

%%%%%%%%%%%%%%%%%%%%%%%%%%%%%%%%%%%%
\subsection{Experimental Datasets and Initial Experiments}
\label{sec4.1}

Several publicly available Hyperspectral datasets have been used to evaluate the performance of our proposed pipeline. These datasets are acquired at different times and locations using different sensors such as Hyperion NASA EO-1 satellite, Reflective Optics System Imaging Spectrometer (ROSIS), and Airborne Visible/Infrared Imaging Spectrometer (AVIRIS) sensor. Further information regarding the experimental datasets can be found from \cite{9307220, PURR1947, Grupointel,hong2021joint}. As earlier explained, the performance of our proposed pipeline is evaluated using four publicly available HSI datasets namely Indian Pines, Pavia University, Botswana, and Salinas. For each of the above datasets, the samples are randomly splited into three subsets i.e., training, validation, and test sets. Table \ref{Tab.2} provides a summary description of each dataset used in the following experiments.

%%%%%%%%%%%%%%%%%%%%%%%%%%%%%%%%%%%%
\begin{table*}[!hbt]
    \centering
    \caption{Summary of the HIS datasets used in the following experiments.}
    \resizebox{\textwidth}{!}{
    \begin{tabular}{lccccccccc} \hline \hline 
        \textbf{Dataset} & \textbf{Year} & \textbf{Source} & \textbf{Spatial dimensions} & \textbf{Spectral} & \textbf{Wavelength} & \textbf{Samples} & \textbf{Classes} & \textbf{Sensor} & \textbf{Resolution} \\ \hline \hline 
        Botswana & 2001-2004 & NASA EO-1 & $1496 \times 256$ & 242 bands & 400-2500 & 3248 & 14 & Satellite & 30 \\ 
        Indian Pines & 1992 & NASA AVIRIS & $145 \times 145$ & 220 bands & 400 - 2500 & 10249 & 16 & Aerial & 20 \\ 
        Salinas & 1998 & NASA AVIRIS & $512 \times 217$ & 224 bands & 360 - 2500 & 54129 & 16 & Aerial & 3.7 \\ 
        Pavia University & 2001 & ROSIS-03 sensor & $610 \times 610$ & 115 bands & 430 - 860 & 42776 & 9 & Aerial & 1.3 \\ 
        \hline \hline 
    \end{tabular}}
    \label{Tab.2}
\end{table*}
%%%%%%%%%%%%%%%%%%%%%%%%%%%%%%%%%%%%

%%%%%%%%%%%%%%%%%%%%%%%%%%%%%%%%%%%%
\subsubsection{\textbf{Indian Pines}}

Indian Pines dataset was acquired back in 1992, June 12 over the Purdue University Agronomy farms, northwest of West Lafayette and the surrounding area using AVIRIS sensor. This dataset was mainly acquired to facilitate soil research being initiated by Prof. Marion Baumgardner and his graduate students. Indian Creek and Pine Creek watersheds contain most of the part of the dataset thus known by Indian Pines and include two flight lines: 1): Flown East-West, 2): Flown North-South. There are three $2 \times 2$ miles intensive test sites with the area as; 1): Northern portion of North-South flight line, 2): Near the center, 3): Southern portion. 

\begin{figure}[!hbt]
\centering
	\begin{subfigure}{0.09\textwidth}
		\includegraphics[width=0.99\textwidth]{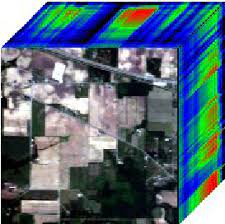}
		\centering
		\caption{IP Cube}
		\label{Fig3A}
	\end{subfigure}
	\begin{subfigure}{0.09\textwidth}
		\includegraphics[width=0.99\textwidth]{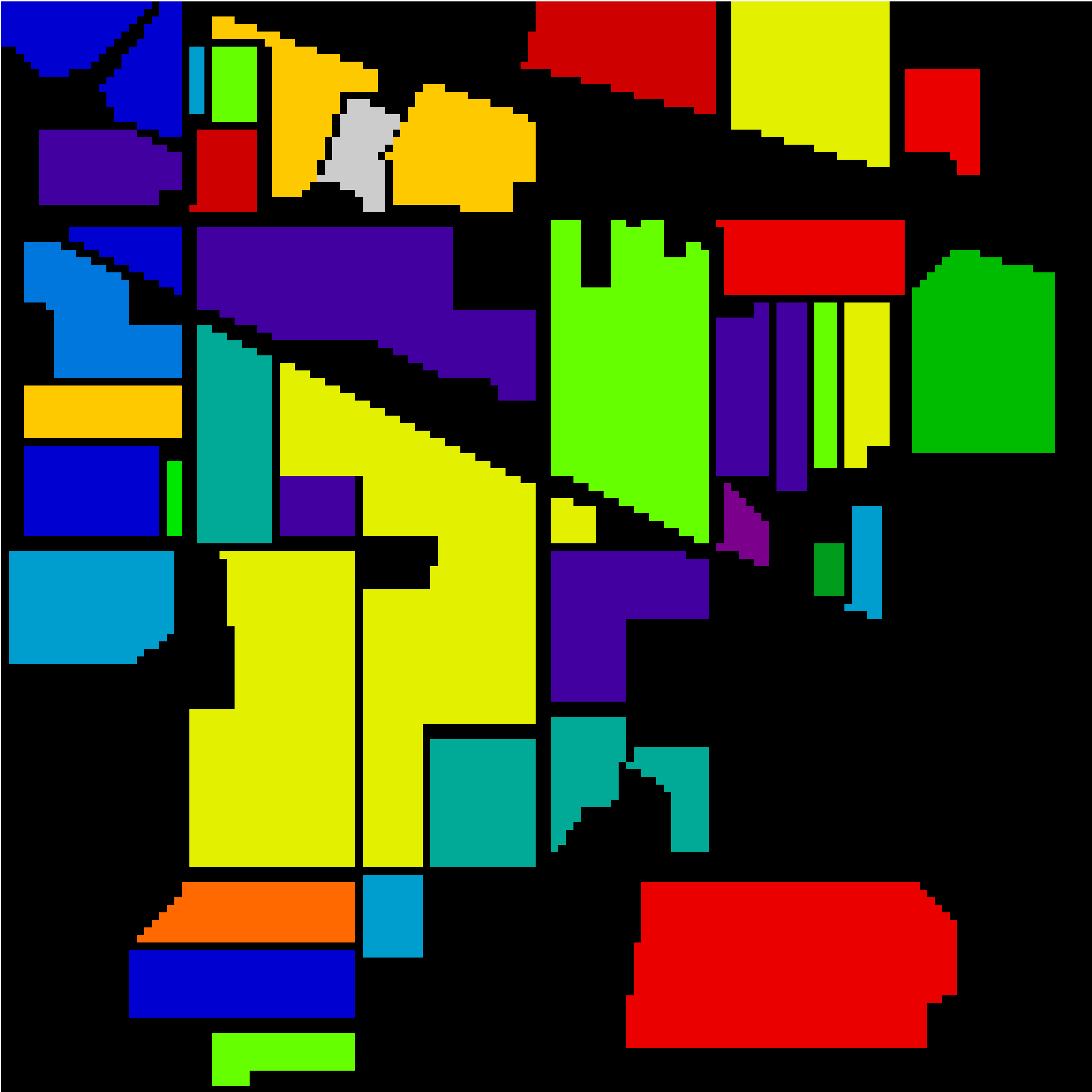}
		\centering
		\caption{GT}
		\label{Fig3B}
	\end{subfigure}
	\begin{subfigure}{0.09\textwidth}
		\includegraphics[width=0.99\textwidth]{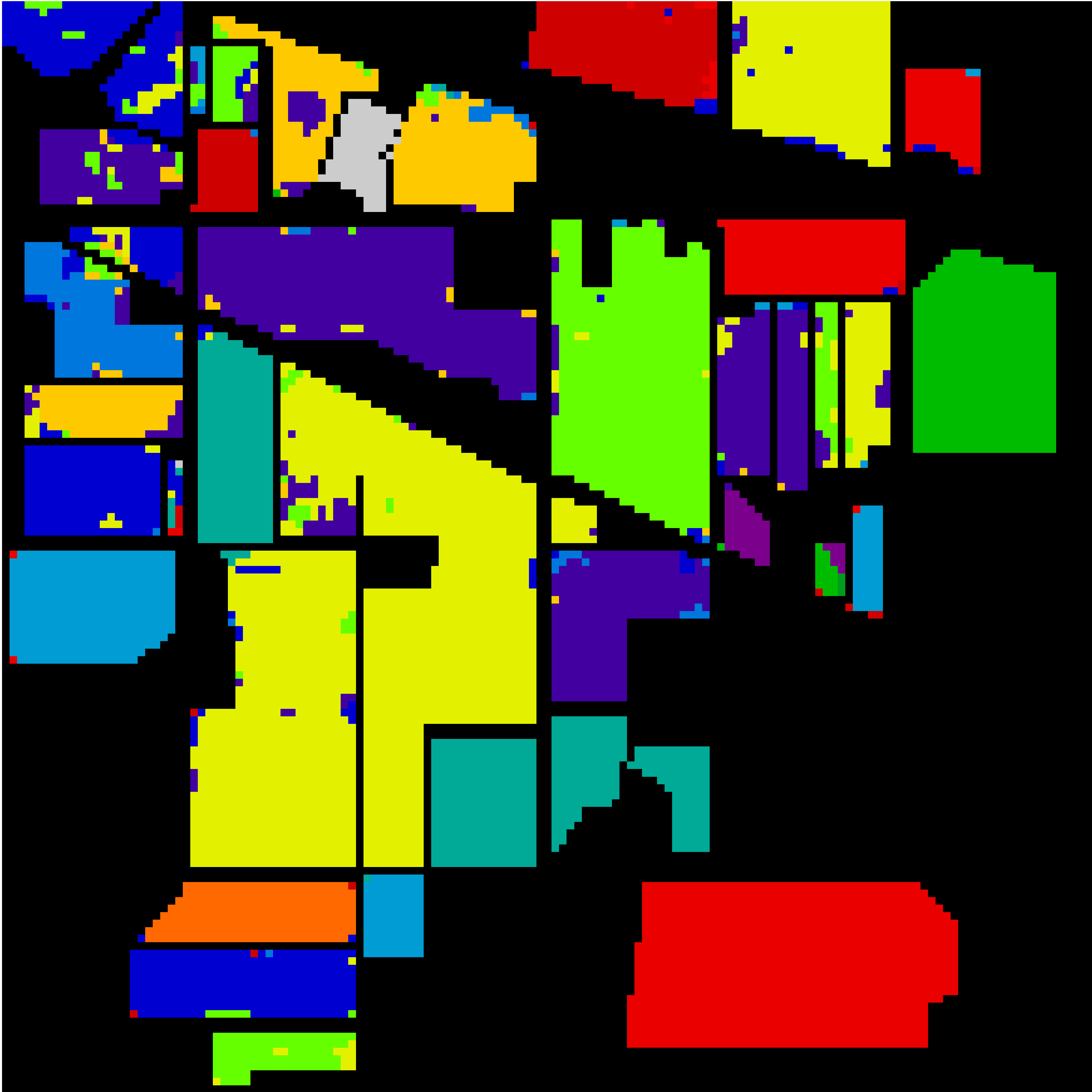}
		\centering
		\caption{Hybrid} 
		\label{Fig3C}
	\end{subfigure}
	\begin{subfigure}{0.09\textwidth}
		\includegraphics[width=0.99\textwidth]{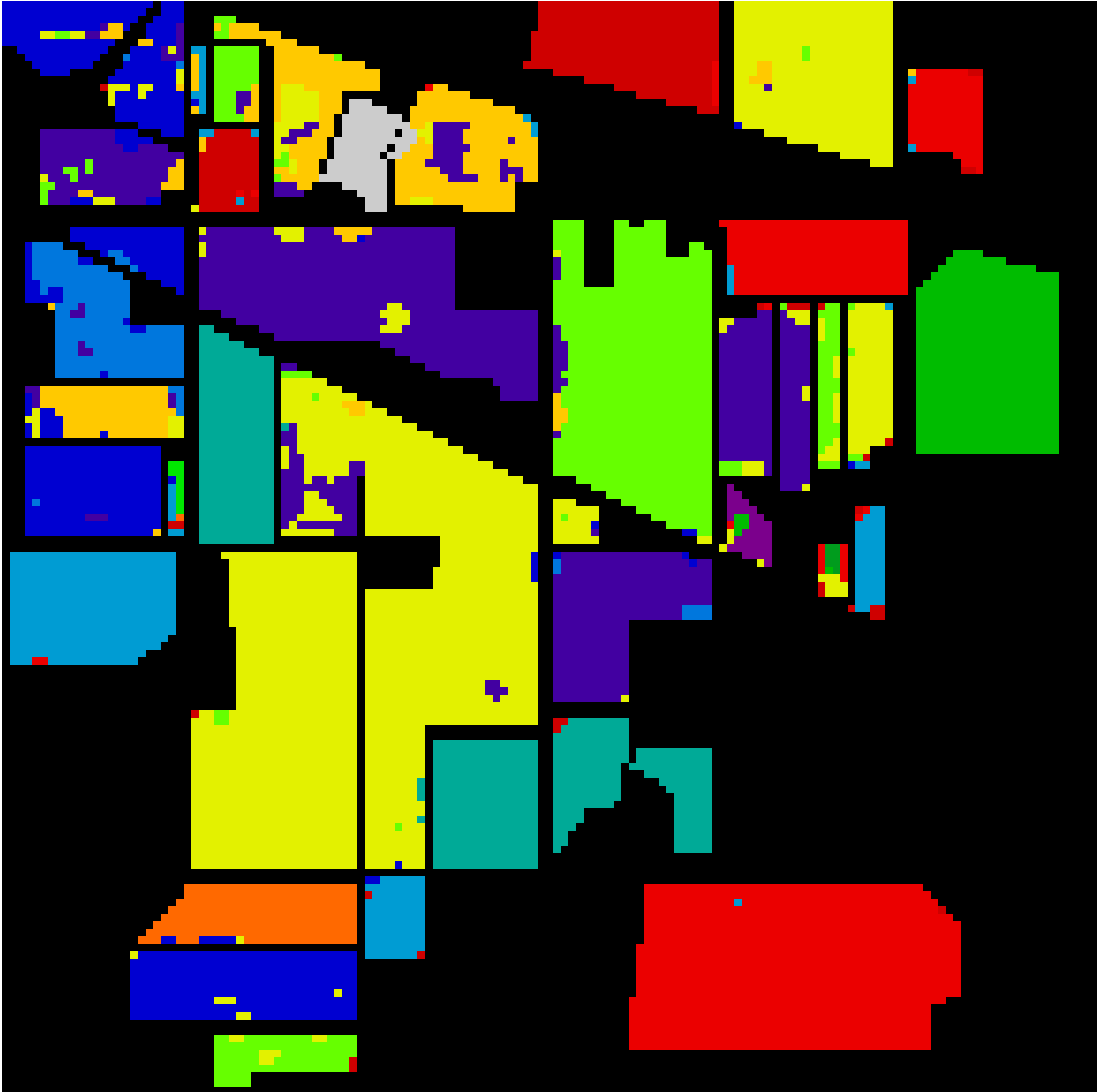}
		\centering
		\caption{2D} 
		\label{Fig3D}
	\end{subfigure}
	\begin{subfigure}{0.09\textwidth}
		\includegraphics[width=0.99\textwidth]{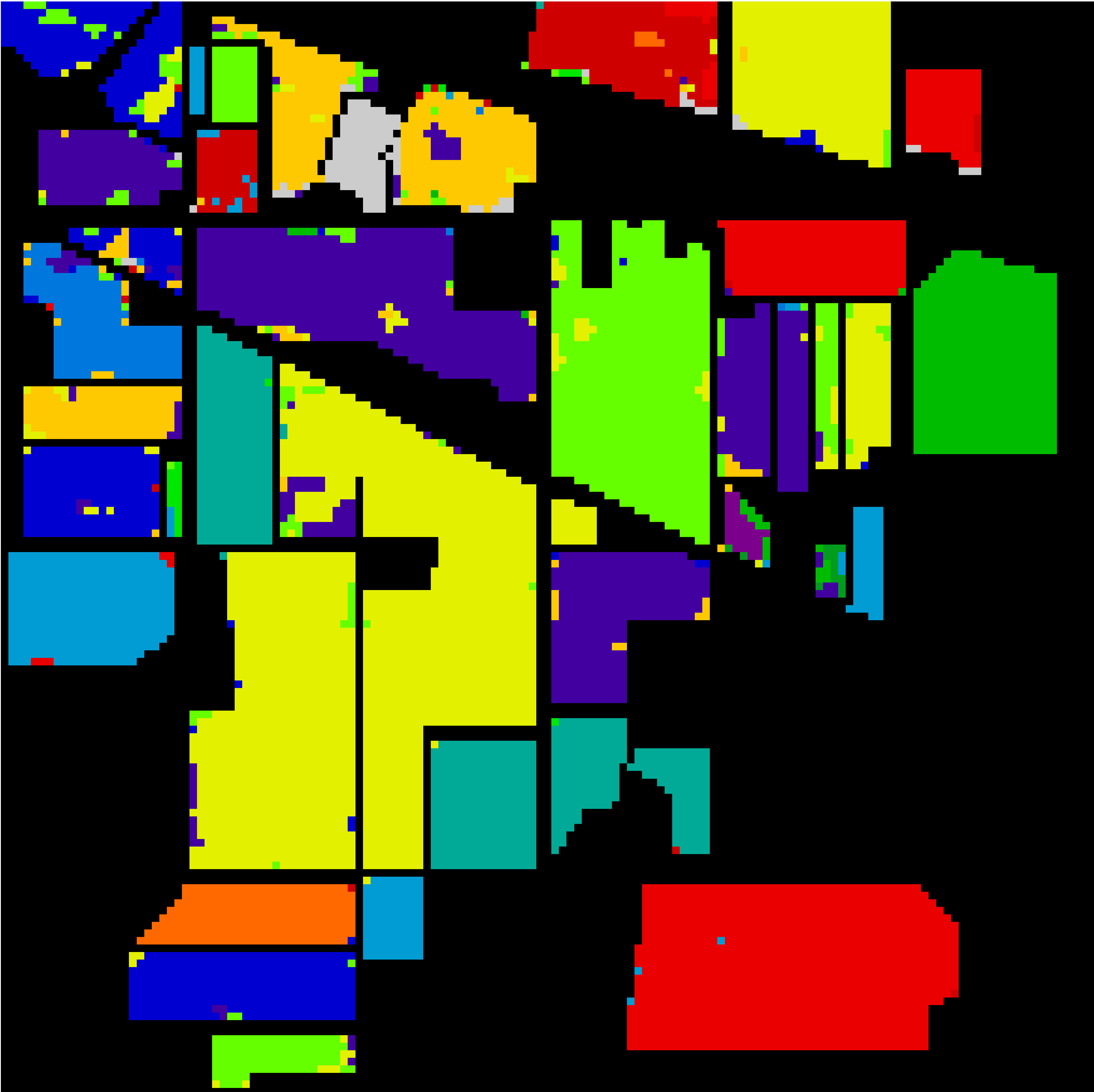}
		\centering
		\caption{3D}
		\label{Fig3E}
	\end{subfigure}
\caption{Indian Pines Dataset illustration in-terms of HSI Cube and Ground Truth (GT) label maps. The classification maps obtained using Hybrid AfNet with overall accuracy of 92.82 (\%), 2D Inception Net with overall accuracy of 94.18 (\%) and 3D Inception Net with overall accuracy of 81.51 (\%).}
\label{Fig.3}
\end{figure}

Indian Pines dataset consists of $145 \times 145$ spatial dimensions per spectral band and in total 224 spectral bands in the wavelength range $0.4-2.5 10^{-6}$ meters. The scene used in this research is a subset of a larger scene. It consists of 1/3 forest, 2/3 agriculture, and natural perennial vegetation, a rail line, two major dual-lane highways, low-density housing, other structures, and small roads as earlier explained. Some crops e.g., corn, soybeans are in the early stages of growth i.e., less than 5\% coverage due to the reason that the dataset was acquired in June. The ground truths are available and distinguished into 16 non-mutual exclusive classes. The image cube and true ground truths label maps are shown in Figures \ref{Fig3A} and \ref{Fig3B} whereas Figures \ref{Fig3C}, \ref{Fig3D}, and \ref{Fig3E} show the classification performance in terms of classification maps (ground truth label maps) for three different models, i.e. Hybrid Attention Fused Network (AfNet), 3D Attention Inception Net and 2D Attention Inception Net. These maps clearly show that the proposed method performs better than 3D as well as 2D Attention Inception Networks. The higher accuracies are emphasized.

%%%%%%%%%%%%%%%%%%%%%%%%%%%%%%%%%%%%

%%%%%%%%%%%%%%%%%%%%%%%%%%%%%%%%%%%%
\subsubsection{\textbf{Botswana}}

The Hyperion NASA EO-1 satellite acquired the Botswana dataset over OkavangoDelta, Botswana back in 2001-2004. The WO-1 sensor acquired the subject data at 30 m pixel resolution over a 7.7 km strip in 242 spectral bands covering the 400-2500 nm portion of the spectrum in 10 nm window. 

\begin{figure}[!hbt]
	\begin{subfigure}{0.45\textwidth}
	    \centering
		\includegraphics[angle=90,width=0.90\textwidth]{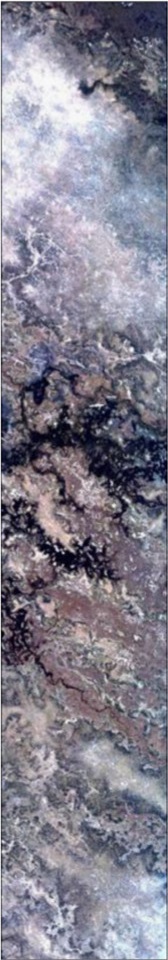}
		\caption{BS Cube}
		\label{Fig4A}
	\end{subfigure}
	\begin{subfigure}{0.45\textwidth}
	    \centering
		\includegraphics[angle=90,width=0.90\textwidth]{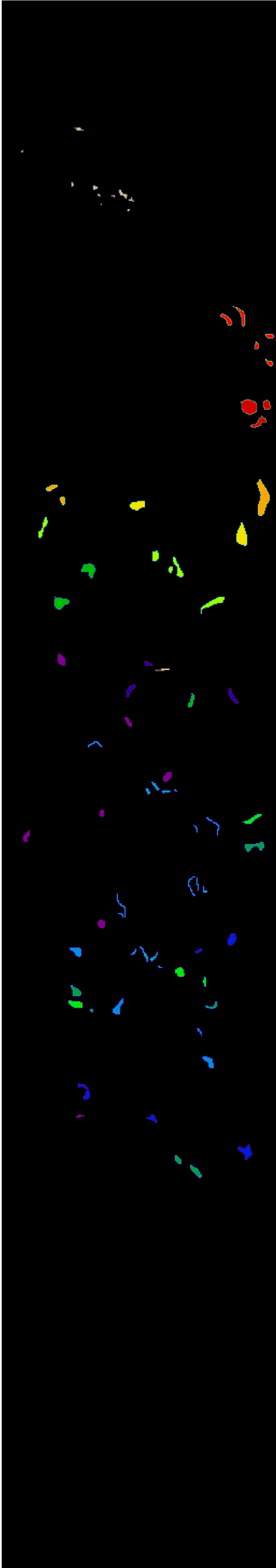}
		\caption{GT}
		\label{Fig4B}
	\end{subfigure}
	\begin{subfigure}{0.45\textwidth}
		\includegraphics[angle=90,width=0.90\textwidth]{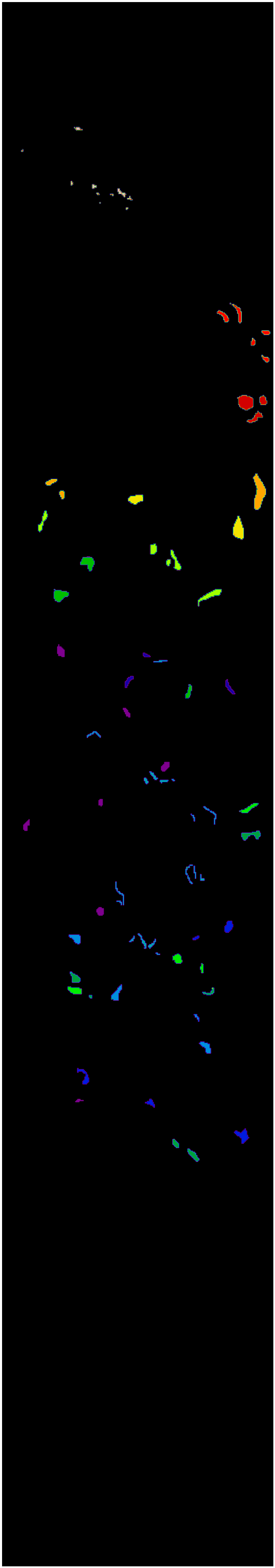}
		\centering
		\caption{Hybrid}
		\label{Fig4C}
	\end{subfigure}
	\begin{subfigure}{0.45\textwidth}
		\includegraphics[angle=90,width=0.90\textwidth]{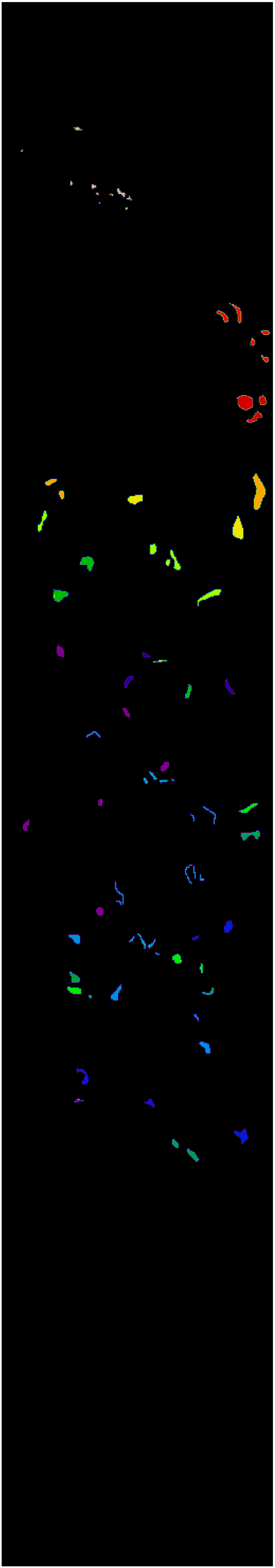}
		\centering
		\caption{2D INet}
		\label{Fig4D}
	\end{subfigure}
	\begin{subfigure}{0.45\textwidth}
		\includegraphics[angle=90,width=0.90\textwidth]{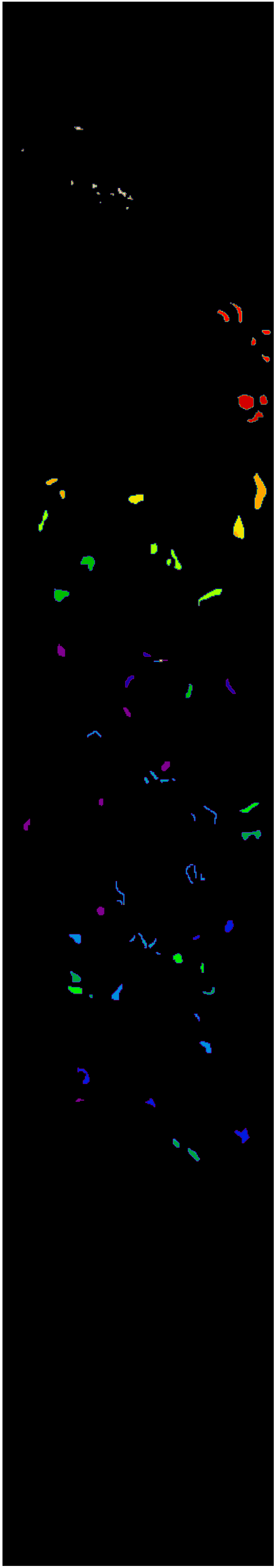}
		\centering
		\caption{3D INet}
		\label{Fig4E}
	\end{subfigure}
\caption{Botswana Dataset illustration in-terms of HSI Cube and Ground Truth (GT) label maps. The classification maps obtained using Hybrid AfNet with overall accuracy of \textbf{98.94 (\%)}, 2D Inception Net with overall accuracy of 98.81 (\%) and 3D Inception Net with overall accuracy of 98.46 (\%).}
\label{Fig.4}
\end{figure}

To mitigate the effects of bad detectors, inter detector miscalibration, and intermittent anomalies, extensive preprocessing has been carried out by UT Center for space research. While processing, uncalibrated and noisy bands (i.e., water absorption features) were removed and the remaining 145 spectral bands are used for experimental purposes. The data used in this study was acquired back on May 31, 2001, and it consist of observations from 14 mutually exclusive classes which represent the land cover types in seasonal swamps, occasional swamps, and drier woodlands located in the distal portion of the Delta. The image cube and true ground truths label maps are shown in Figures \ref{Fig4A} - \ref{Fig4B} whereas Figures \ref{Fig4C} - \ref{Fig4D} show the classification performance in terms of classification maps (ground truth label maps) for three different models, i.e. Hybrid Attention Fused Network (AfNet), 3D Attention Inception Net and 2D Attention Inception Net. These maps clearly show that the proposed method performs better than 3D as well as 2D Attention Inception Networks. The higher accuracies are emphasized.

%%%%%%%%%%%%%%%%%%%%%%%%%%%%%%%%%%%%

%%%%%%%%%%%%%%%%%%%%%%%%%%%%%%%%%%%%
\subsubsection{\textbf{Pavia University}}

\begin{figure}[!hbt]
    \centering
	\begin{subfigure}{0.09\textwidth}
		\includegraphics[width=0.99\textwidth]{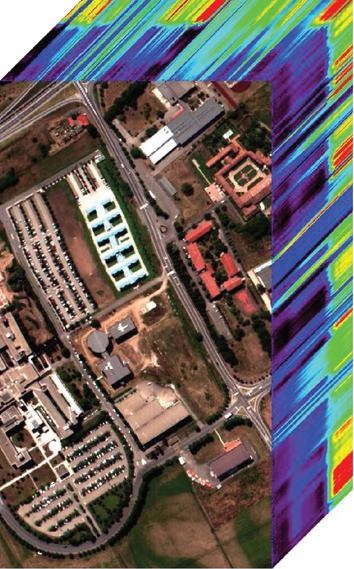}
		\caption{PU Cube} 
		\label{Fig6A}
	\end{subfigure}
	\begin{subfigure}{0.09\textwidth}
		\includegraphics[width=0.90\textwidth]{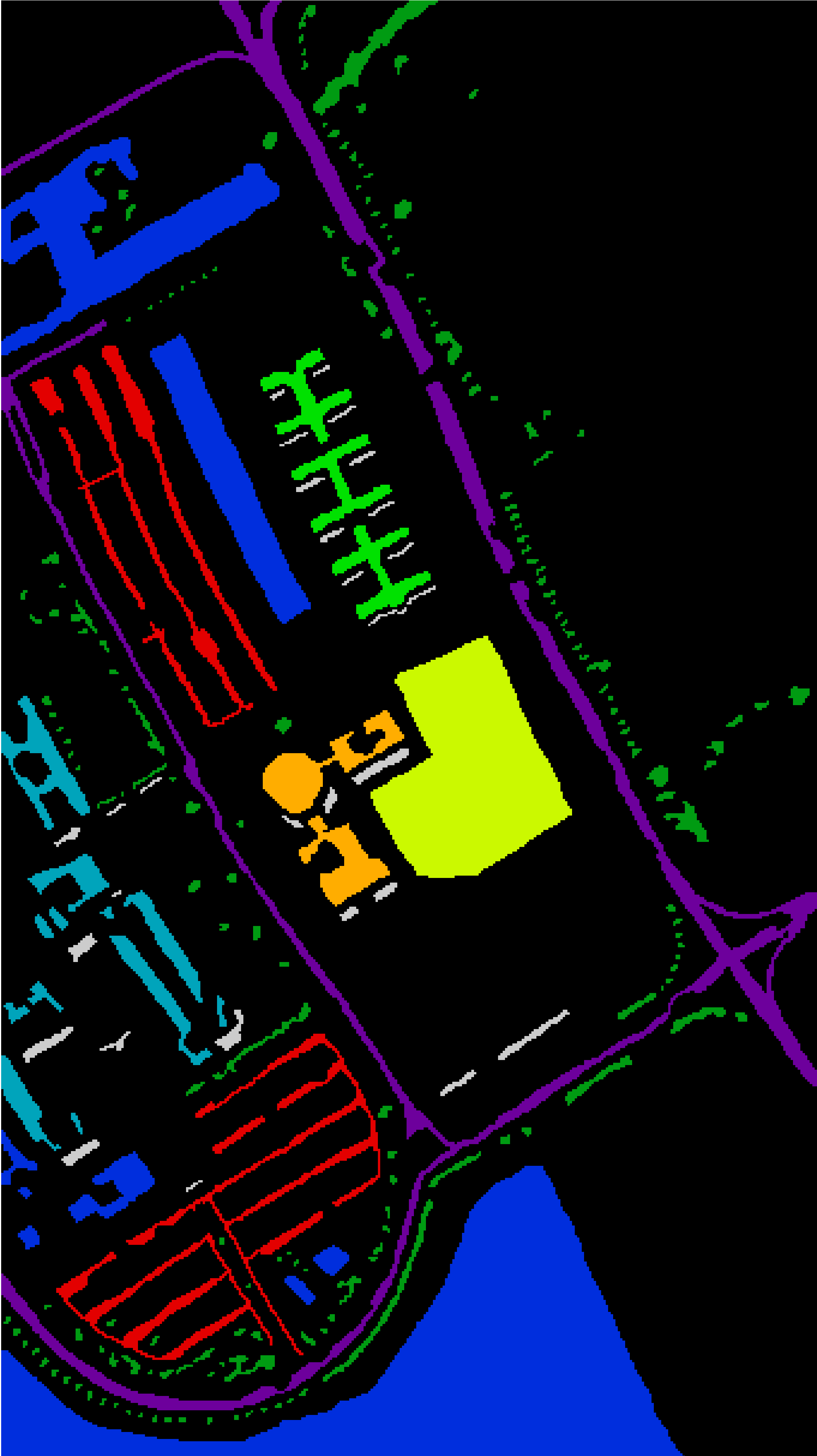}
		\caption{GT}
		\label{Fig6B}
	\end{subfigure}
	\begin{subfigure}{0.09\textwidth}
		\includegraphics[width=0.90\textwidth]{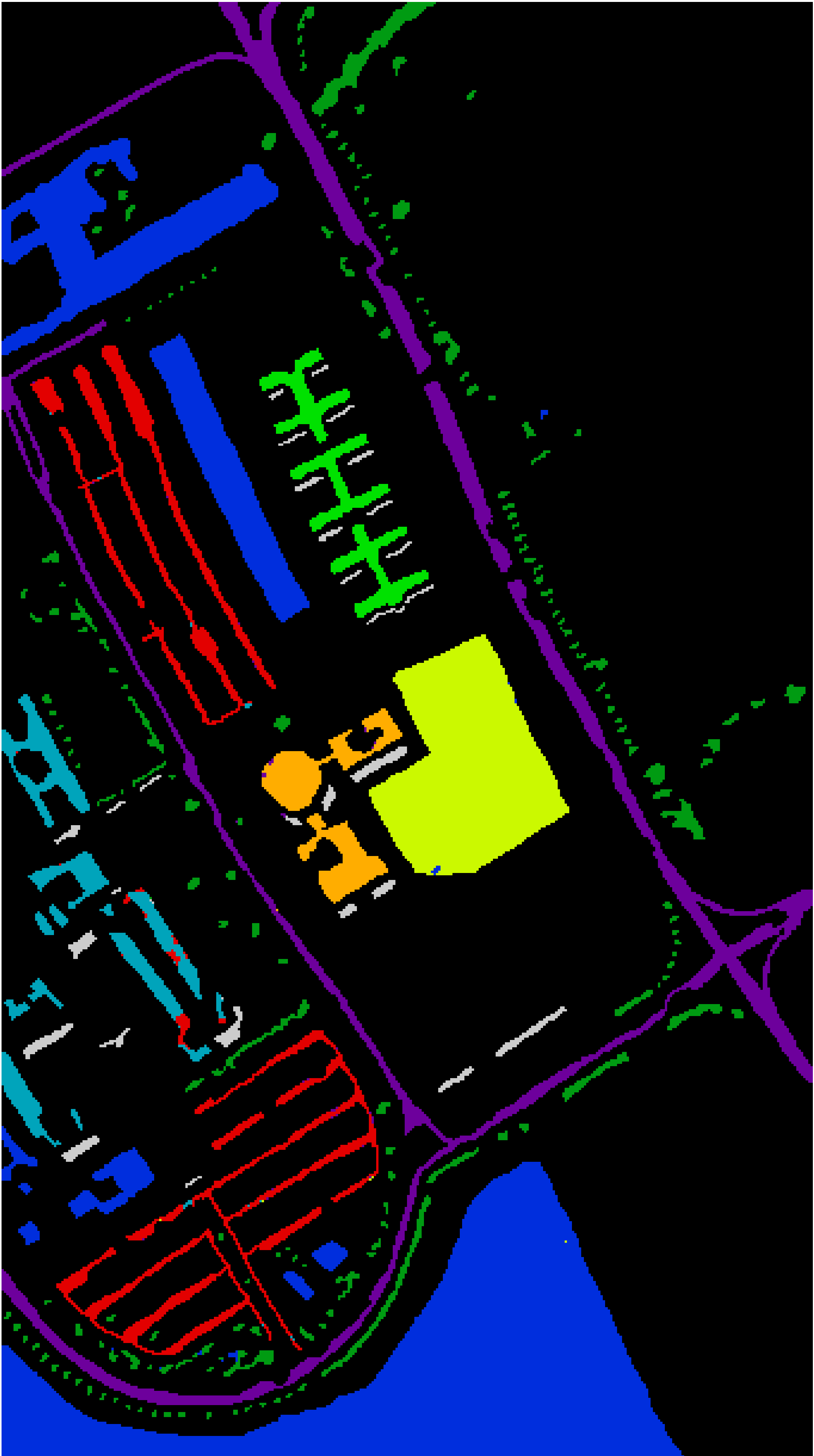}
		\centering
		\caption{Hybrid} 
		\label{Fig6C}
	\end{subfigure}
	\begin{subfigure}{0.09\textwidth}
		\includegraphics[width=0.90\textwidth]{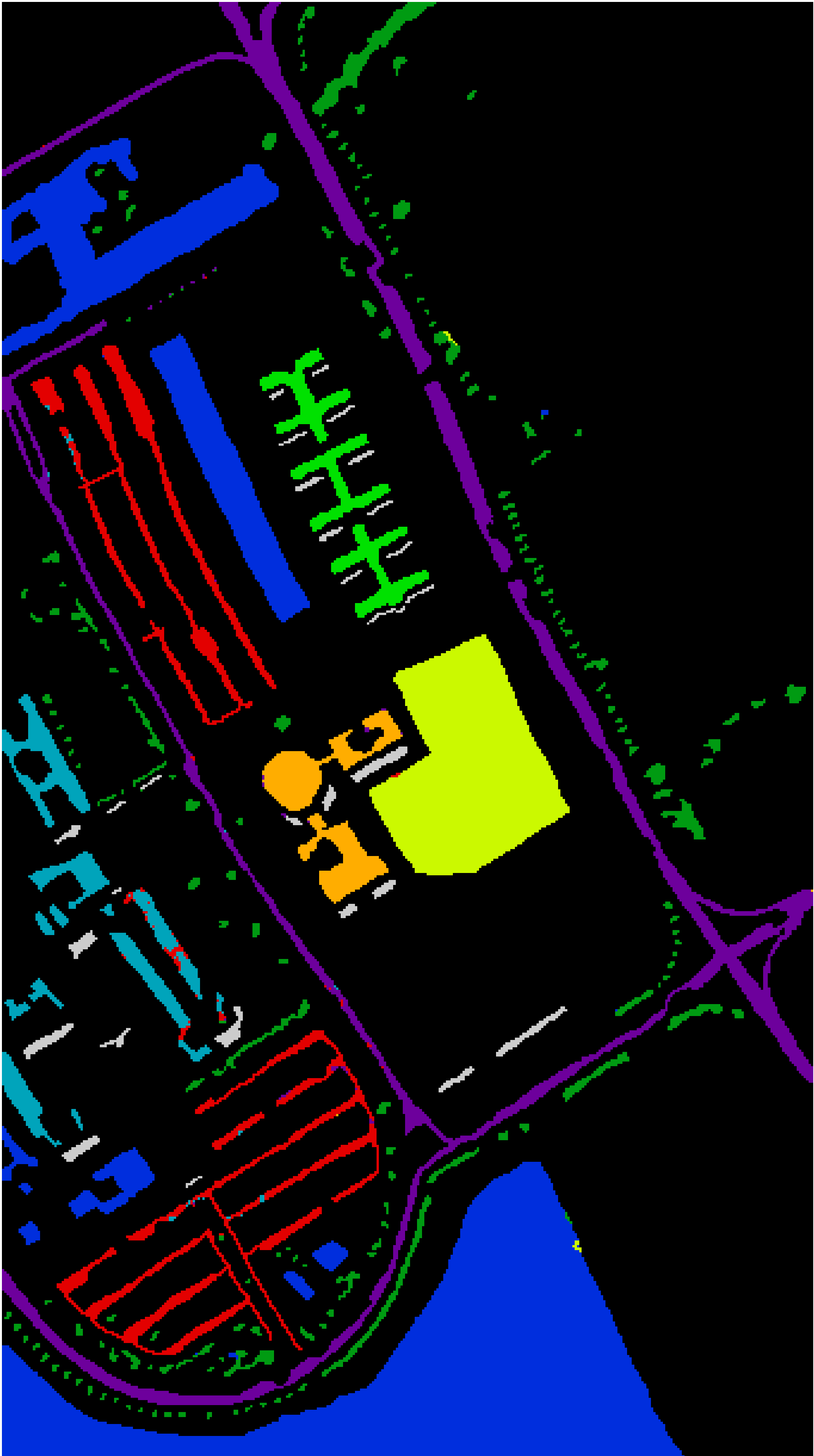}
		\centering
		\caption{2D} 
		\label{Fig6D}
	\end{subfigure}
	\begin{subfigure}{0.09\textwidth}
		\includegraphics[width=0.90\textwidth]{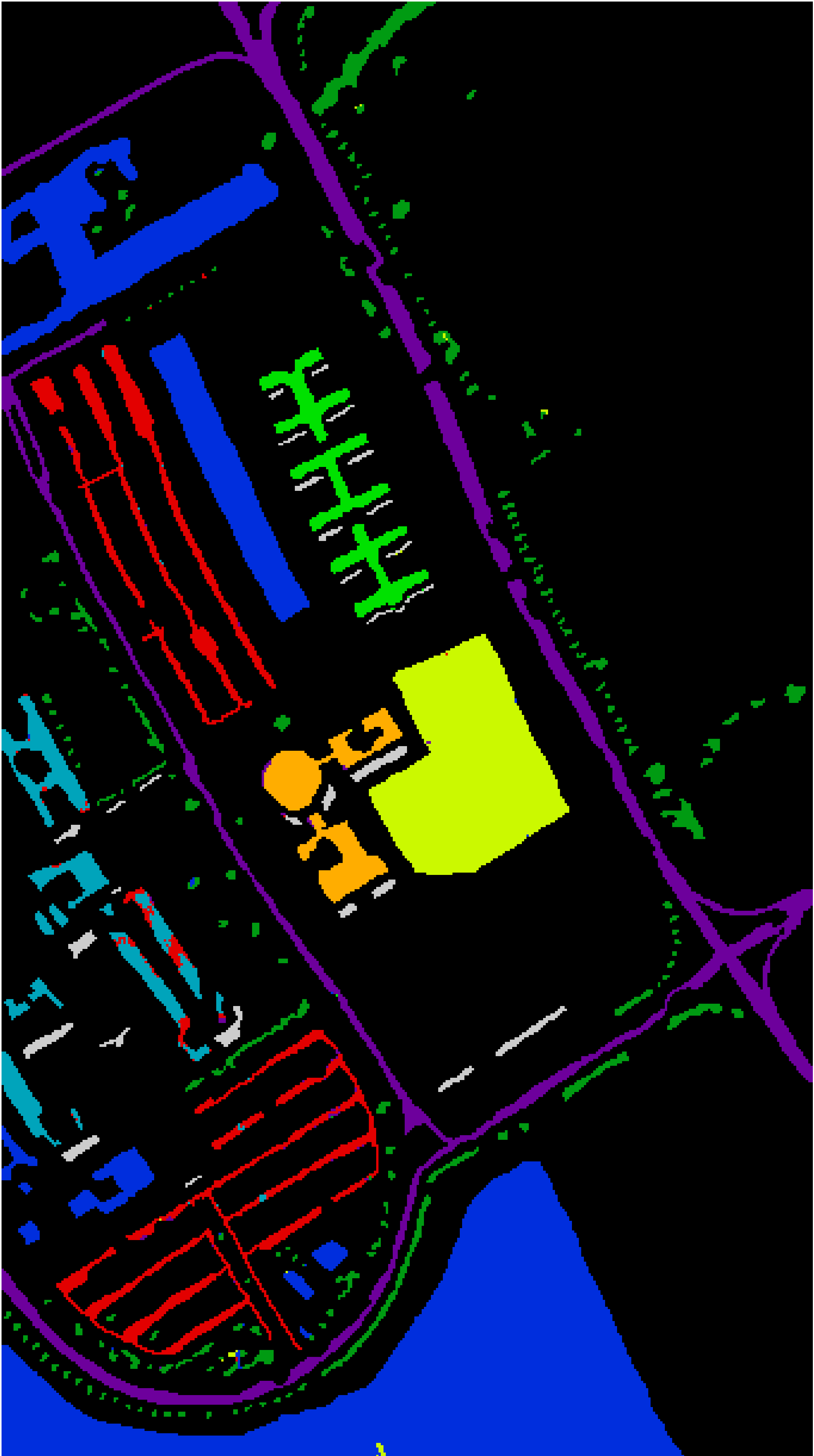}
		\centering
		\caption{3D}
		\label{Fig6E}
	\end{subfigure}
\caption{Pavia University Dataset illustration in-terms of HSI Cube and Ground Truth (GT) label maps. The classification maps obtained using Hybrid AfNet with overall accuracy of \textbf{99.27 (\%)}, 2D Inception Net with overall accuracy of 99.09 (\%) and 3D Inception Net with overall accuracy of 98.55 (\%).}
\label{Fig.6}
\end{figure}

Pavia university dataset was acquired using ROSIS sensor during a flight campaign over Pavia, northern Italy. The total number of spectral bands is 103 in which each spectral band covers $610 \times 610$ spatial dimensions per spectral band. Some of the samples contain no information in the above spatial dimensions, thus have to be discarded before the analysis. The geometric resolution is 1.3 meters. Pavia Center image ground truths differentiate 9 mutually exclusive classes. The image cube and true ground truths label maps are shown in Figures \ref{Fig6A} - \ref{Fig6B} whereas Figures \ref{Fig6C} - \ref{Fig6D} show the classification performance in terms of classification maps (ground truth label maps) for three different models, i.e. Hybrid Attention Fused Network (AfNet), 3D Attention Inception Net and 2D Attention Inception Net. These maps clearly show that the proposed method performs better than 3D as well as 2D Attention Inception Networks. The higher accuracies are emphasized.

%%%%%%%%%%%%%%%%%%%%%%%%%%%%%%%%%%%%

%%%%%%%%%%%%%%%%%%%%%%%%%%%%%%%%%%%%
\subsubsection{\textbf{Salinas}}

\begin{figure}[!hbt]
    \centering
	\begin{subfigure}{0.09\textwidth}
		\includegraphics[width=0.99\textwidth]{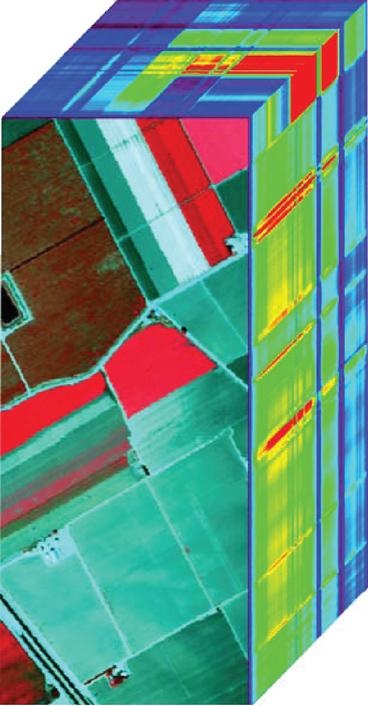}
		\caption{SA Cube} 
		\label{Fig7A}
	\end{subfigure}
	\begin{subfigure}{0.08\textwidth}
		\includegraphics[width=0.90\textwidth]{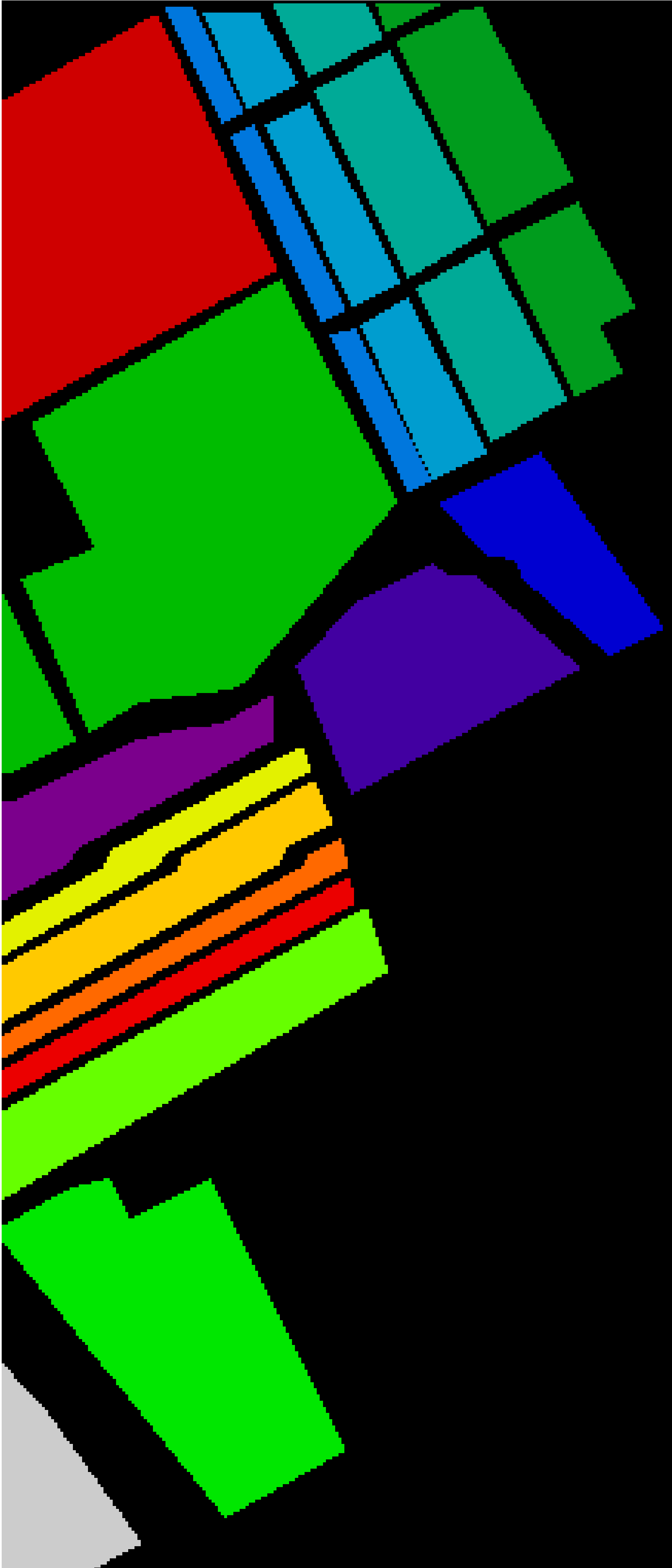}
		\caption{GT}
		\label{Fig7B}
	\end{subfigure}
	\begin{subfigure}{0.08\textwidth}
		\includegraphics[width=0.90\textwidth]{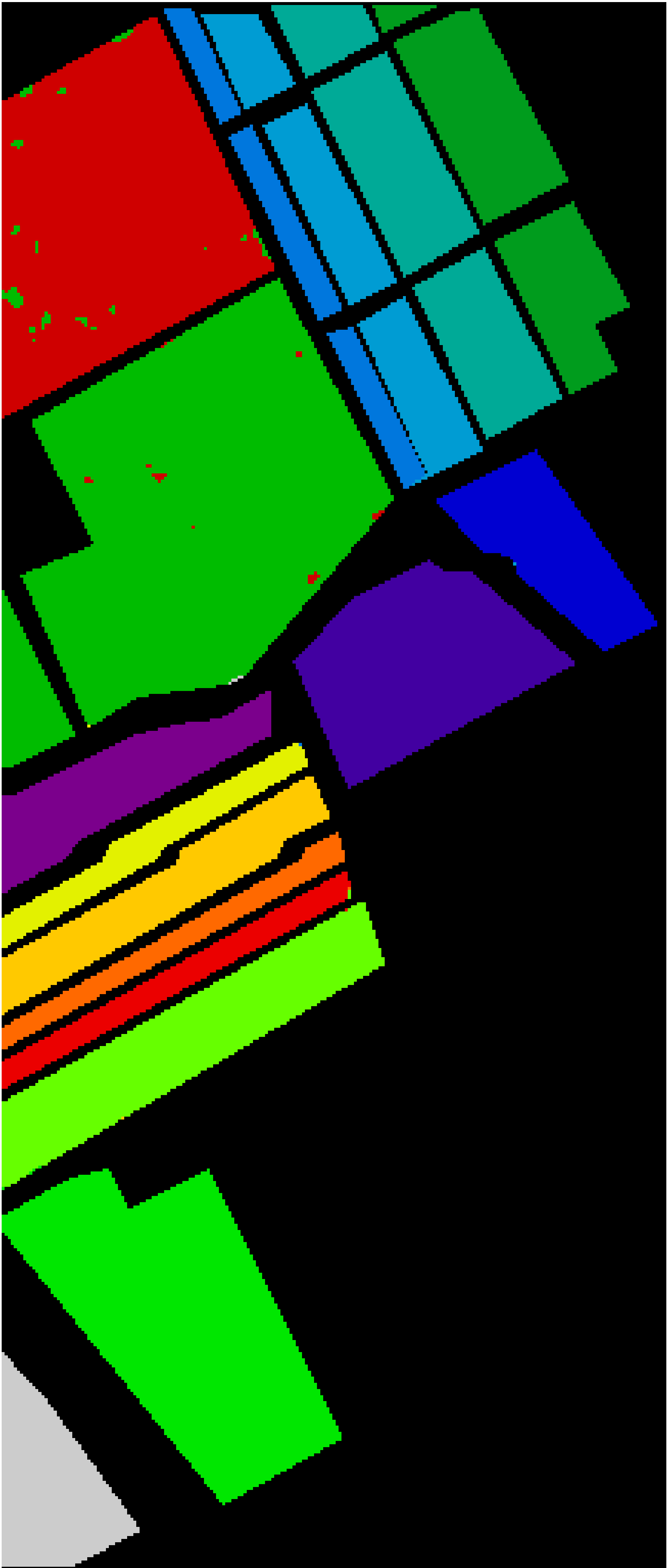}
		\centering
		\caption{Hybrid} 
		\label{Fig7C}
	\end{subfigure}
	\begin{subfigure}{0.08\textwidth}
		\includegraphics[width=0.90\textwidth]{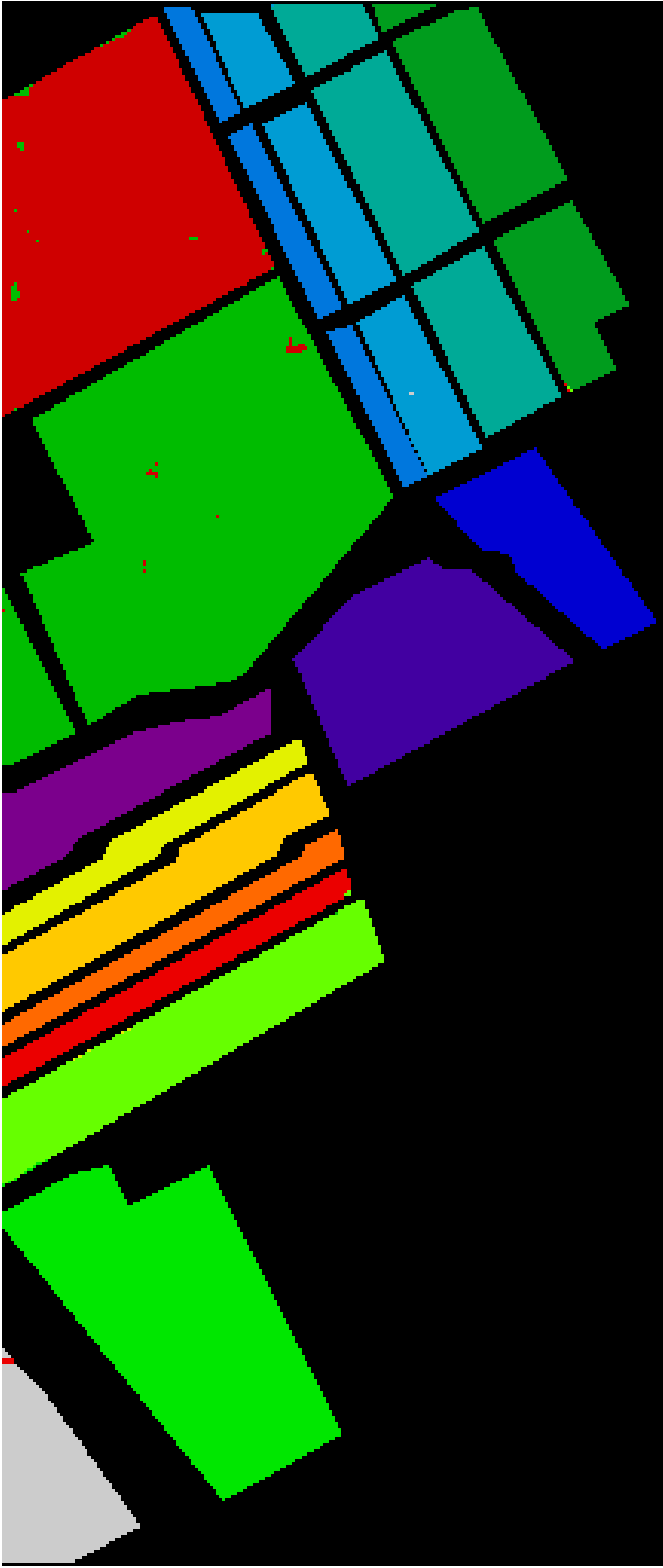}
		\centering
		\caption{2D} 
		\label{Fig7D}
	\end{subfigure}
	\begin{subfigure}{0.08\textwidth}
		\includegraphics[width=0.90\textwidth]{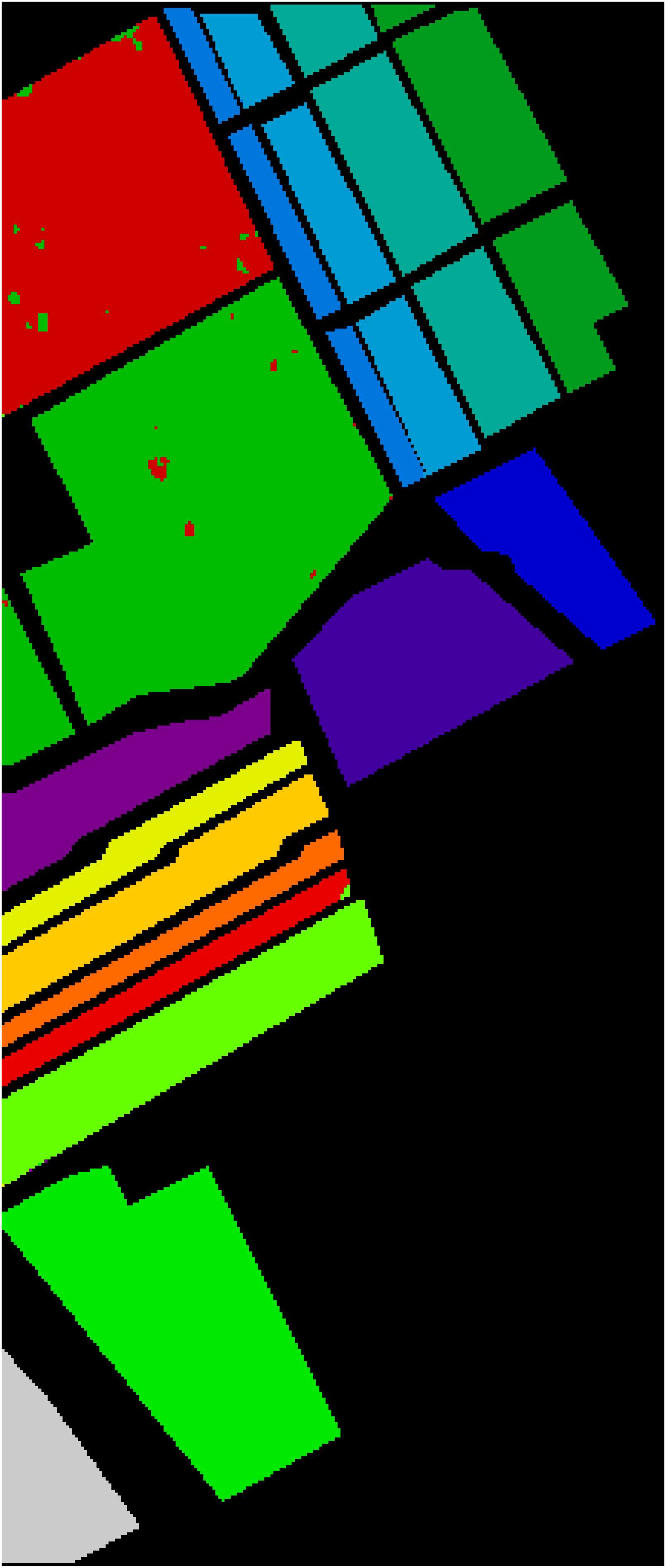}
		\centering
		\caption{3D}
		\label{Fig7E}
	\end{subfigure}
\caption{Salinas Dataset illustration in-terms of HSI Cube and Ground Truth (GT) label maps. The classification maps obtained using Hybrid AfNet with overall accuracy of 99.59 (\%), 2D Inception Net with overall accuracy of \textbf{99.76 (\%)} and 3D Inception Net with overall accuracy of 99.29 (\%).}
\label{Fig.7}
\end{figure}

The Salinas dataset was acquired using the AVIRIS sensor over Salinas Valley, California, and is characterized by high spatial resolution with 3.7 meters per pixel with 224 spectral bands. The area covered by each spectral band is $512 \times 217$ samples. As with the Indian Pines dataset, 20 water absorption bands which are [108-112], [154-167], and 224 were discarded. Salinas dataset is only available as sensor radiance data. It includes vegetables, bare soils, and vineyard fields. Salinas ground truths contain 16 mutually exclusive classes. The image cube and true ground truths label maps are shown in Figures \ref{Fig7A} - \ref{Fig7B} whereas Figures \ref{Fig7C} - \ref{Fig7D} show the classification performance in terms of classification maps (ground truth label maps) for three different models, i.e. Hybrid Attention Fused Network (AfNet), 3D Attention Inception Net and 2D Attention Inception Net. These maps clearly show that the proposed method performs better than 3D as well as 2D Attention Inception Networks. The higher accuracies are emphasized.

%%%%%%%%%%%%%%%%%%%%%%%%%%%%%%%%%%%%

%%%%%%%%%%%%%%%%%%%%%%%%%%%%%%%%%%%%
\subsection{Artefacts of Spatial Dimensions}
\label{sec4.2}

To process the HSI cube in any CNN, the Spatial dimensions is being considered an important component and have an important impact on classification results \cite{9307220}. This section experimentally illustrates the impact of spatial dimensions on classification results i.e., Overall (OA), Average Accuracy (AA), and Kappa ($\kappa$) accuracy irrespective of the processing time and the computational cost which gradually increases as the spatial dimensions increase. The OA, AA, and $\kappa$ accuracy is presented on all four experimental datasets to explore the impact of spatial dimensions on our proposed Hybrid AfNet. All the experimental settings and tuning parameters for this particular experiment remain the same except spatial dimensions in which we tested the models on several different sizes i.e., $9 \times 9 \times B$, $11\times 11 \times B$, $13 \times 13 \times B$, and $15 \times 15 \times B$.

%%%%%%%%%%%%%%%%%%%%%%%%%%%%%%%%%%%%
\begin{table}[!hbt]
    \centering
    \caption{Classification Performance and Training (Tr) and Testing (Te) Time (in seconds) on different spatial dimensions i.e., $9 \times 9 \times B$, $11\times 11 \times B$, $13 \times 13 \times B$, and $15 \times 15 \times B$. The training, validation, and test sets are randomly divided into three parts i.e. 15\%/15\%/70\% (Train/Validation/Test).}
    \begin{tabular}{cc||cccc} \\ \cline{1-6} 
    
        \multirow{2}{*}{\textbf{Dataset}} & \multirow{2}{*}{\textbf{Measure}} & \multicolumn{4}{c}{\textbf{Spatial Dimensions}} \\ \cline{3-6}
        
        & & $9 \times 9$ & $11\times 11$ & $13 \times 13$ & $15 \times 15$ \\ \cline{1-6}
        
        \multirow{5}{*}{IP} & \textbf{$\kappa$ (\%)} & 91.79 & 94.34 & 95.04 & 95.97 \\ \cline{2-6}
        & \textbf{OA (\%)} & 92.82 & 95.04 & 95.65 & 96.47 \\ \cline{2-6}
        & \textbf{AA (\%)} & 83.26 & 79.53 & 86.62 & 86.13 \\ \cline{2-6}
        & \textbf{Tr Time} & 83.92 & 83.88 & 143.23 & 263.93 \\ \cline{2-6}
        & \textbf{Te Time} & 1.60 & 1.83 & 2.62 & 3.39 \\ \hline 

        \multirow{5}{*}{BS} & \textbf{$\kappa$ (\%)} & 98.57 & 98.76 & 99.18 & 98.90 \\ \cline{2-6}
        & \textbf{OA (\%)} & 98.68 & 98.85 & 99.25 & 98.98 \\ \cline{2-6}
        & \textbf{AA (\%)} & 97.83 & 98.01 & 98.85 & 98.774 \\ \cline{2-6}
        & \textbf{Tr Time} & 22.42 & 42.14 & 83.95 & 144.14 \\ \cline{2-6}
        & \textbf{Te Time} & 0.94 & 0.91 & 1.30 & 1.62 \\ \hline
        
        \multirow{5}{*}{SA} & \textbf{$\kappa$ (\%)} & 99.54 & 99.72 & 99.94 & 99.85 \\ \cline{2-6}
        & \textbf{OA (\%)} & 99.59 & 99.74 & 99.94 & 99.86 \\ \cline{2-6}
        & \textbf{AA (\%)} & 99.76 & 99.85 & 99.96 & 99.84 \\ \cline{2-6}
        & \textbf{Tr Time} & 248.53 & 443.90 & 716.78 & 1043.87 \\ \cline{2-6}
        & \textbf{Te Time} & 10.77 & 10.90 & 21.23 & 21.40 \\ \hline

        \multirow{5}{*}{PU} & \textbf{$\kappa$ (\%)} & 99.27 & 99.57 & 99.79 & 99.47 \\ \cline{2-6}
        & \textbf{OA (\%)} & 99.45 & 99.67 & 99.84 & 99.60 \\ \cline{2-6}
        & \textbf{AA (\%)} & 98.92 & 99.38 & 99.68 & 99.28 \\ \cline{2-6}
        & \textbf{Tr Time} & 203.92 & 323.88 & 552.19 & 821.60 \\ \cline{2-6}
        & \textbf{Te Time} & 4.74 & 10.71 & 10.81 & 11.74 \\ \hline

    \end{tabular}
    \label{Tab.3}
\end{table}

\begin{figure}[!hbt]
    \centering
	\begin{subfigure}{0.11\textwidth}
		\includegraphics[width=0.99\textwidth]{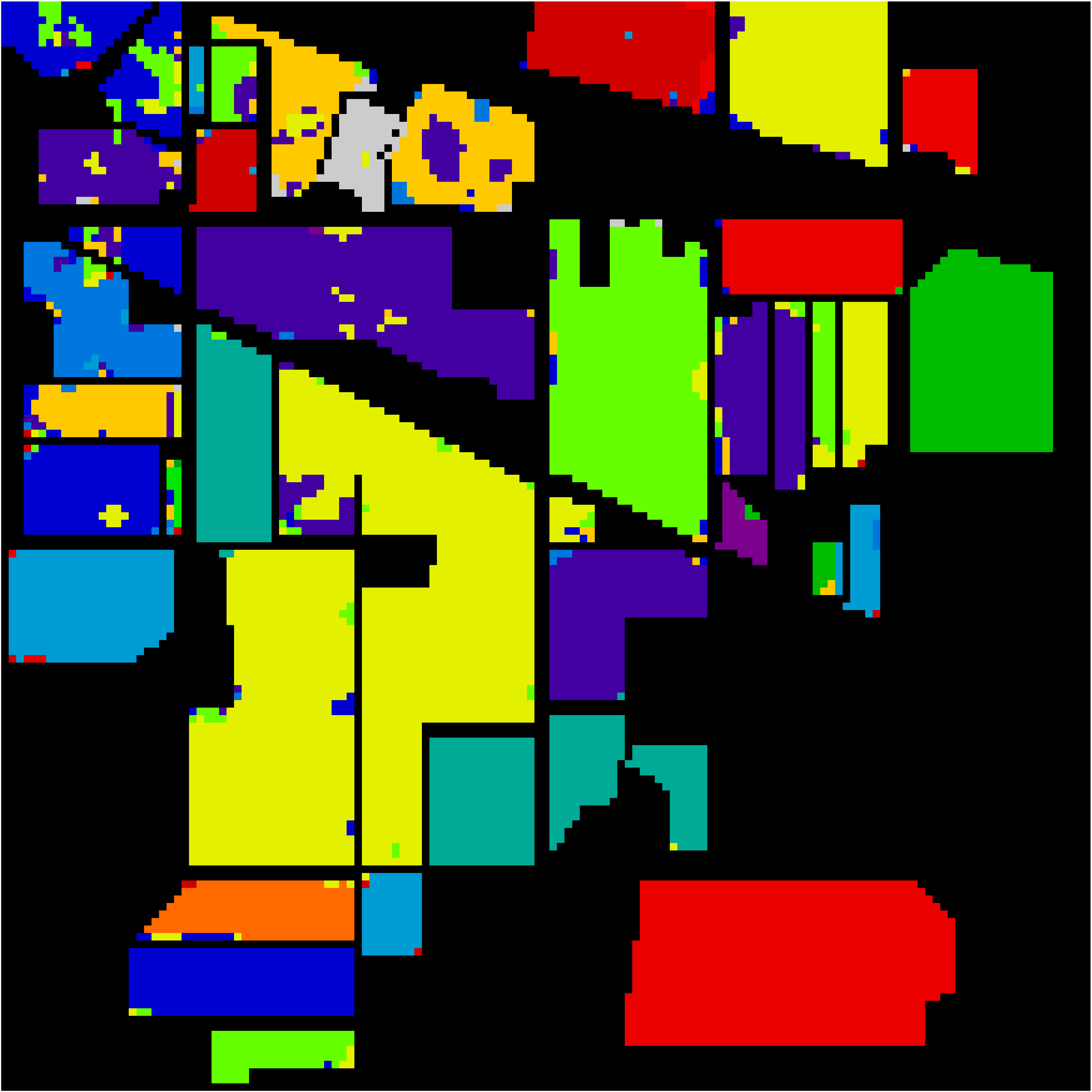}
		\caption{IP:$9\times 9$} 
		\label{Fig8A}
	\end{subfigure}
	\begin{subfigure}{0.11\textwidth}
		\includegraphics[width=0.99\textwidth]{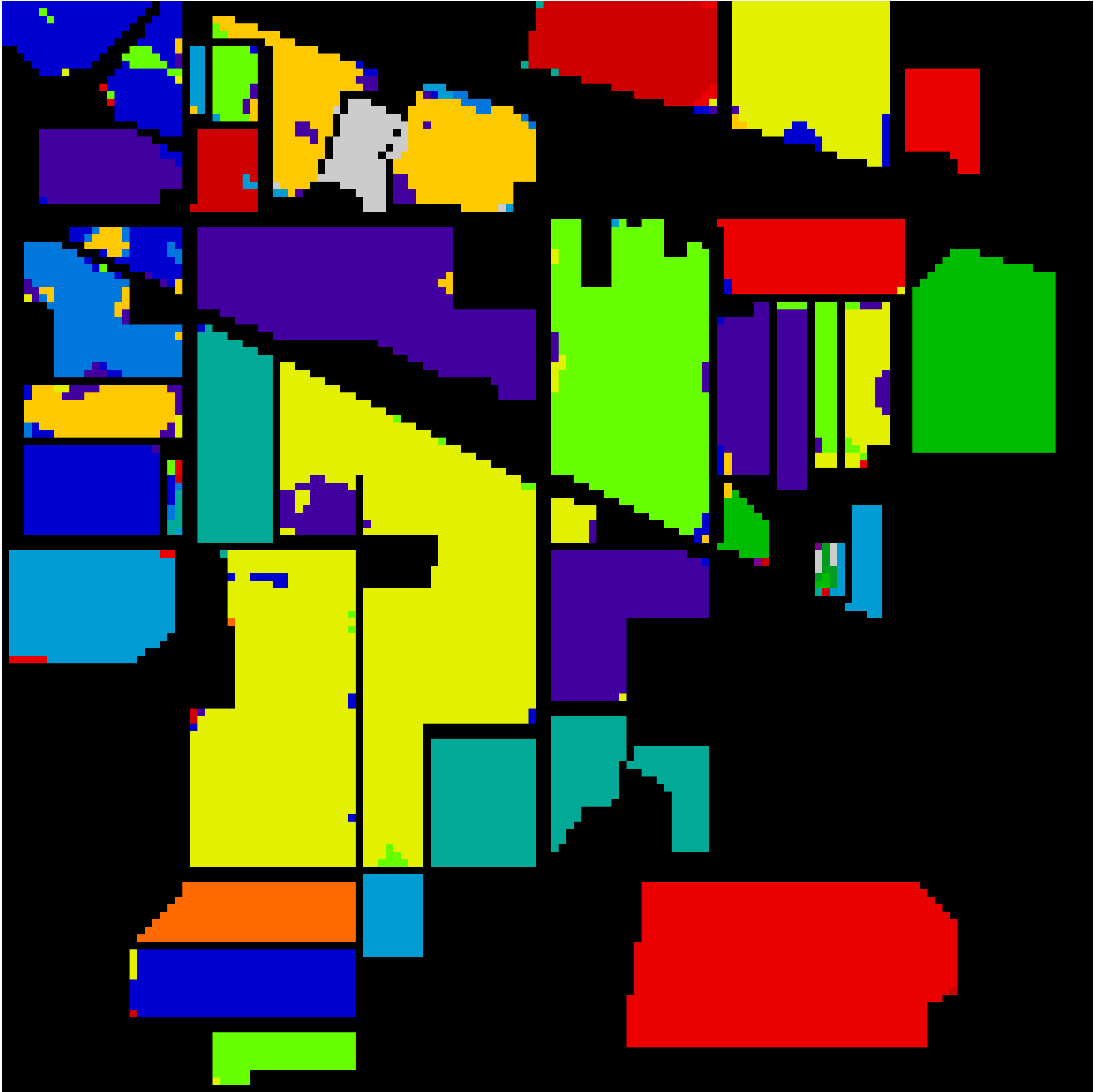}
		\caption{IP:$11\times 11$}
		\label{Fig8B}
	\end{subfigure}
	\begin{subfigure}{0.11\textwidth}
		\includegraphics[width=0.99\textwidth]{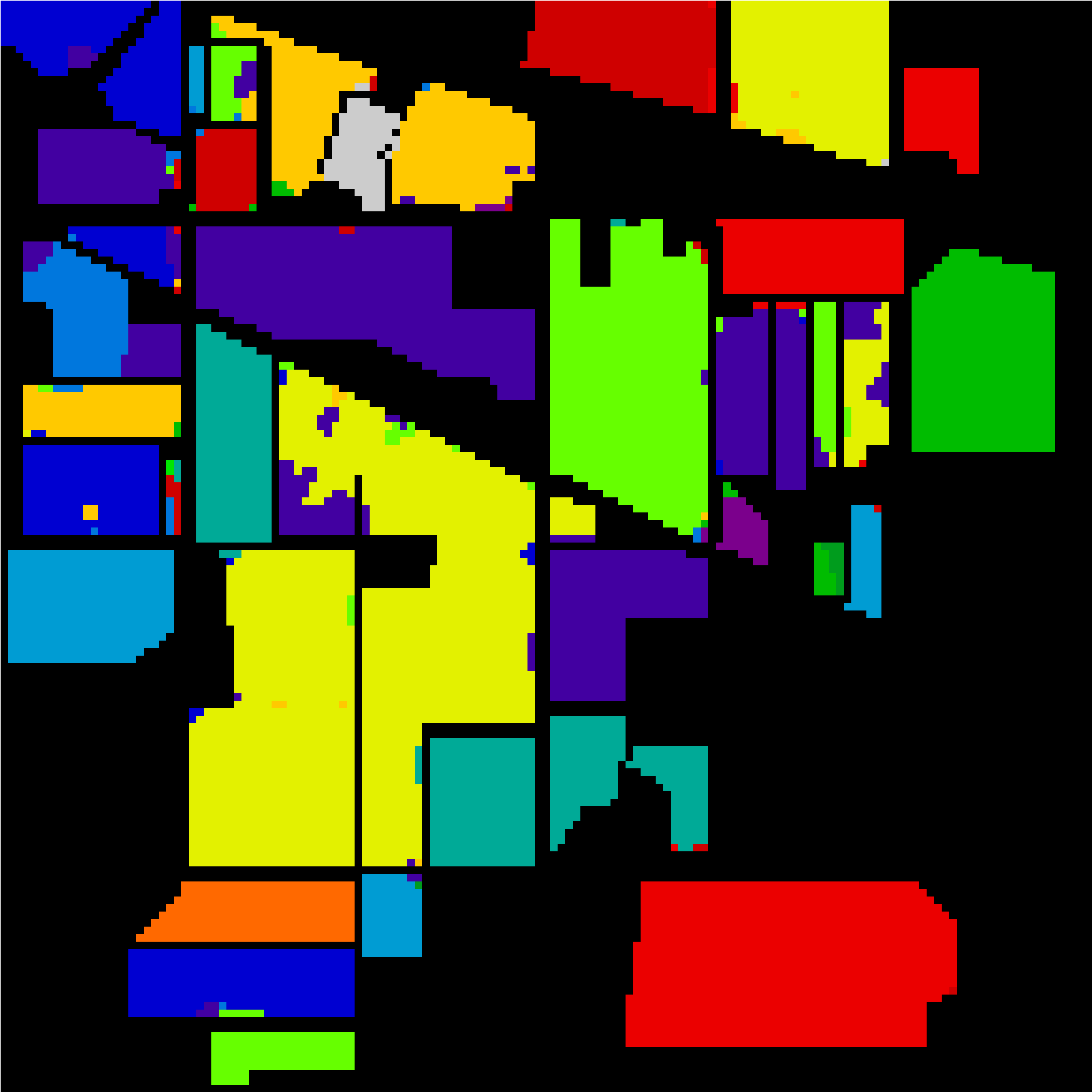}
		\centering
		\caption{IP:$13\times 13$} 
		\label{Fig8C}
	\end{subfigure}
	\begin{subfigure}{0.11\textwidth}
		\includegraphics[width=0.99\textwidth]{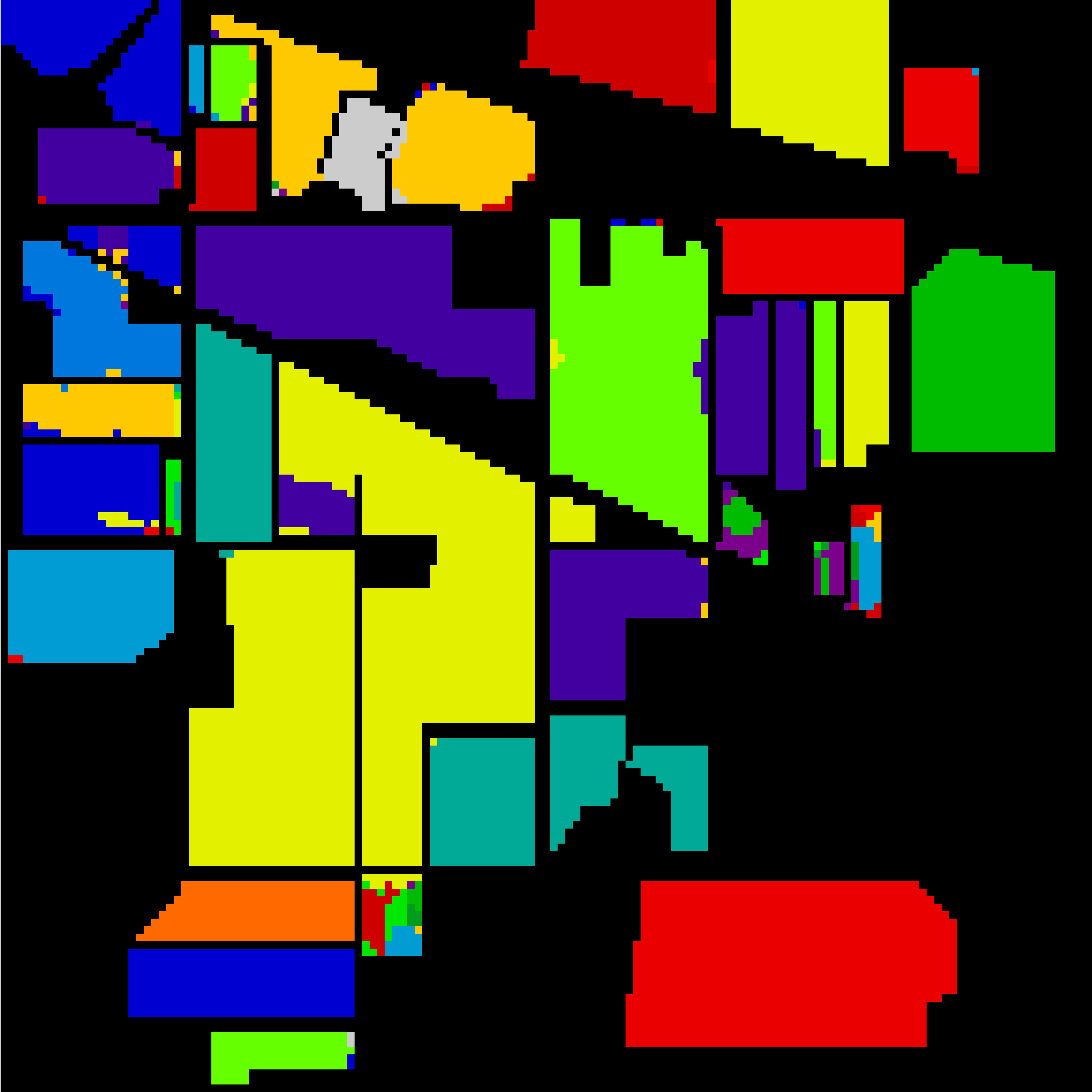}
		\centering
		\caption{IP:$15\times 15$} 
		\label{Fig8D}
	\end{subfigure} \\ \vspace{0.3cm}
    
    \begin{subfigure}{0.24\textwidth}
        \centering
		\includegraphics[angle=90,width=0.99\textwidth]{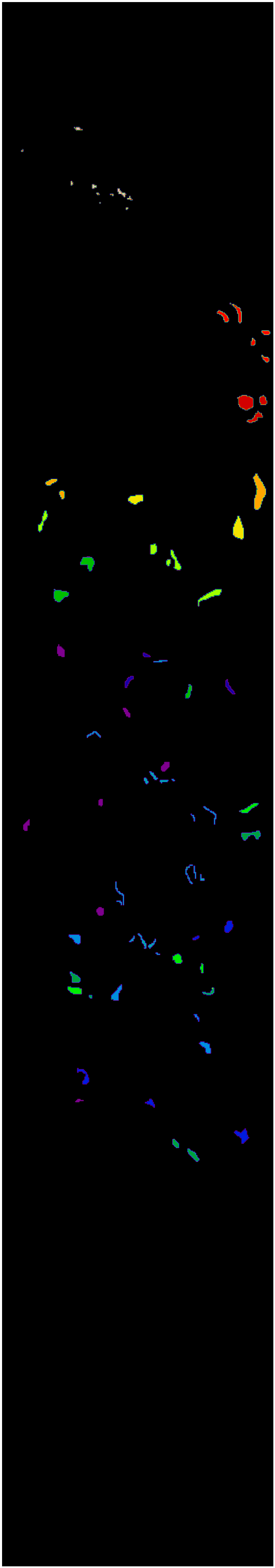}
		\caption{BS:$9\times 9$} 
		\label{Fig8E}
	\end{subfigure}
	\begin{subfigure}{0.24\textwidth}
	\centering
		\includegraphics[angle=90,width=0.99\textwidth]{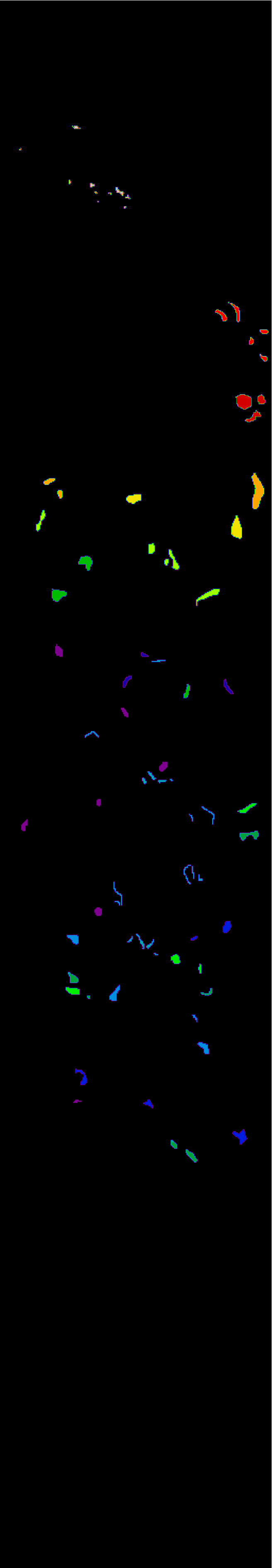}
		\caption{BS:$11\times 11$}
		\label{Fig8FB}
	\end{subfigure}
	\begin{subfigure}{0.24\textwidth}
	\centering
		\includegraphics[angle=90,width=0.99\textwidth]{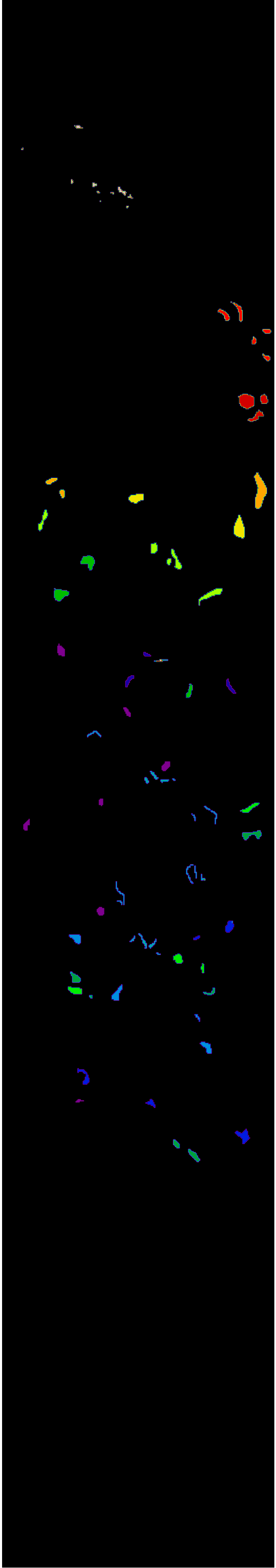}
		\caption{BS:$13\times 13$} 
		\label{Fig8G}
	\end{subfigure}
	\begin{subfigure}{0.24\textwidth}
	\centering
		\includegraphics[angle=90,width=0.99\textwidth]{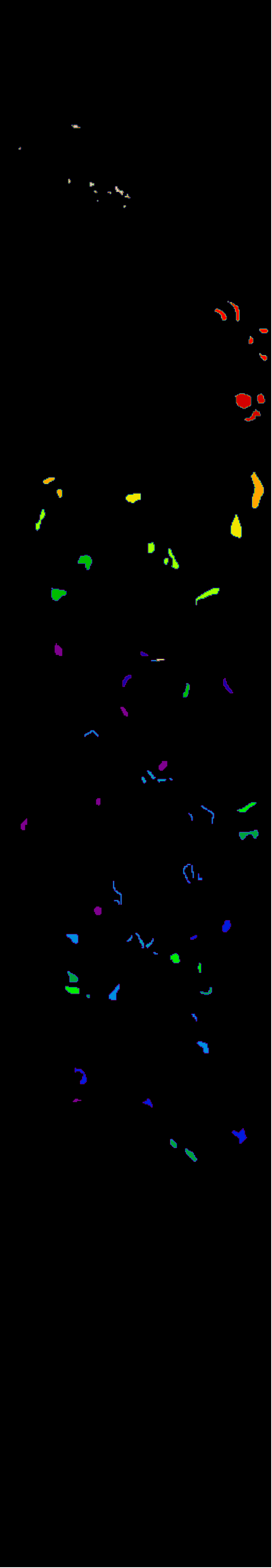}
		\caption{BS:$15\times 15$} 
		\label{Fig8H}
	\end{subfigure} \\ \vspace{0.3cm}

    \begin{subfigure}{0.11\textwidth}
        \centering
		\includegraphics[width=0.99\textwidth]{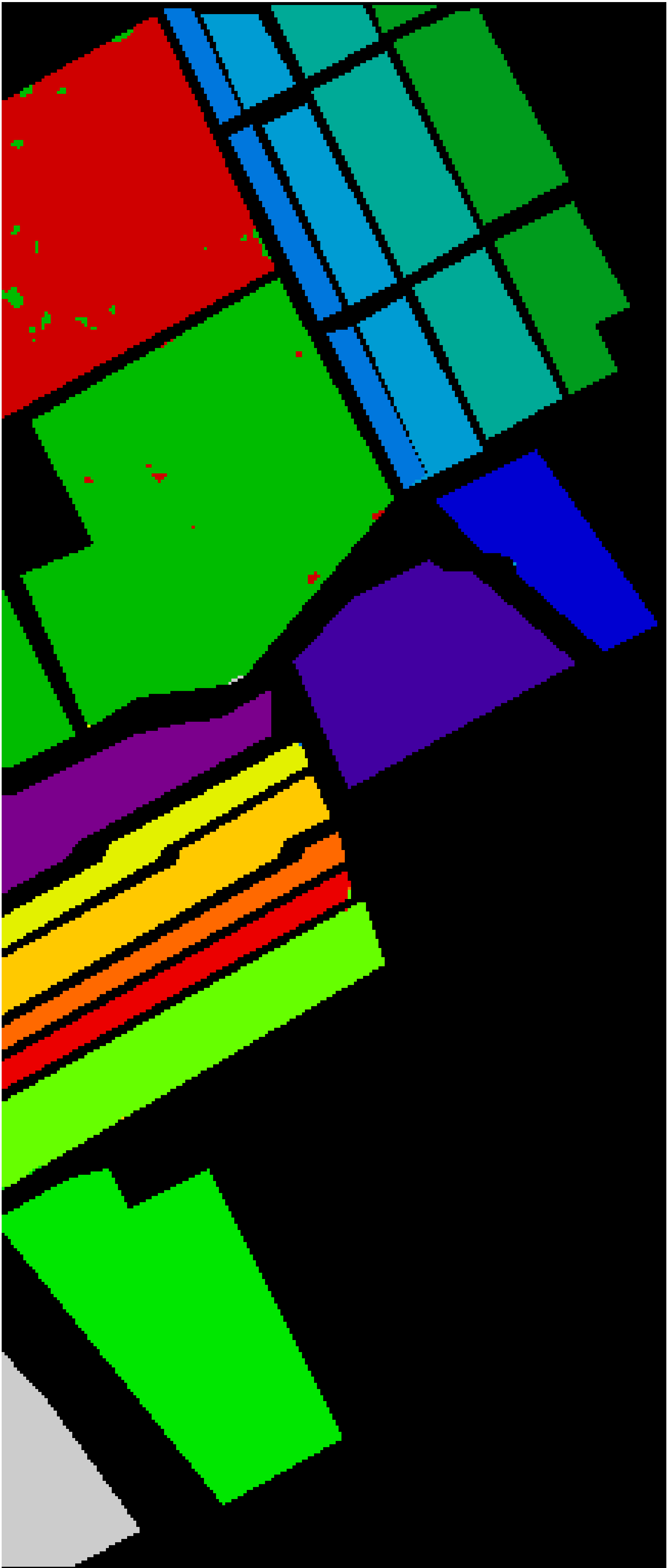}
		\caption{SA:$9\times 9$} 
		\label{Fig8I}
	\end{subfigure}
	\begin{subfigure}{0.11\textwidth}
		\includegraphics[width=0.99\textwidth]{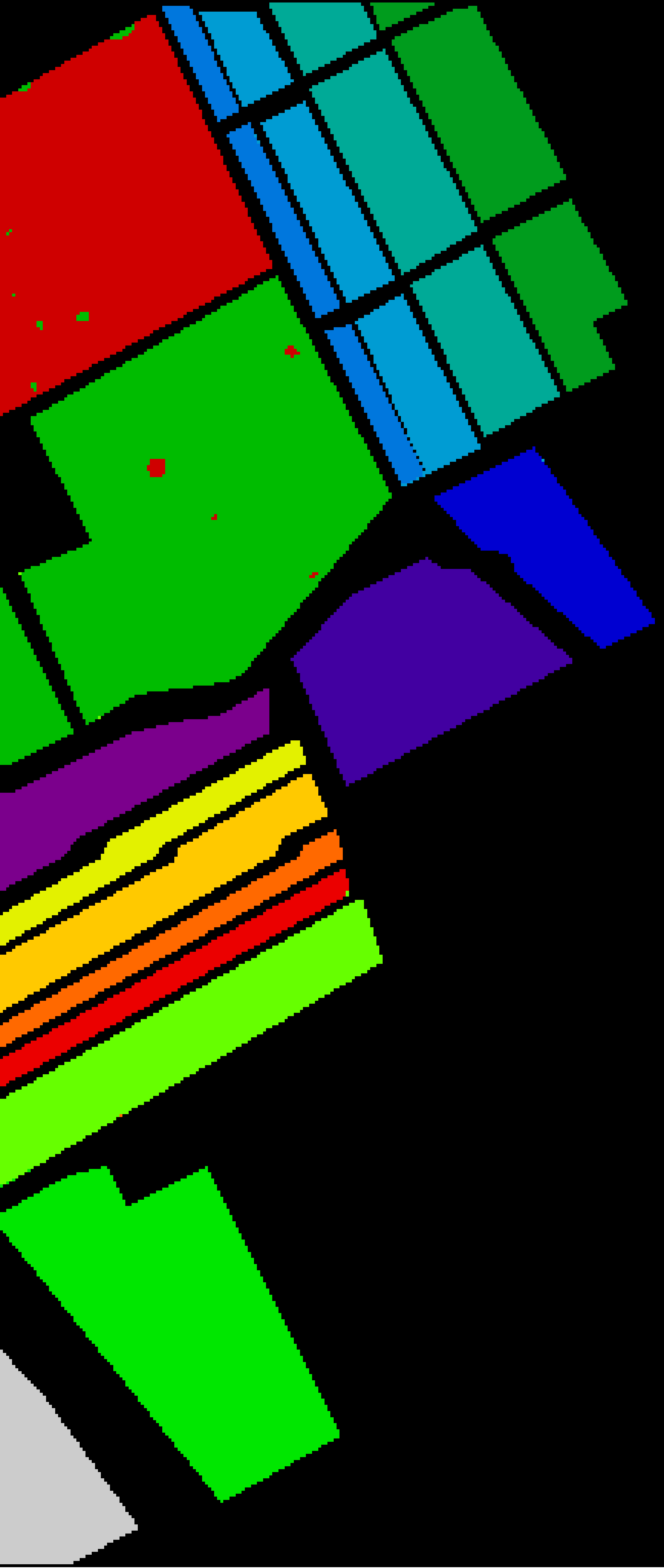}
		\caption{SA:$11\times 11$}
		\label{Fig8J}
	\end{subfigure}
	\begin{subfigure}{0.11\textwidth}
		\includegraphics[width=0.99\textwidth]{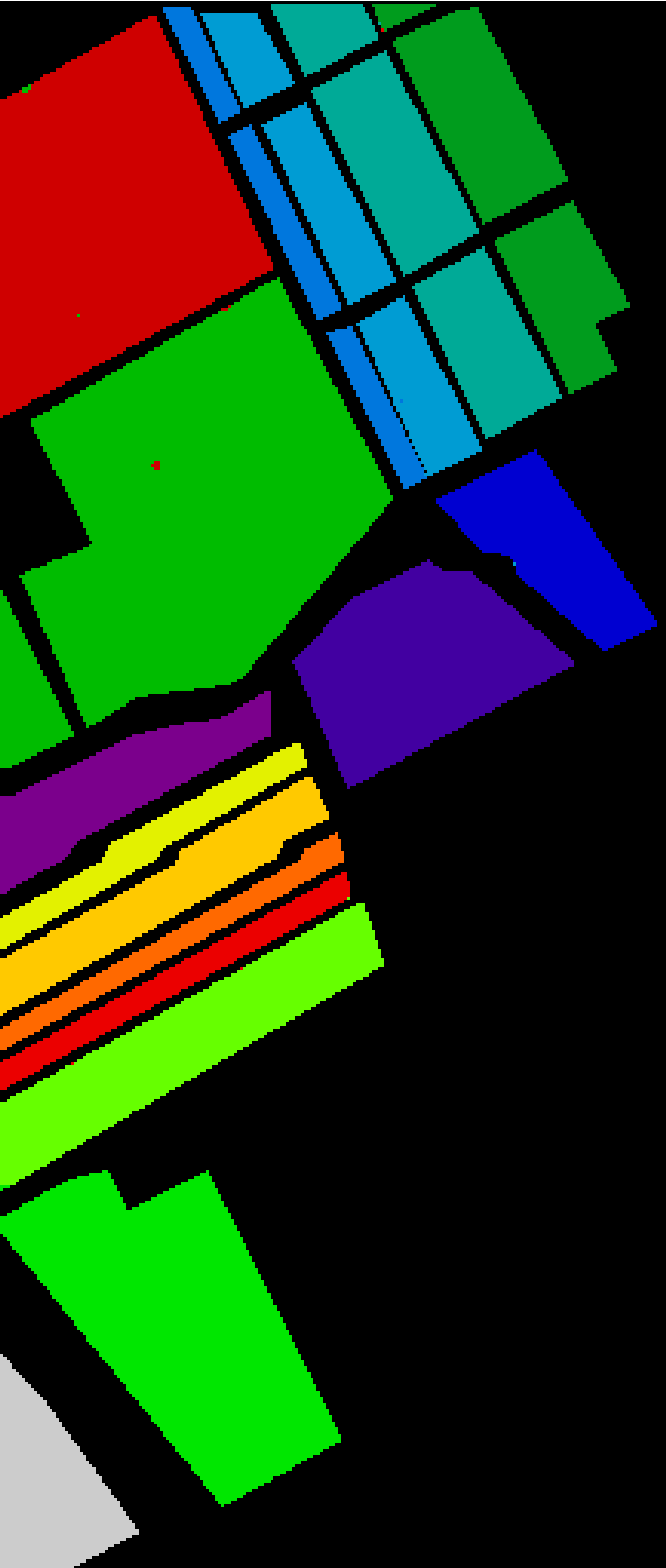}
		\centering
		\caption{SA:$13\times 13$}
		\label{Fig8K}
	\end{subfigure}
	\begin{subfigure}{0.11\textwidth}
		\includegraphics[width=0.99\textwidth]{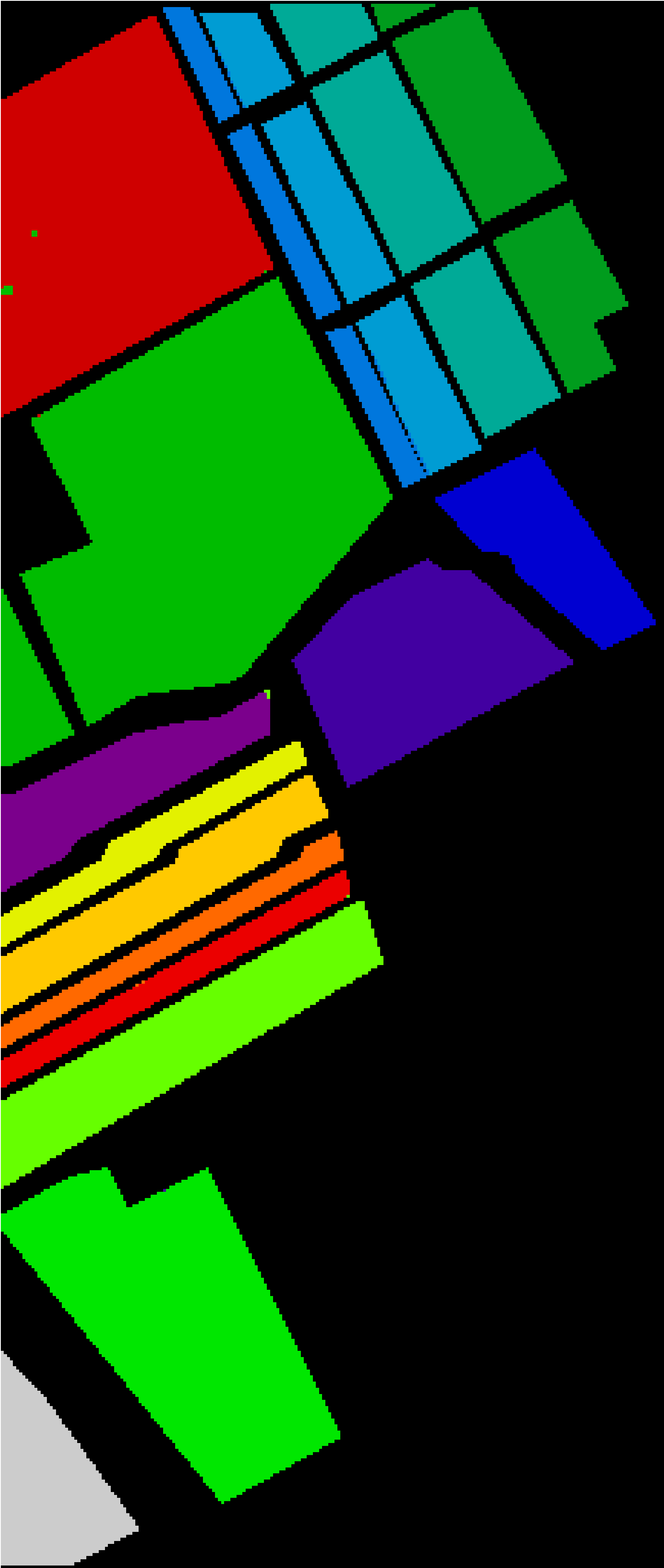}
		\centering
		\caption{SA:$15\times 15$} 
		\label{Fig8L}
	\end{subfigure} \\ \vspace{0.3cm}
	
	\begin{subfigure}{0.11\textwidth}
        \centering
		\includegraphics[width=0.99\textwidth]{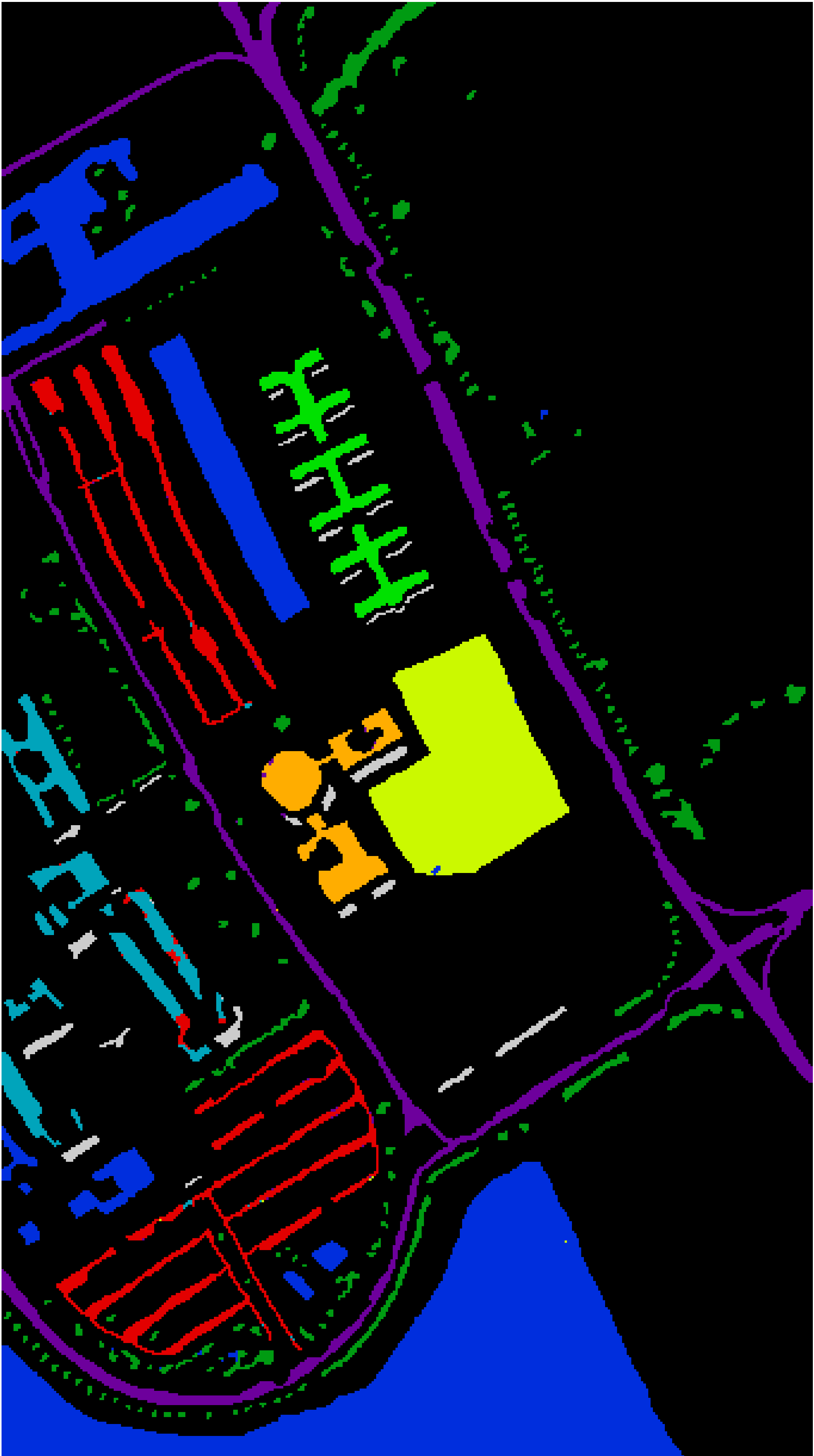}
		\caption{PU:$9\times 9$} 
		\label{Fig8M}
	\end{subfigure}
	\begin{subfigure}{0.11\textwidth}
		\includegraphics[width=0.99\textwidth]{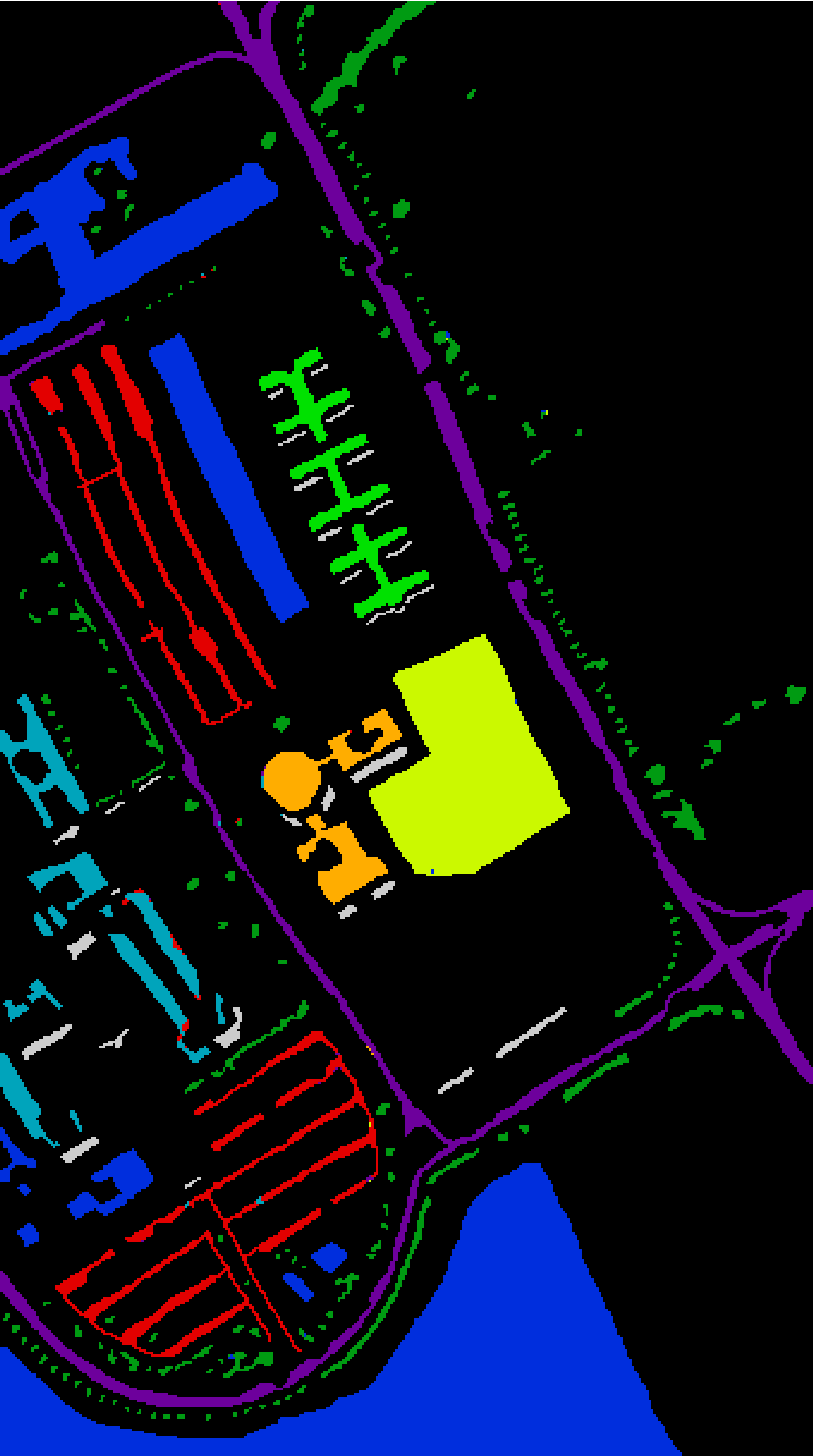}
		\caption{PU:$11\times 11$}
		\label{Fig8N}
	\end{subfigure}
	\begin{subfigure}{0.11\textwidth}
		\includegraphics[width=0.99\textwidth]{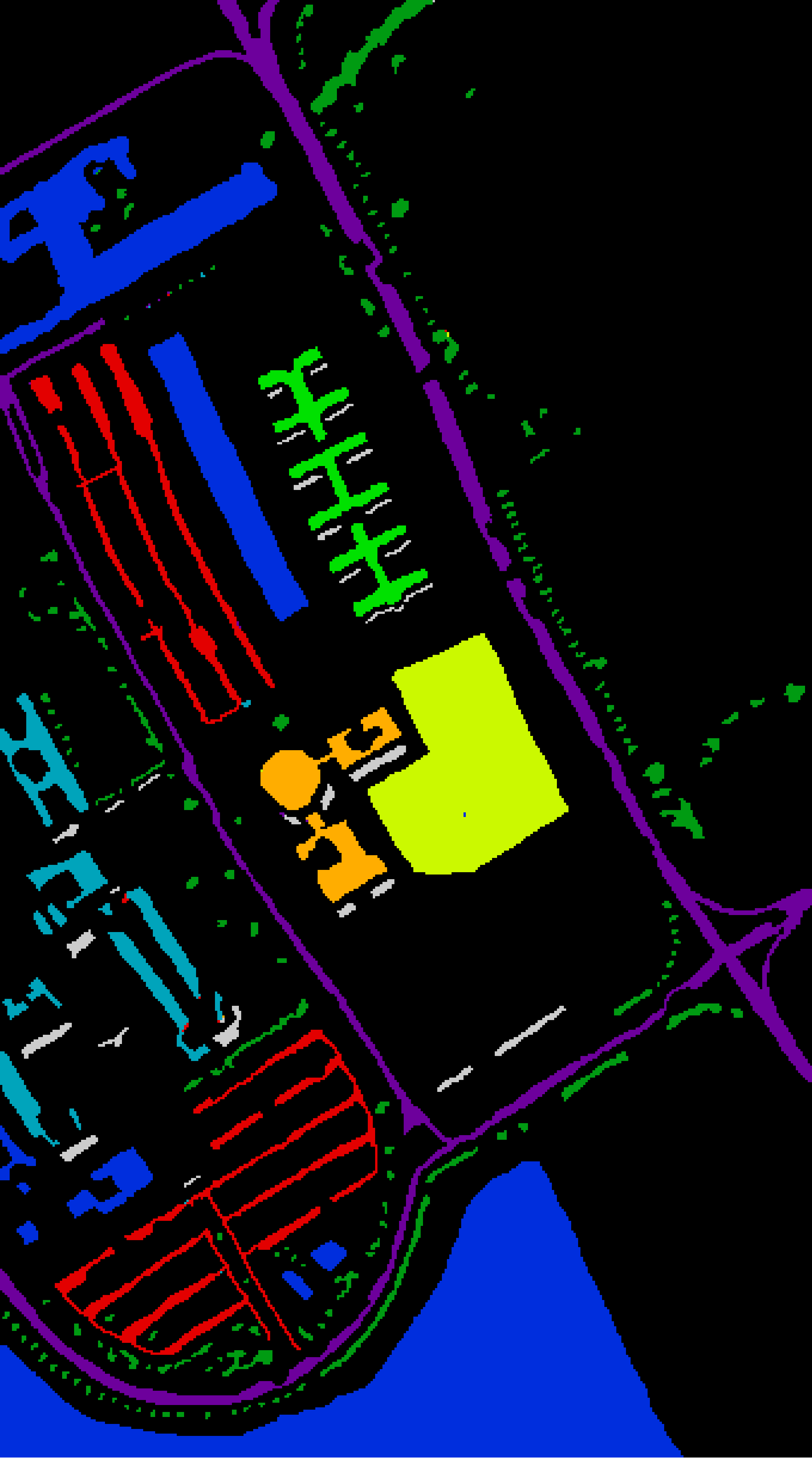}
		\centering
		\caption{PU:$13\times 13$}
		\label{Fig8O}
	\end{subfigure}
	\begin{subfigure}{0.11\textwidth}
		\includegraphics[width=0.99\textwidth]{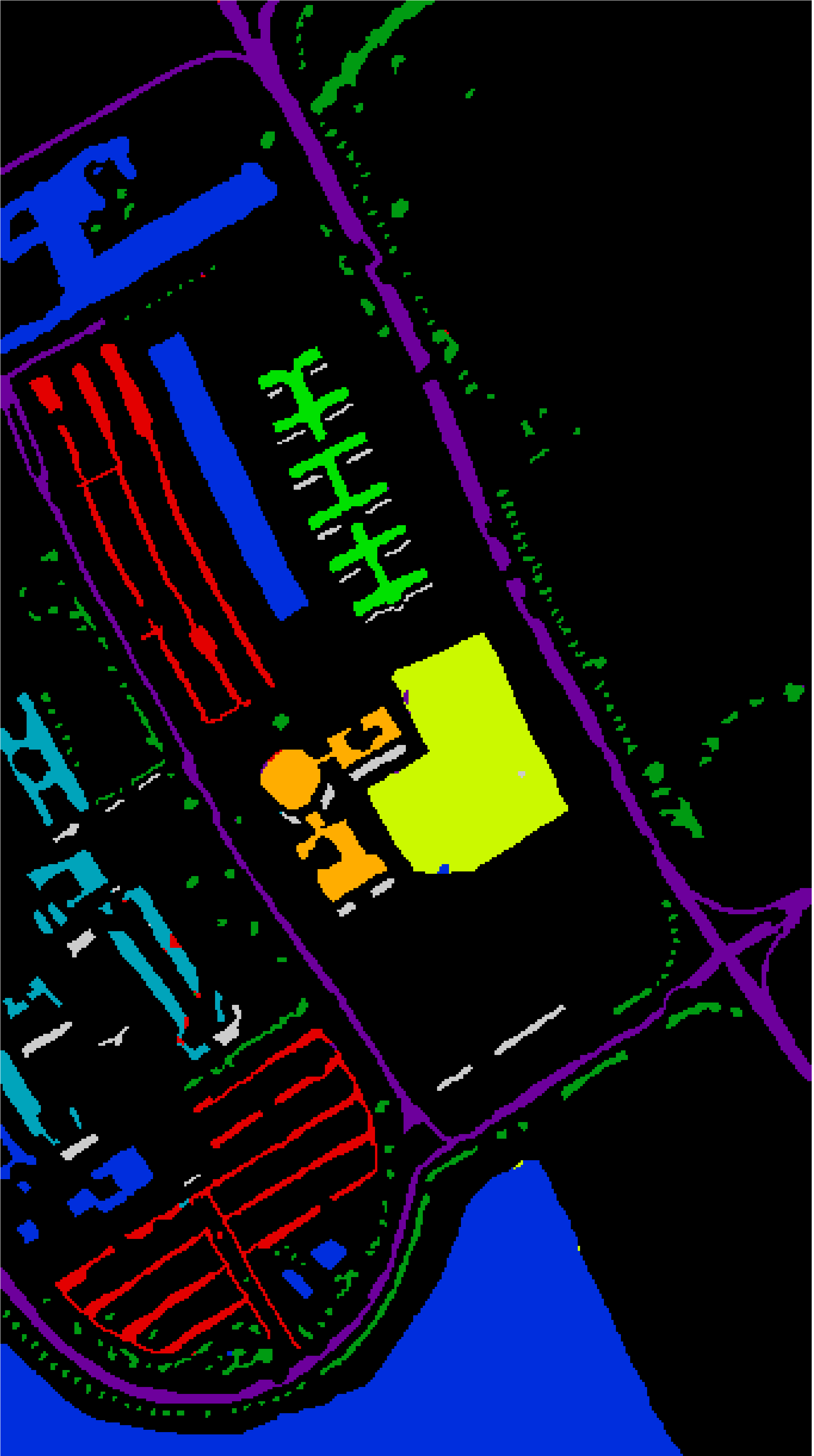}
		\centering
		\caption{PU:$15\times 15$} 
		\label{Fig8P}
	\end{subfigure}
\caption{Classification performance in terms of ground truth maps on different spatial dimensions.}
\label{Fig.8}
\end{figure}

%%%%%%%%%%%%%%%%%%%%%%%%%%%%%%%%%%%%
All these experimental results are presented in Figure \ref{Fig.8} and Table \ref{Tab.3} in which one can observe that the classification accuracy improves as the spatial size improves. The reason behind this trend is that “the larger spatial dimensions contains more samples”. However, this trend does not remain the same for all the spatial dimensions, as it may contain redundant samples, or by increasing the spatial dimension may contain interfering samples in a spatial patch or may contain the overlapping regions which bring nothing new to the classifier but just confuse the classifier and deteriorate the classification performance with redundant samples. Thus, in a nutshell, an appropriate size of spatial dimension with respect to the characteristics of the data is quite important to attain reliable accuracy.

%%%%%%%%%%%%%%%%%%%%%%%%%%%%%%%%%%%%

%%%%%%%%%%%%%%%%%%%%%%%%%%%%%%%%%%%%
\subsection{Artefacts of Training Samples}
\label{sec4.3}

CNN's have been extensively utilized for HSIC however, deep CNN requires a large number of annotated training samples for appropriate learning. However, the collection of annotated samples for HSI is expensive and critical which demands human experts or exploration of real-time scenarios. Limited availability of annotated training samples hinders HSIC performance. Thus, an appropriate size of annotated training samples is an important factor for HSIC performance. This section provides the performance evaluation in terms of OA, AA, and $\kappa$ accuracy for different percentages of training samples. These percentages include 5/5/90\% (Train/Validation/Test), 7/7/86\%, 10/10/80\%, 12/12/76\%, and 15/15/70\% respectively. We intentionally did not use below 5\% training samples as several classes of different datasets do not have enough training samples to include, for instance, Oats class of IP dataset which only have 20 samples in total, thus selecting 1-4\% of training samples for this class would only include 1 sample from this class which is not enough to train the model. There is another option to avoid such limitation is to select the number of training samples rather using the percentage. However, this process will lead to another issue which is known as the "Class Imbalance" issue, which is not the problem under study. This may be considered as a potential future research direction.

%%%%%%%%%%%%%%%%%%%%%%%%%%%%%%%%%%%%
\begin{table}[!hbt]
    \centering
    \caption{Classification performance ($\kappa$, OA, and AA in percentage) on different percentages of training samples along with the processing time (in seconds) for both Training (Tr) and Testing (Te) process.}
    \resizebox{\columnwidth}{!}{
    \begin{tabular}{cc||ccccc} \\ \cline{1-7}
        \multirow{2}{*}{\textbf{Dataset}} & \multirow{2}{*}{\textbf{Measure}} & \multicolumn{5}{c}{\textbf{Percentage of Training Samples}} \\ \cline{3-7}
        & & 5\% & 7\% & 10\% & 12\%  & 15\%\\ \hline 
        
        \multirow{5}{*}{\textbf{IP}} & \textbf{$\kappa$(\%)} & 77.41 & 84.32 & 84.06 & 90.15 & 91.79 \\ \cline{3-7}
        & \textbf{OA(\%)} & 80.34 & 86.26 & 86.04 & 91.37 & 92.82 \\ \cline{3-7}
        & \textbf{AA(\%)} & 65.78 & 74.08 & 80.33 & 88.48 & 83.26 \\ \cline{3-7}
        & \textbf{Tr Time} & 22.41 & 42.88 & 33.44 & 42.88 & 83.93 \\ \cline{3-7}
        & \textbf{Te Time} & 1.95 & 1.87 & 2.89 & 2.91 & 1.60 \\ \hline
        
        \multirow{5}{*}{\textbf{BS}} & \textbf{$\kappa$(\%)} & 95.25 & 96.97 & 98.16 & 98.06 & 98.57 \\ \cline{3-7}
        & \textbf{OA(\%)} & 95.62 & 97.20 & 98.30 & 98.21 & 98.68 \\ \cline{3-7}
        & \textbf{AA(\%)} & 96.10 & 96.62 & 98.32 & 97.15 & 97.83 \\ \cline{3-7}
        & \textbf{Tr Time} & 16.69 & 22.42 & 20.26 & 22.47 & 22.42 \\ \cline{3-7}
        & \textbf{Te Time} & 1.07 & 1.08 & 0.90 & 0.91 & 0.94 \\ \hline 

        \multirow{5}{*}{\textbf{SA}} & \textbf{$\kappa$(\%)} & 98.39 & 97.72 & 99.19 & 97.79 & 99.54 \\ \cline{3-7}
        & \textbf{OA(\%)} & 98.56 & 97.95 & 99.27 & 98.01 & 99.59 \\ \cline{3-7}
        & \textbf{AA(\%)} & 99.16 & 98.07 & 99.56 & 98.25 & 99.76 \\ \cline{3-7}
        & \textbf{Tr Time} & 83.81 & 2243.88 & 143.90 & 208.87 & 248.53 \\ \cline{3-7}
        & \textbf{Te Time} & 7.33 & 82.28 & 10.77 & 10.82 & 10.77 \\ \hline 

        \multirow{5}{*}{\textbf{PU}} & \textbf{$\kappa$(\%)} & 98.14 & 98.36 & 99.26 & 98.26 & 99.27 \\ \cline{3-7}
        & \textbf{OA(\%)} & 98.59 & 98.76 & 99.44 & 98.68 & 99.45 \\ \cline{3-7}
        & \textbf{AA(\%)} & 97.39 & 97.90 & 99.05 & 98.19 & 98.92 \\ \cline{3-7}
        & \textbf{Tr Time} & 1164.25 & 1583.49 & 2243.48 & 166.92 & 203.92 \\ \cline{3-7}
        & \textbf{Te Time} & 57.44 & 82.54 & 49.62 & 5.58 & 4.74 \\ \hline
    \end{tabular}}
    \label{Tab.4}
\end{table}

\begin{figure}[!hbt]
    \centering
	\begin{subfigure}{0.09\textwidth}
		\includegraphics[width=0.99\textwidth]{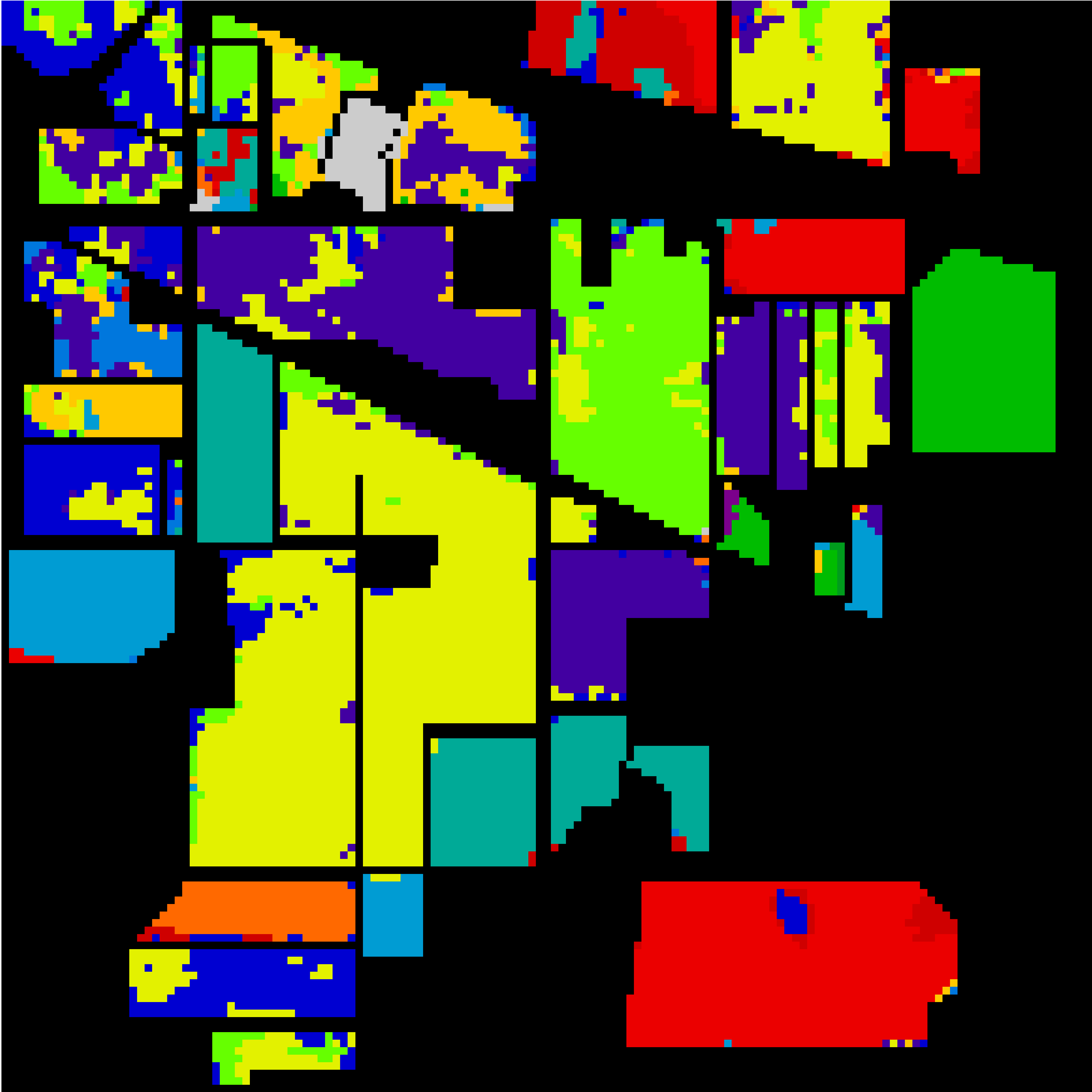}
		\caption{IP:5\%} 
		\label{Fig9A}
	\end{subfigure}
	\begin{subfigure}{0.09\textwidth}
		\includegraphics[width=0.99\textwidth]{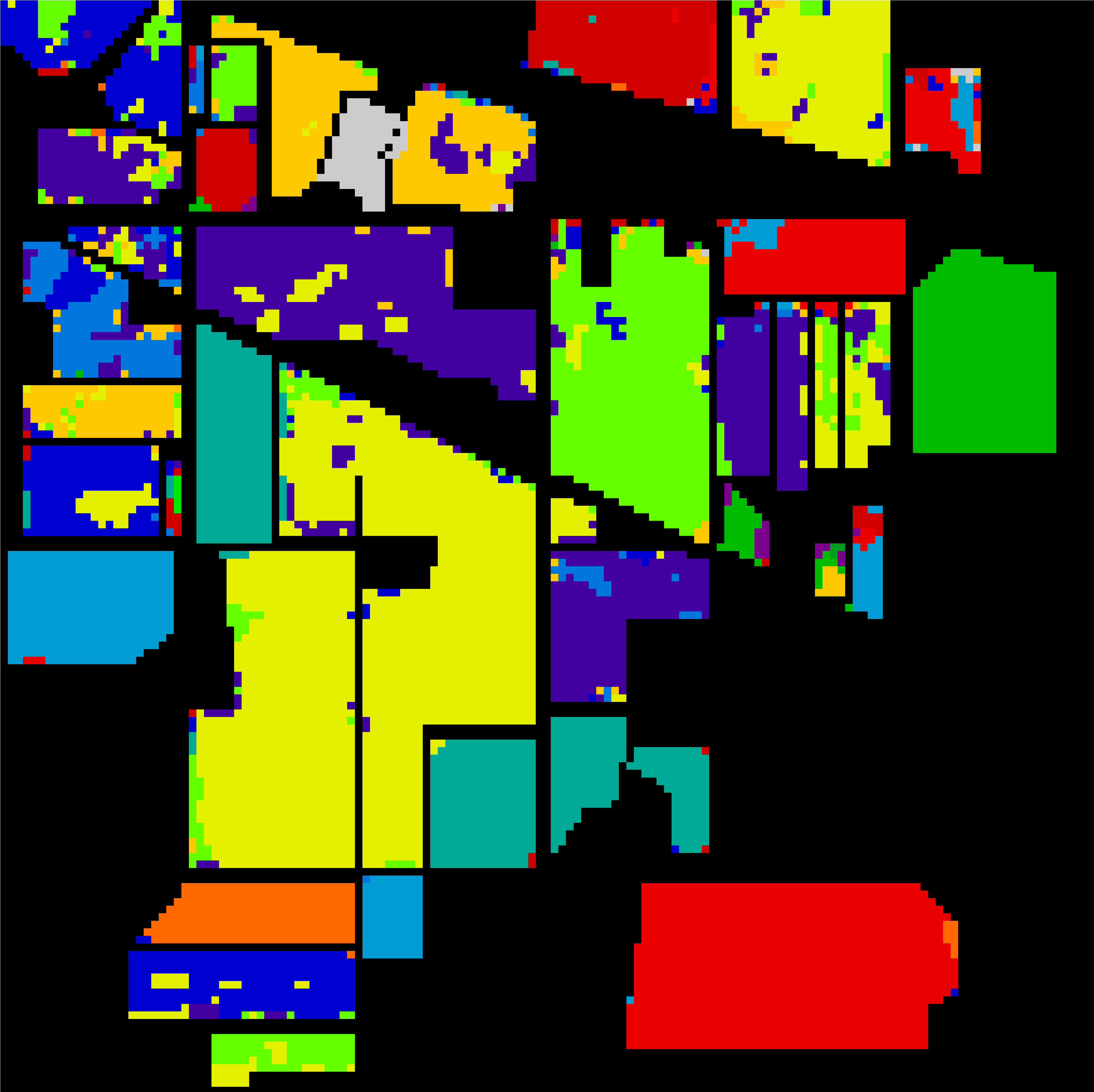}
		\caption{IP:7\%}
		\label{Fig9B}
	\end{subfigure}
	\begin{subfigure}{0.09\textwidth}
		\includegraphics[width=0.99\textwidth]{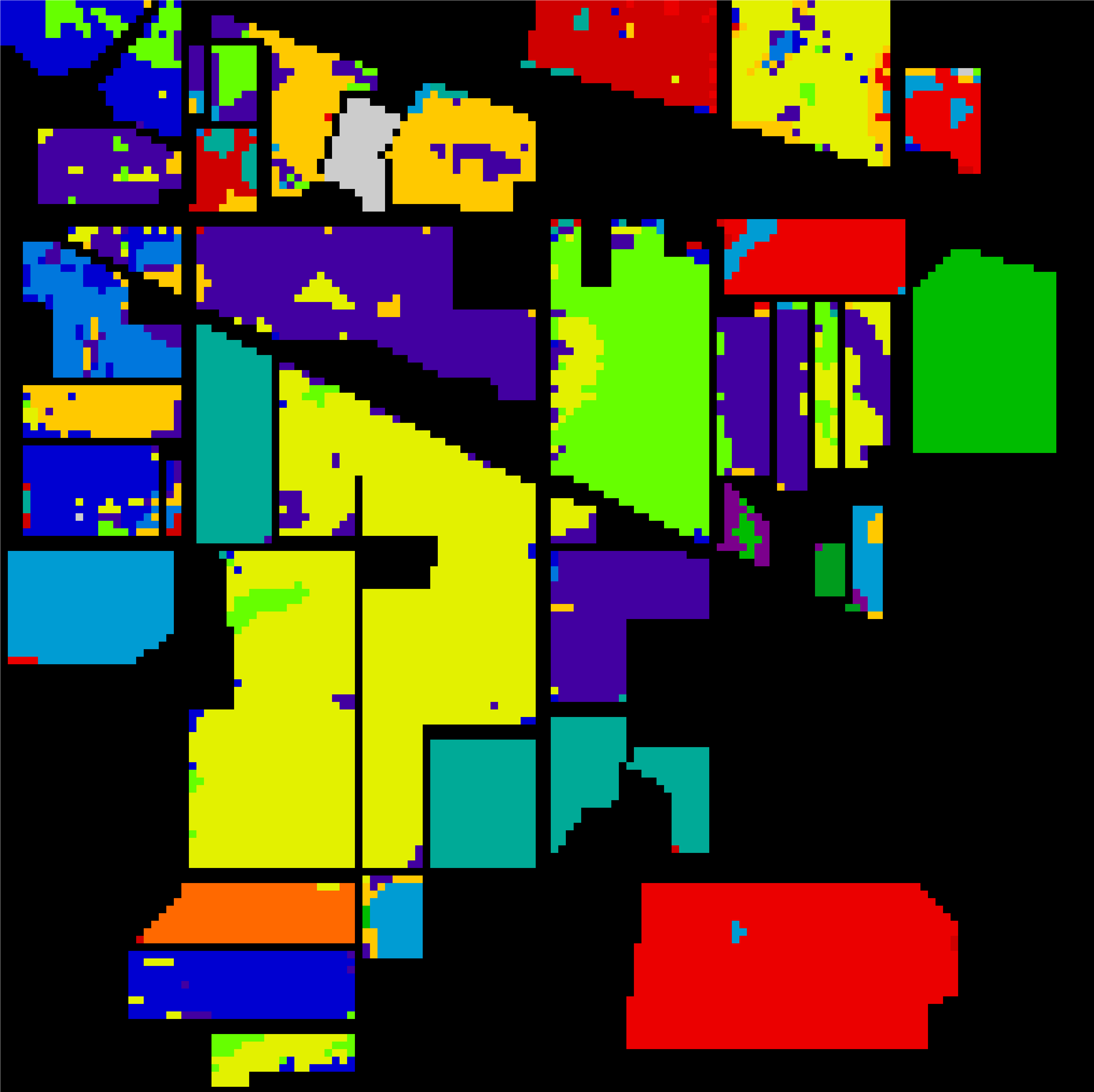}
		\centering
		\caption{IP:10\%} 
		\label{Fig9C}
	\end{subfigure}
	\begin{subfigure}{0.09\textwidth}
		\includegraphics[width=0.99\textwidth]{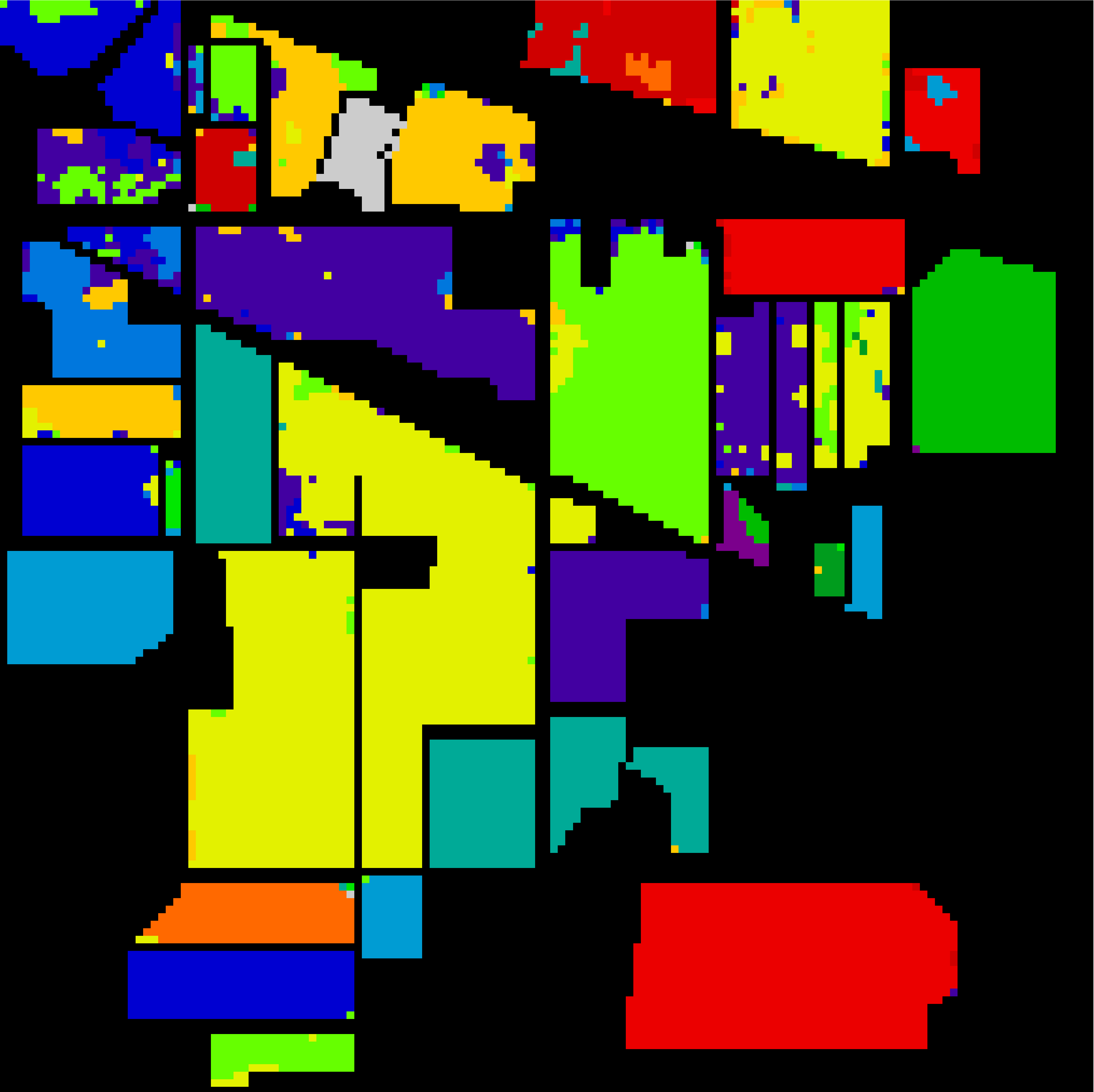}
		\centering
		\caption{IP:12\%} 
		\label{Fig9D}
	\end{subfigure} 
	\begin{subfigure}{0.09\textwidth}
		\includegraphics[width=0.99\textwidth]{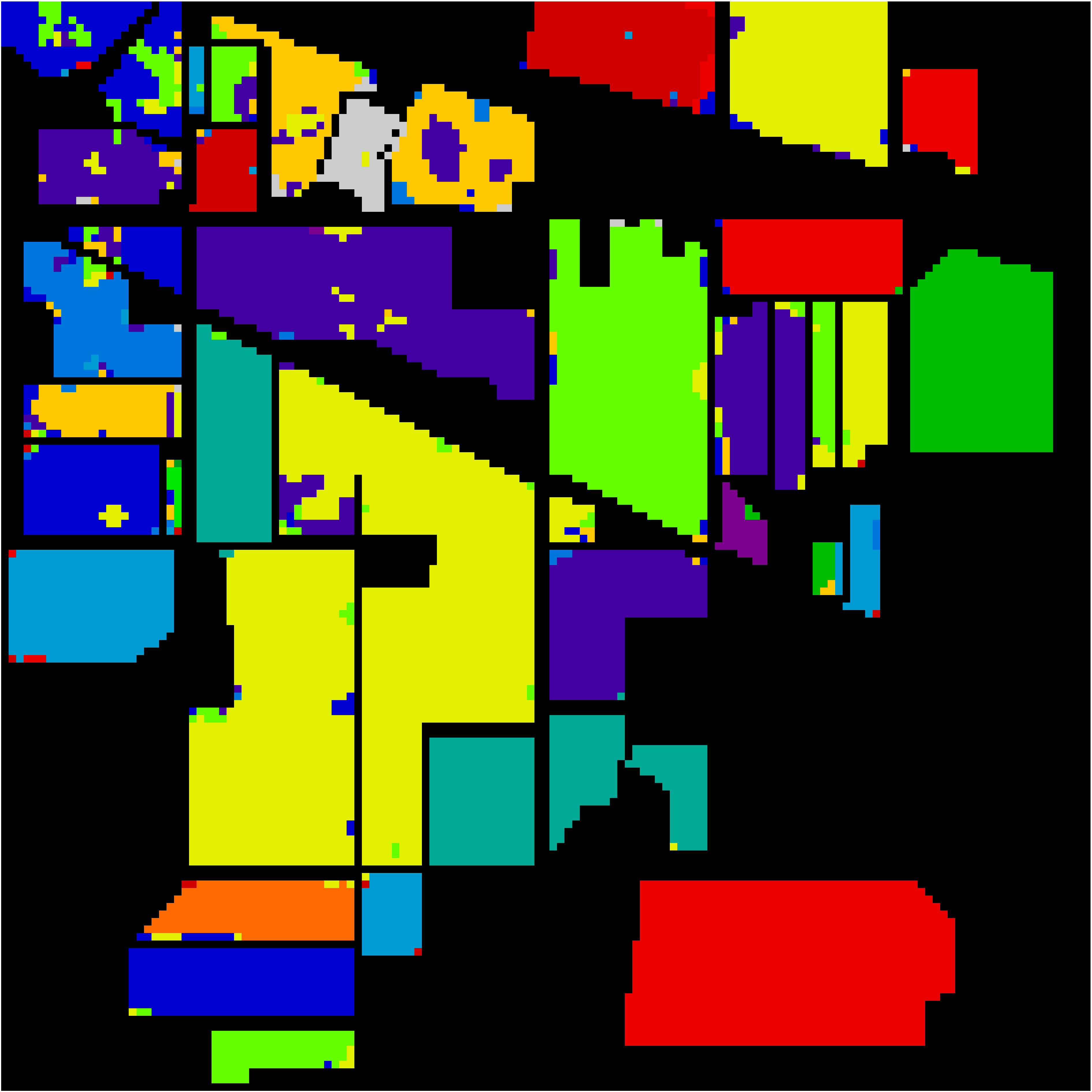}
		\centering
		\caption{IP:15\%} 
		\label{Fig9E}
	\end{subfigure} \\ \vspace{0.3cm}
	
    \begin{subfigure}{0.24\textwidth}
		\includegraphics[angle=90,width=0.99\textwidth]{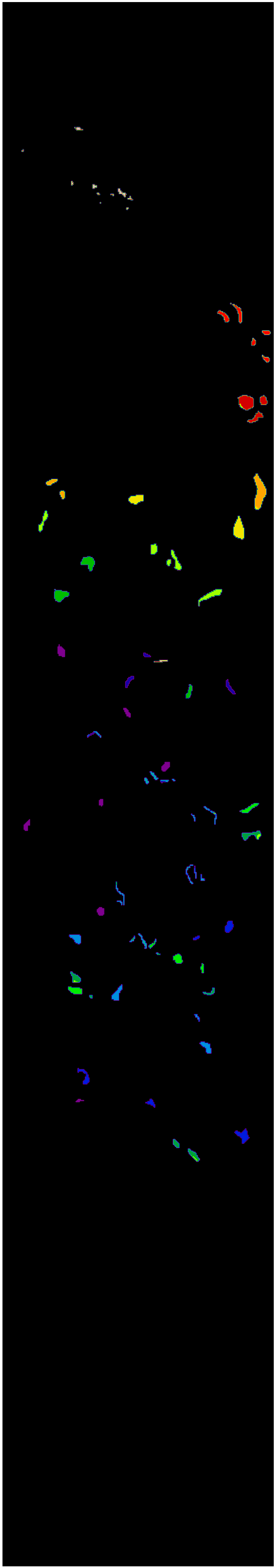}
		\caption{BS:5\%} 
		\label{Fig9A}
	\end{subfigure}
	\begin{subfigure}{0.24\textwidth}
		\includegraphics[angle=90,width=0.99\textwidth]{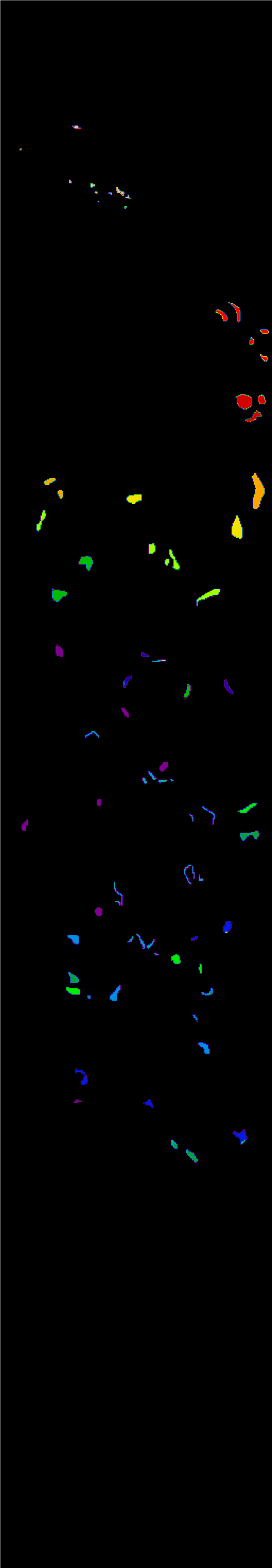}
		\caption{BS:7\%}
		\label{Fig9B}
	\end{subfigure}
	\begin{subfigure}{0.24\textwidth}
		\includegraphics[angle=90,width=0.99\textwidth]{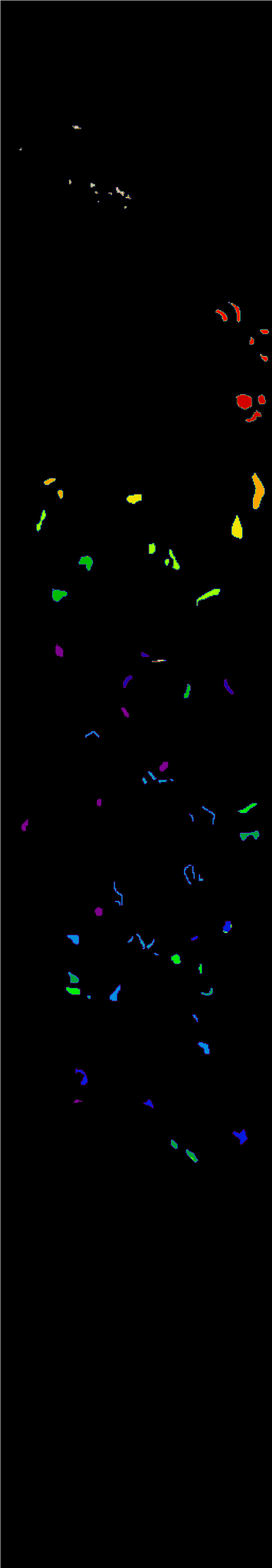}
		\centering
		\caption{BS:10\%} 
		\label{Fig9C}
	\end{subfigure}
	\begin{subfigure}{0.24\textwidth}
		\includegraphics[angle=90,width=0.99\textwidth]{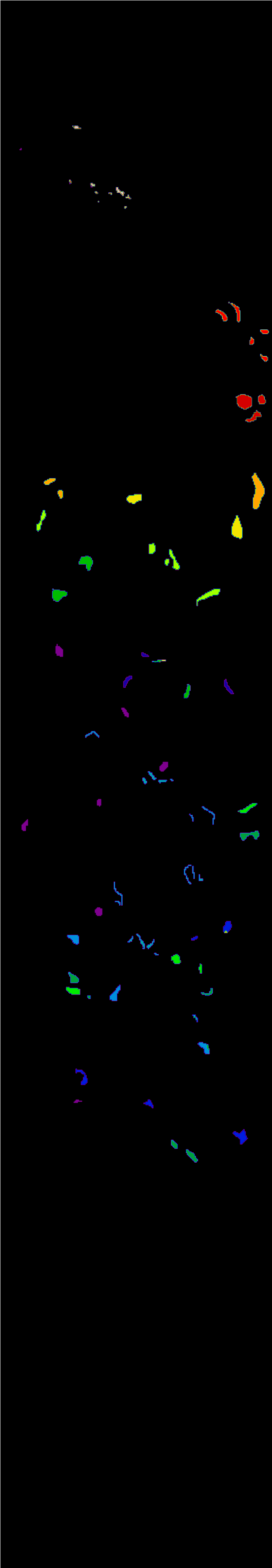}
		\centering
		\caption{BS:12\%} 
		\label{Fig9D}
	\end{subfigure} 
	\begin{subfigure}{0.24\textwidth}
		\includegraphics[angle=90,width=0.99\textwidth]{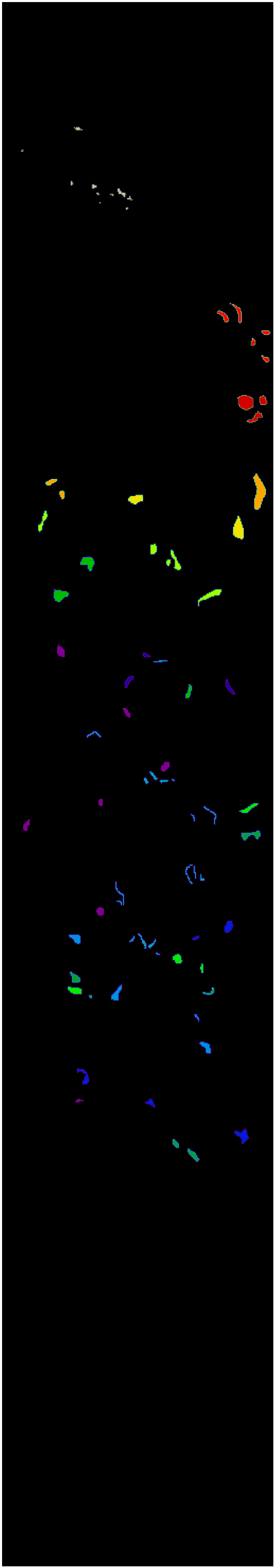}
		\centering
		\caption{BS:15\%} 
		\label{Fig9E}
	\end{subfigure} \\ \vspace{0.3cm}
	
    \begin{subfigure}{0.09\textwidth}
		\includegraphics[width=0.99\textwidth]{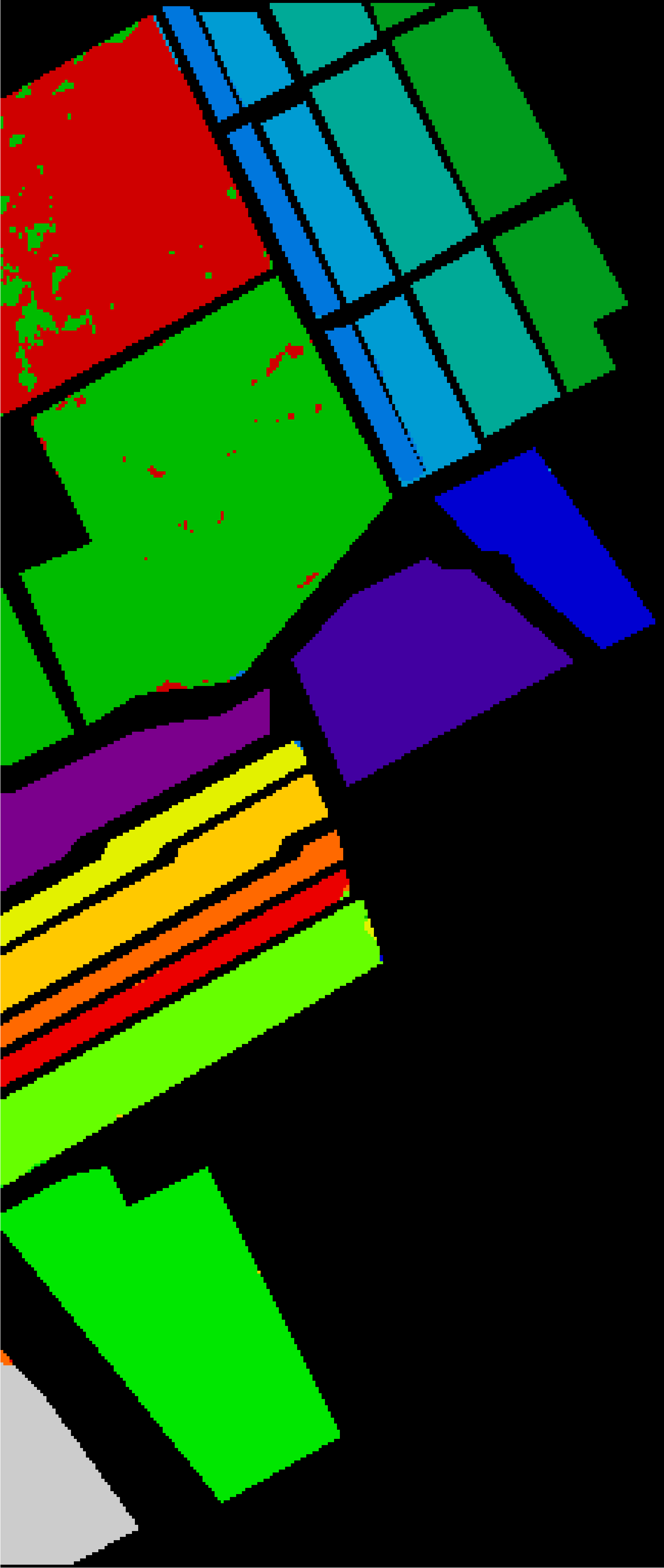}
		\caption{SA:5\%} 
		\label{Fig9A}
	\end{subfigure}
	\begin{subfigure}{0.09\textwidth}
		\includegraphics[width=0.99\textwidth]{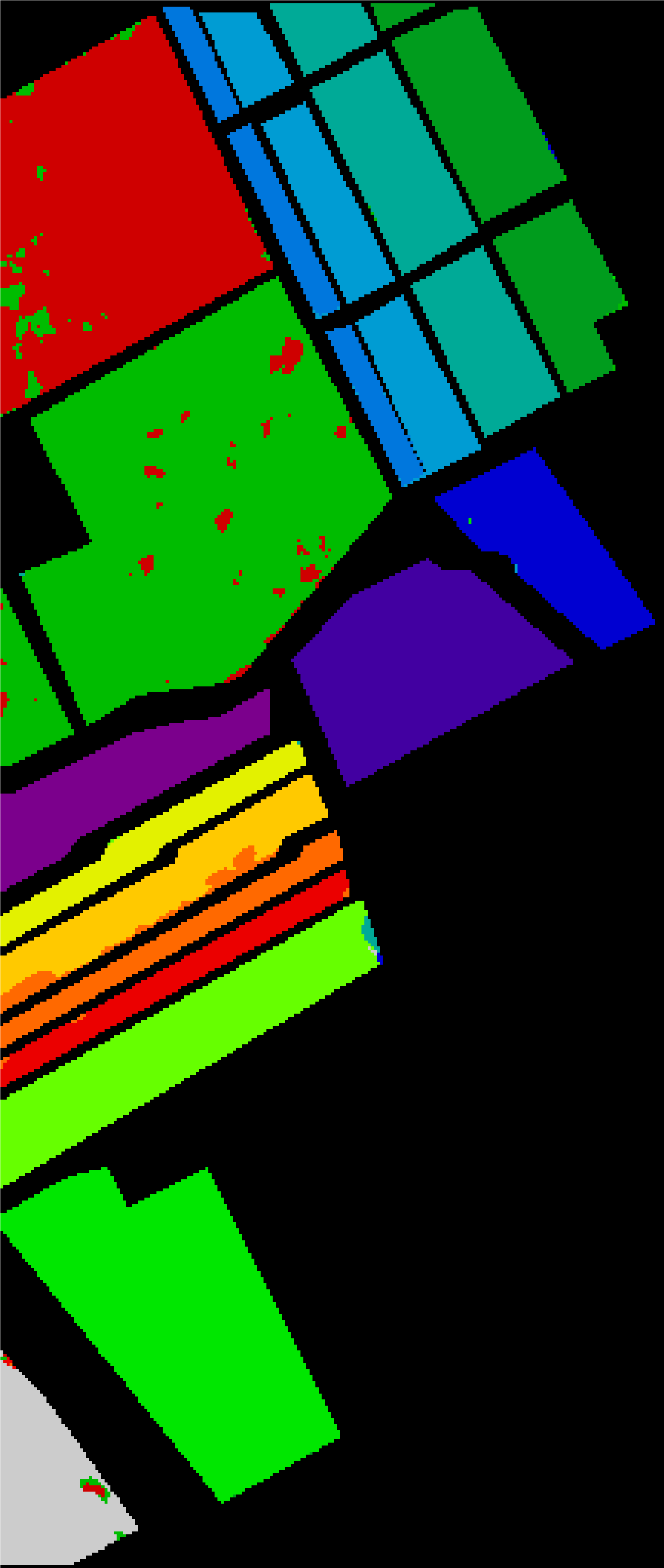}
		\caption{SA:7\%}
		\label{Fig9B}
	\end{subfigure}
	\begin{subfigure}{0.09\textwidth}
		\includegraphics[width=0.99\textwidth]{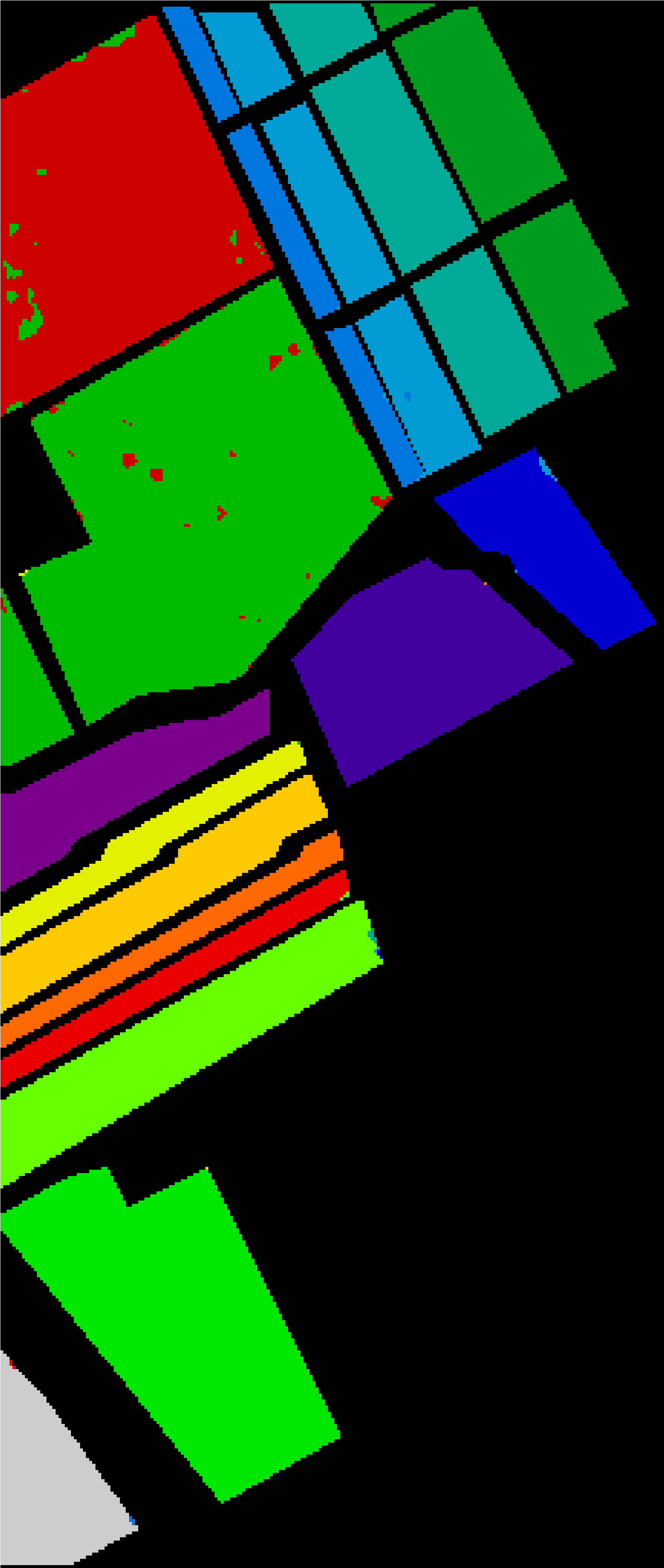}
		\centering
		\caption{SA:10\%}
		\label{Fig9C}
	\end{subfigure}
	\begin{subfigure}{0.09\textwidth}
		\includegraphics[width=0.99\textwidth]{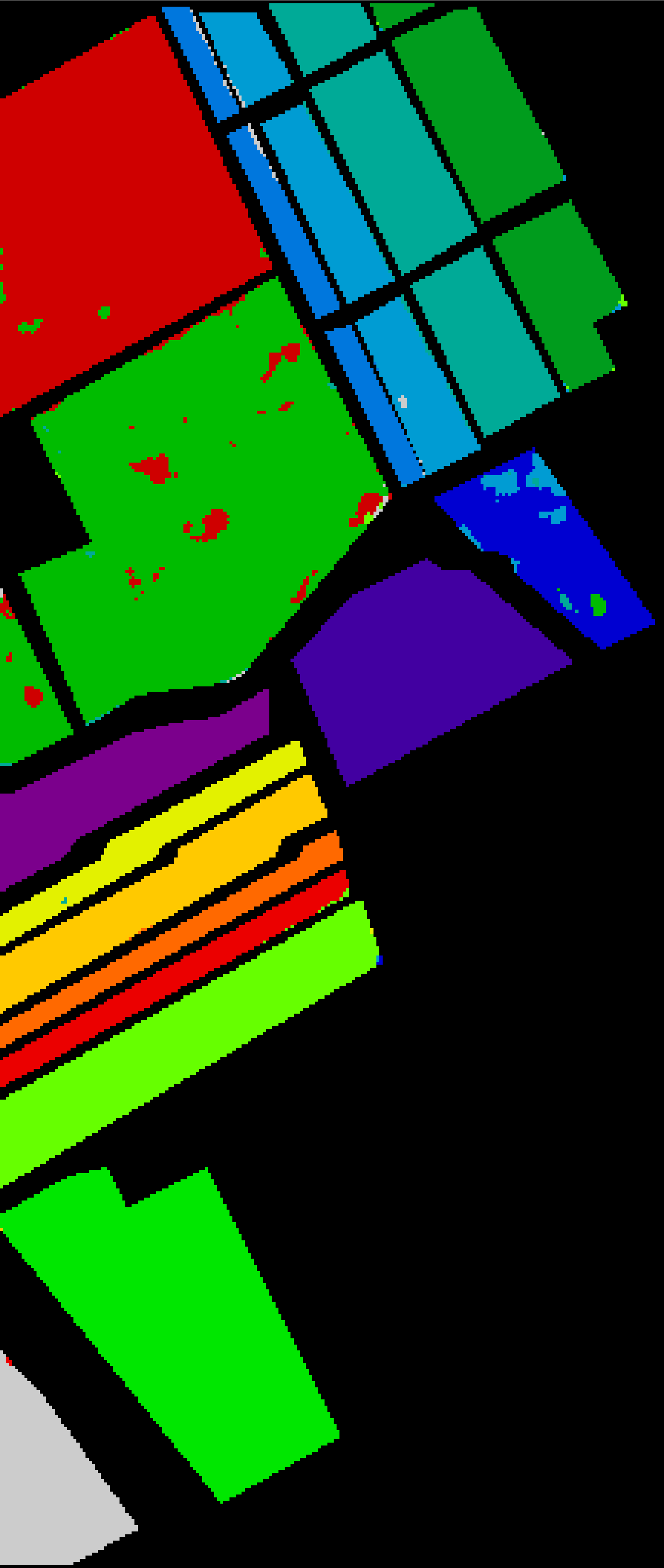}
		\centering
		\caption{SA:12\%} 
		\label{Fig9D}
	\end{subfigure} 
	\begin{subfigure}{0.09\textwidth}
		\includegraphics[width=0.99\textwidth]{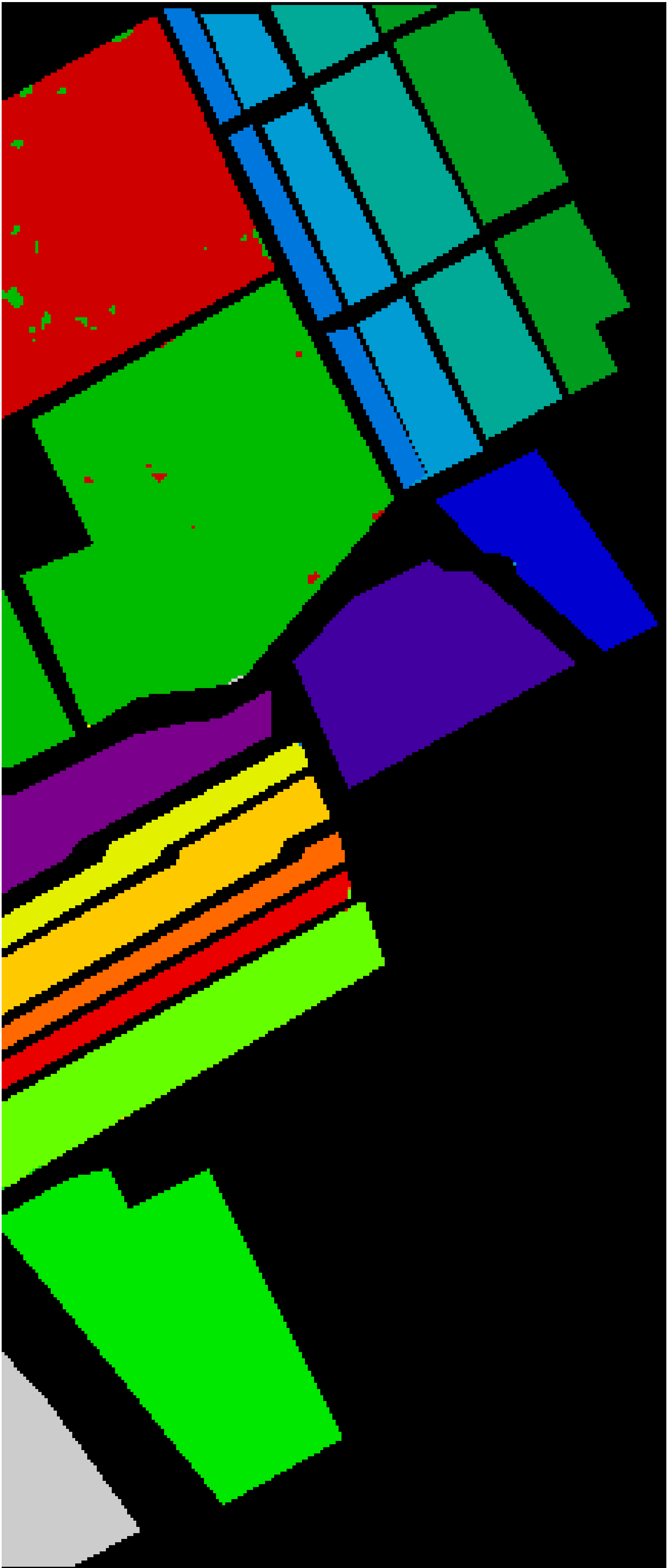}
		\centering
		\caption{SA:15\%} 
		\label{Fig9E}
	\end{subfigure} \\ \vspace{0.3cm}
	
    \begin{subfigure}{0.09\textwidth}
		\includegraphics[width=0.99\textwidth]{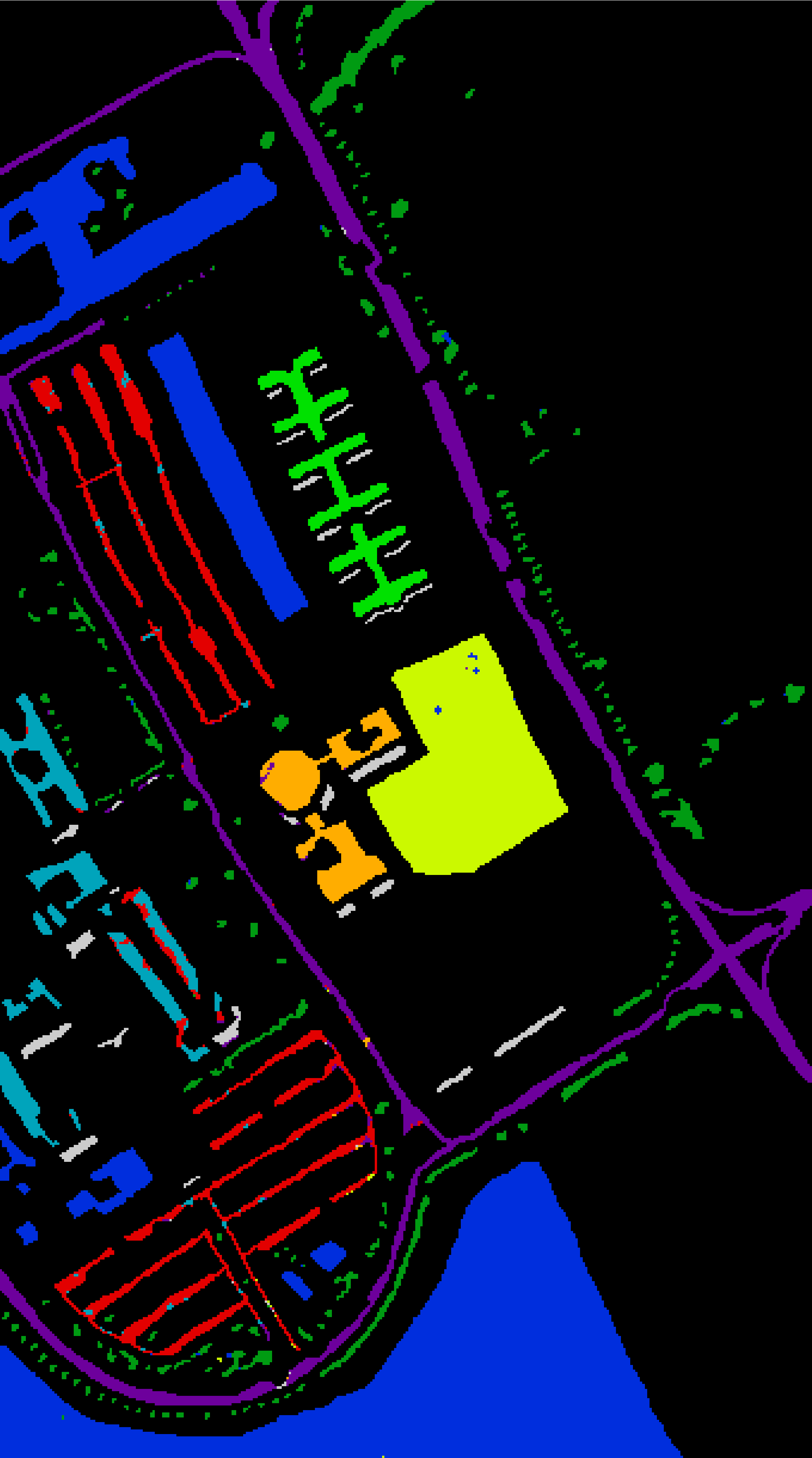}
		\caption{PU:5\%} 
		\label{Fig9A}
	\end{subfigure}
	\begin{subfigure}{0.09\textwidth}
		\includegraphics[width=0.99\textwidth]{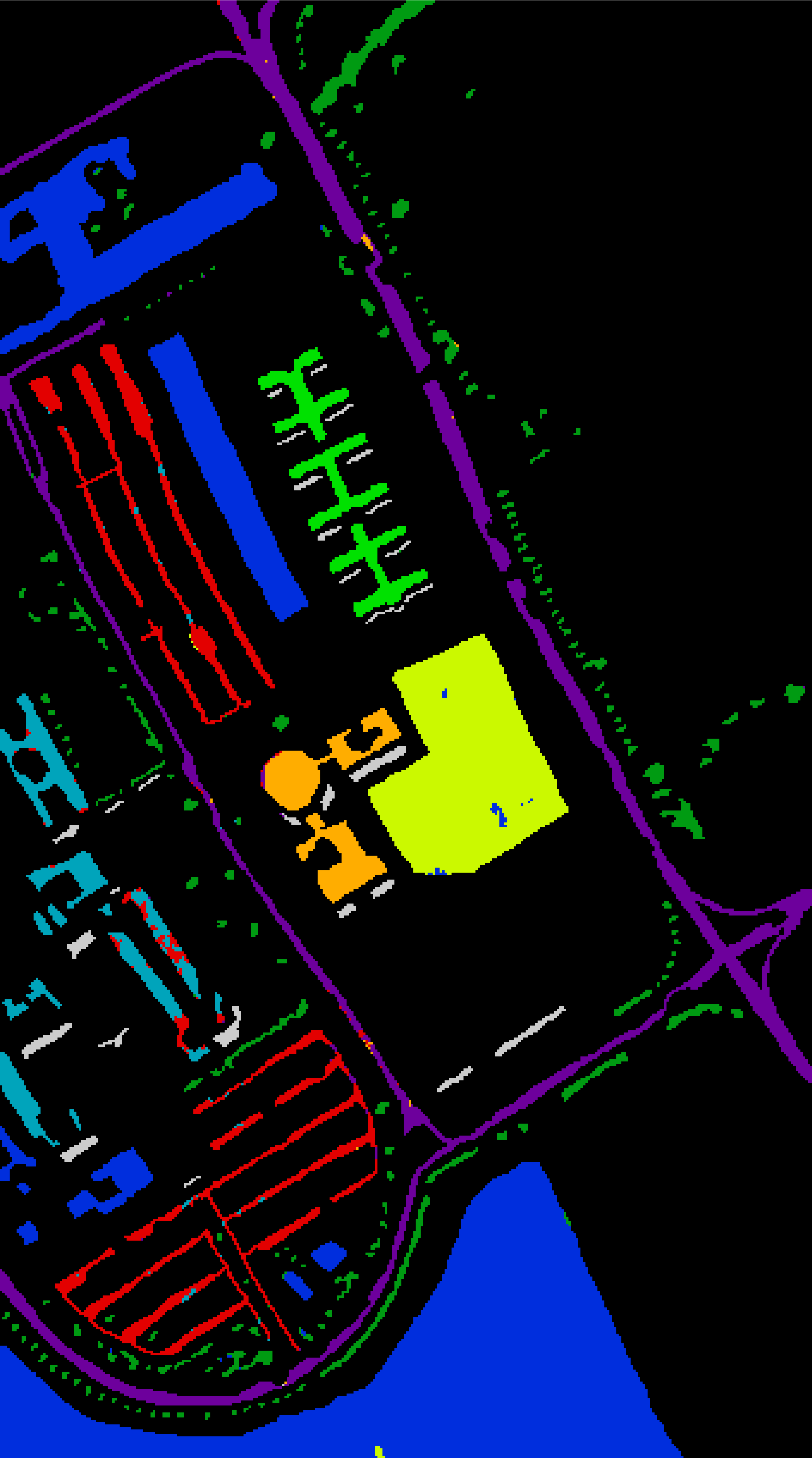}
		\caption{PU:7\%}
		\label{Fig9B}
	\end{subfigure}
	\begin{subfigure}{0.09\textwidth}
		\includegraphics[width=0.99\textwidth]{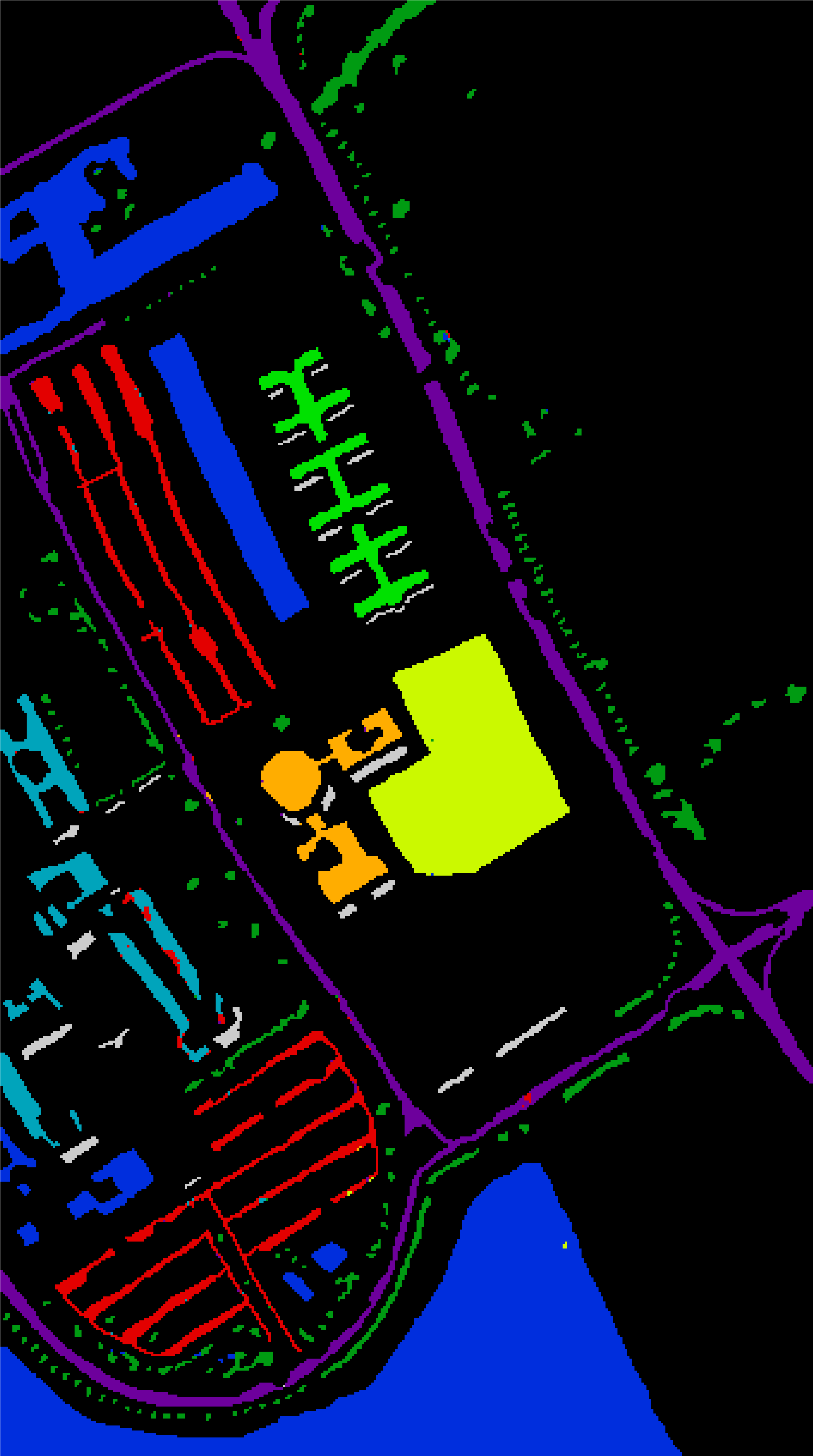}
		\centering
		\caption{PU:10\%} 
		\label{Fig9C}
	\end{subfigure}
	\begin{subfigure}{0.09\textwidth}
		\includegraphics[width=0.99\textwidth]{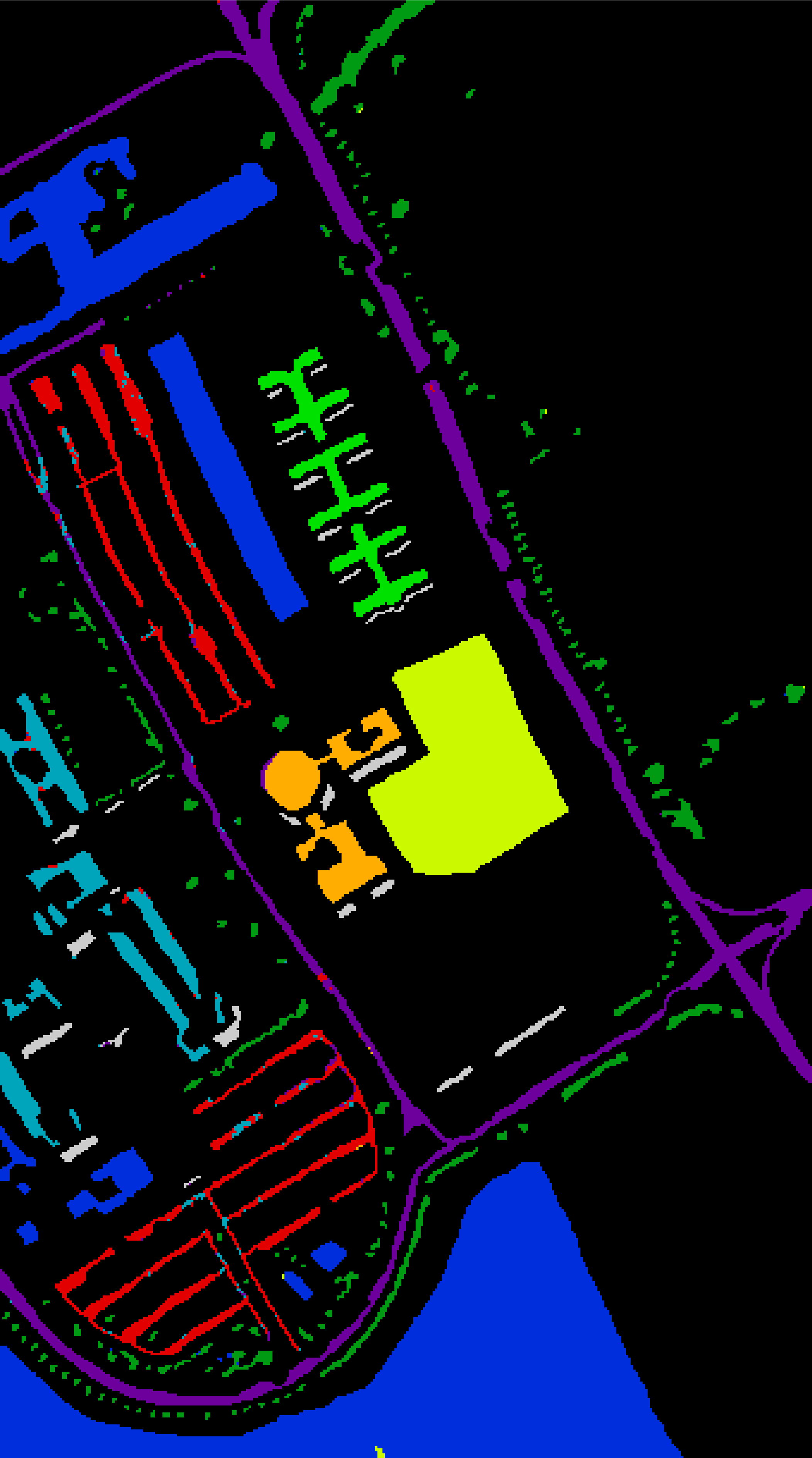}
		\centering
		\caption{PU:12\%} 
		\label{Fig9D}
	\end{subfigure} 
	\begin{subfigure}{0.09\textwidth}
		\includegraphics[width=0.99\textwidth]{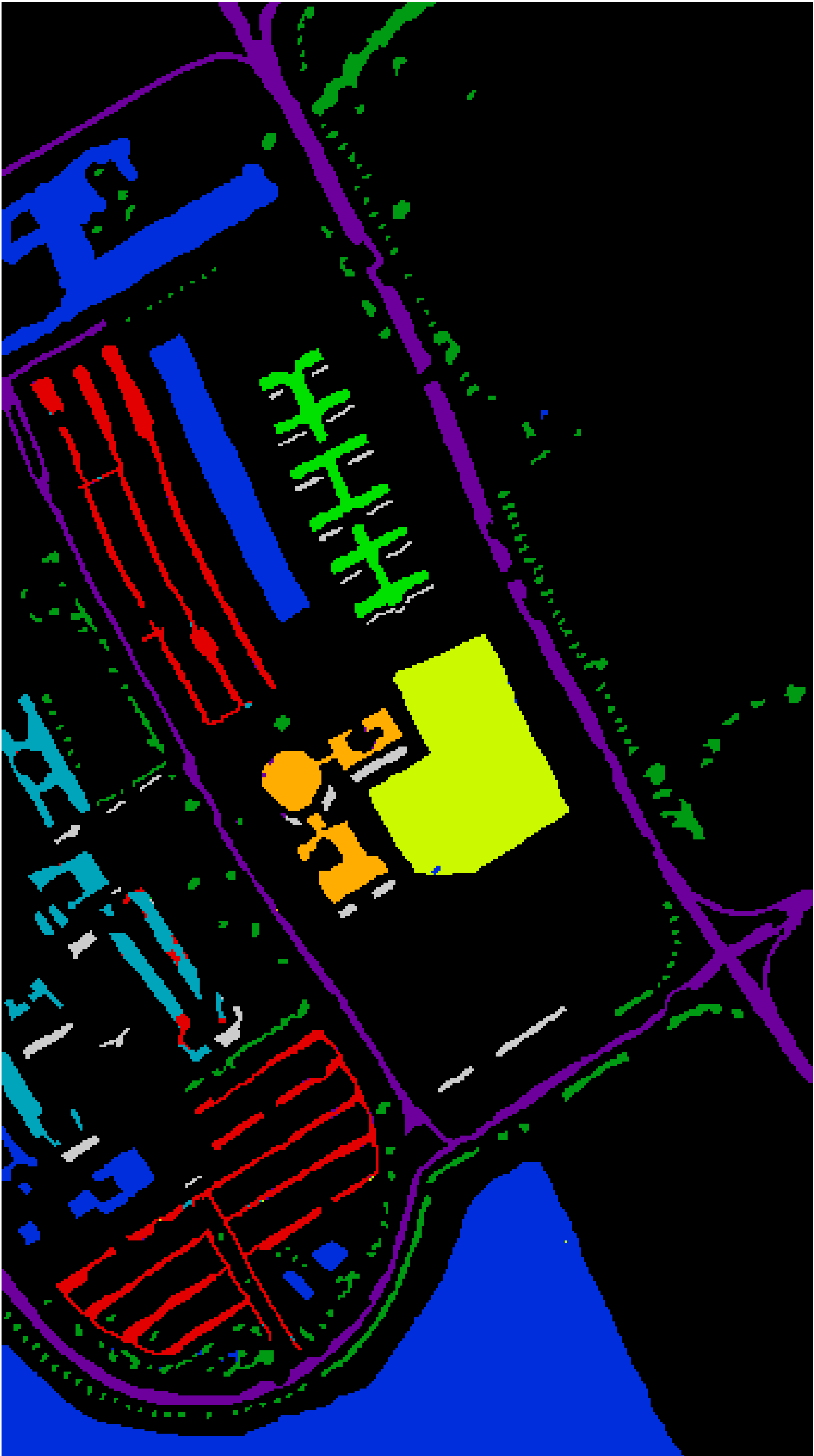}
		\centering
		\caption{PU:15\%} 
		\label{Fig9E}
	\end{subfigure} \\ \vspace{0.3cm}
\caption{Classification performance of different percentages of training samples in terms of ground truth maps.}
\label{Fig.9}
\end{figure}

%%%%%%%%%%%%%%%%%%%%%%%%%%%%%%%%%%%%
Table \ref{Tab.4} and Figure \ref{Fig.9} show the classification performance of our proposed method with different percentages of annotated training samples. One can observe from these results, as the number of annotated training samples increases the classification performance significantly improves. However, the trend is for some certain stage not for the entire group of percentages. This is due to the redundancy among the training samples, as the higher number of annotated samples may contain samples spectrally similar to each other, brings nothing new information, or may lead to confusion for learning. The performance evaluation indicates the quality of spatial-spectral features learned by our proposed model, i.e. the features learned by AfNet. From these results, one can also conclude that the 7-10\% training samples are enough to get satisfactory results. For all these experimental results, $9 \times 9 \times B$ spatial dimension are used, rest of the experimental protocols remains the same except the number of training samples which are further explained in Table \ref{Tab.4}. From the computational time, similar observations can be made, as the number of training samples increases, the training and testing time significantly increase as well the accuracy increases. 

%%%%%%%%%%%%%%%%%%%%%%%%%%%%%%%%%%%%

%%%%%%%%%%%%%%%%%%%%%%%%%%%%%%%%%%%%
\section{Conclusion}
\label{sec6}

Convolution Neural Networks (CNNs) are known to overcome the nonlinearity issues with fixed kernel sizes which are not flexible enough because these kernels are specific and not conducive to feature learning thus impairs classification accuracy. However, CNN with different kernel sizes may capture more discriminative and important features. Thus, taking into account the aforesaid advantages, this work proposed a hybrid (3D-2D) Inception net with an attention mechanism to boost the classification performance. The proposed attention fused hybrid network (AfNet) used attention-based six parallel hybrid sub-nets with different kernels in each sub-block to enhance the final ground-truth maps. The proposed AfNet selectively filters out the discriminative feature i.e., the critical features for classification. AfNet has been tested on several Hyperspectral datasets and shows competitive results as compared to the state-of-the-art models except for a few expensive choices.

%%%%%%%%%%%%%%%%%%%%%%%%%%%%%%%%%%%%
{\footnotesize
\bibliographystyle{IEEEtran}
\bibliography{Sample}}
%%%%%%%%%%%%%%%%%%%%%%%%%%%%%%%%%%%%
\end{document}